\documentclass{article} % For LaTeX2e
\usepackage{iclr2023_conference,times}
\usepackage[T1]{fontenc}

\usepackage{amsfonts, amssymb, amsmath, amsthm, bbm, graphicx, color, tikz, booktabs, multirow, bm, cases, mathtools,  mathrsfs, subcaption, pgfplots, csvsimple,algorithm,algorithmic,wrapfig}
\usetikzlibrary{intersections, pgfplots.fillbetween, shapes, arrows, positioning, decorations.markings}
\definecolor {processblue}{cmyk}{0.96,0,0,0}
\tikzstyle{int}=[draw, fill=blue!20, minimum size=2em]
\tikzstyle{init} = [pin edge={to-,thin,black}]
\usepackage{pgfplotstable}
\usepackage{pgfplots}
\pgfplotsset{compat=1.14}
\usepackage{hyperref, graphics}
\hypersetup{colorlinks=true,linkcolor=blue,filecolor=gray, urlcolor=blue, citecolor=blue}
\newtheorem{defn}{Definition}[section]
\newtheorem{lem}{Lemma}[section]
\newtheorem{rem}{Remark}[section]

\newtheorem{thm}{Theorem}[section]

\newcommand{\ex}[2]{{\ifx&#1& \mathbb{E} \else {\mathbb{E}_{#1}} \fi \left[#2\right]}}

\usepackage[toc,page,header]{appendix}
\usepackage{minitoc}

\usepackage{hyperref}
\usepackage{url}

\title{Over-training with Mixup May Hurt \\ Generalization}

\author{Zixuan Liu$^{1}$\thanks{Equal contribution.} , Ziqiao Wang$^{1}$\footnotemark[1] , Hongyu Guo$^{2, 1}$,Yongyi Mao $^{1}$ \\
$^{1}$University of Ottawa  $^{2}$National Research Council Canada\\
\texttt{\{zliu282, zwang286, hongyu.guo, ymao\}@uottawa.ca}
}
\iclrfinalcopy % Uncomment for camera-ready version, but NOT for submission.
\begin{document}

\maketitle

\begin{abstract}
% Mixup is a regularization technique by using which the model is trained on the virtual examples formulated by convexly combining pairs of the original training examples. In various tasks, Mixup has empirically shown efficient improving the trained models' robustness and generalization compared with the empirical risk minimization (ERM) training policy. However, such a , some of our recent experimental results show that overtrain the models with Mixup may lead to significant degradation in the models' testing performance. In this paper, we 
Mixup, which creates synthetic training instances by linearly interpolating random sample pairs, is a simple and yet effective regularization technique to boost the performance of deep models trained with SGD. 
%Owning to random mixing with randomized mixing coefficient, Mixup is  able to dramatically increase the training sample size, and thus being expected to be beneficial to  longer training time, compared to the  empirical risk minimization (ERM) training scheme. 
In this work, we report a previously unobserved phenomenon in Mixup training: 
on a number of standard datasets, the performance of Mixup-trained models starts to decay after training for a large number of epochs, giving rise to a  U-shaped generalization curve. This behavior is further aggravated when the size of original dataset is reduced. 
%Similar U-shaped generalization curves are also observed when neural networks are trained with a dataset containing a fraction of random labels, where neural networks will learn the clean data first then will overfit to the noisy data.
To help understand such a behavior of Mixup, we show theoretically that Mixup training may introduce undesired data-dependent label noises to the synthesized data. Via  analyzing a least-square regression problem with a random feature model, we explain why  noisy labels may cause the U-shaped curve to occur: Mixup improves generalization through fitting the clean patterns at the early training stage,  but as training progresses, Mixup  becomes over-fitting to the noise in the synthetic data. Extensive experiments are performed on a variety of benchmark datasets, validating this explanation.
\end{abstract}

%%%%%%%%%%%%%%%%%%%%%%%%%%%%%%%%%%%%%%%%%%%%%%%%%%%%

\section{Introduction}
%%%%%%%%%%%%%%%%%%%%%%%%%%%%%%%%%%%%%%%%%%%%%%%%%%%%
% Deep neural networks possess excellent expressive power, and they have gained notable performance in various machine learning tasks. However, under the training scheme of empirical risk minimization (ERM), over-train a network may instead lead to the degradation of its generalization. This issue is what we know as \textbf{over-fitting}. In a intuitive way, over-fitting causes the testing performance of the network to show a U-shaped curve as the training proceeds.

%\begin{figure}[htbp]

Mixup
%, a simple interpolation-based regularization technique,  
has empirically shown its effectiveness in improving the generalization and robustness of 
deep classification models~\citep{zhang2018mixup,guo2019mixup,guotext19,thulasidasan2019mixup,zhang2022and}. 
%the trained networks over a variety of benchmark datasets for classification problems~\citep{}. 
Unlike the vanilla empirical risk minimization (ERM), in which  networks are trained using the original training set, Mixup  trains the networks with %virtual training examples formulated by taking convex combinations of pairs of the real training examples:
synthetic examples. % ($\tilde{\textbf{x}}, \tilde{\textbf{y}}$).
These  examples are created by linearly interpolating  both the input features and the labels of instance pairs %$(\textbf{x}_i,\textbf{y}_i)$ and $(\textbf{x}_j,\textbf{y}_j)$,, 
randomly sampled from the original training set. 
% \iffalse
% , as follows: 
% \begin{equation}
%     \begin{aligned}
%         \tilde{\textbf{x}}&=\lambda\textbf{x}_i+(1-\lambda)\textbf{x}_j,\\
%         \tilde{\textbf{y}}&=\lambda\textbf{y}_i+(1-\lambda)\textbf{y}_j,
%     \end{aligned}
% \end{equation}
% where $\textbf{y}_i$ and $\textbf{y}_j$ are both one-hot labels, and $\lambda\in[0,1]$ is the interpolation coefficient randomly drawn from the Beta distribution $Beta(\alpha,\alpha)$ with a prefixed $\alpha$. 
% \fi
\begin{wrapfigure}{r}{0.7\textwidth}
 \vspace{-2mm}
    \centering
    \subfloat%[Trainloss\label{cifar10 resnet18 trainloss}]
    {%
       \includegraphics[width=0.5078\linewidth]{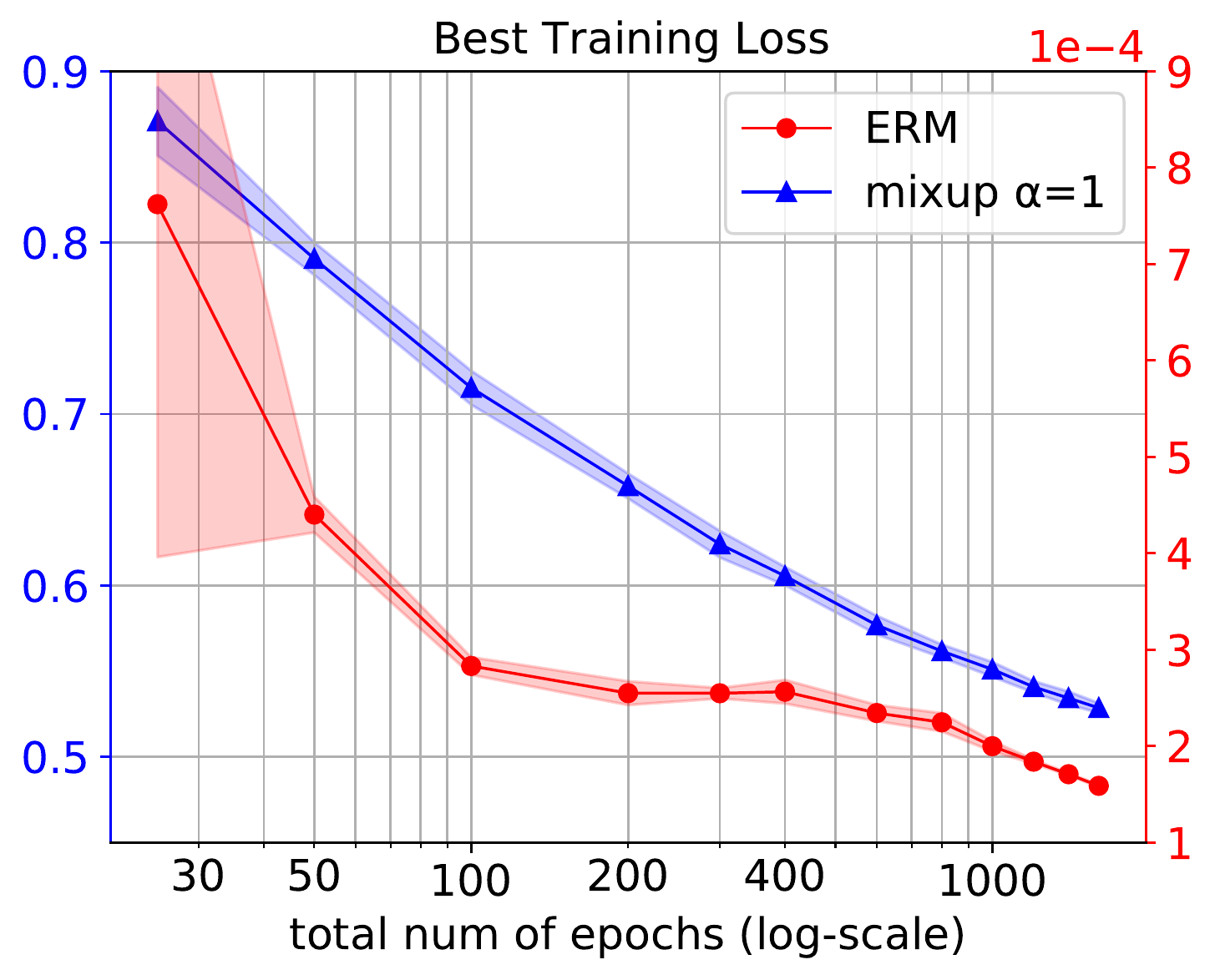}}
       %\hfill
    \subfloat %[Testacc\label{cifar10 resnet18 testacc}]
    {%
       \includegraphics[width=0.4822\linewidth]{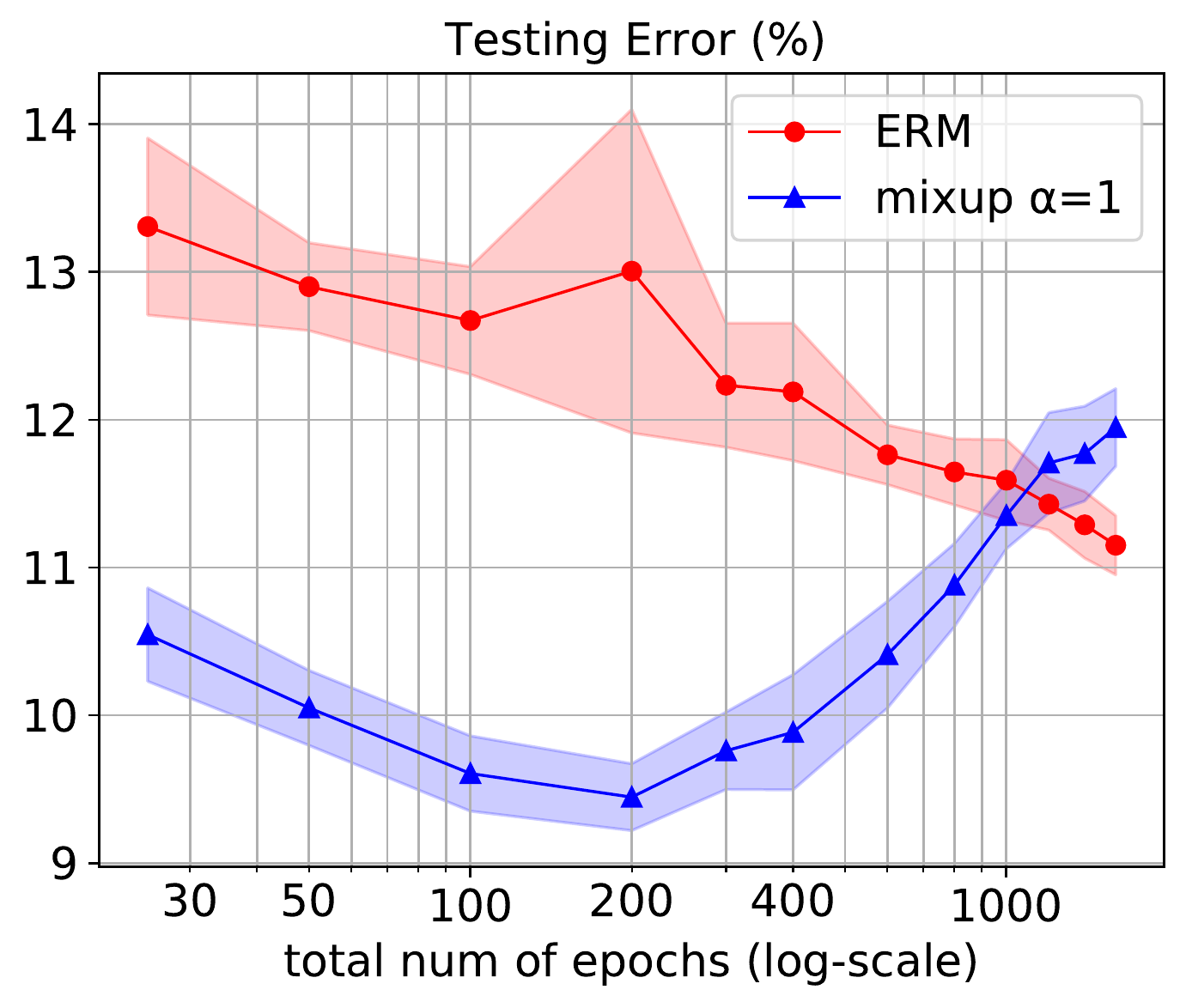}}\\
\vspace{-2mm}
    \caption{Over-training ResNet18 on CIFAR10. %The red and the blue polylines refer to the corresponding results of ERM training and Mixup trainig respectively.
    %It shows that over-training reduces both the ERM and  Mixup training loss, but the testing error of the Mixup-trained ResNet18 decreases first and then gradually increases, while that of the ERM-trained ResNet18 keeps decreasing. 
    }
    \vspace{-3mm}
    \label{fig: CIFAR10 loss & acc curves: introduction}
%\end{figure}
\end{wrapfigure}
% In this work, we report a previously unobserved phenomenon in Mixup training: 
% on a number of standard datasets, the performance of Mixup-trained models starts to decay after training for a large number of epoches, giving rise to a  U-shaped generalization curve. This behavior is further aggravated when the size of original dataset is reduced. 
% To help understand such a behavior of Mixup, we show theoretically that Mixup training may introduce undesired data-dependent label noises to the synthesized data. Via  analyzing a least-square regression problem with a random feature model, we explain why  noisy labels may cause the U-shaped curve to occur: Mixup improve generalization through fitting the clean patterns at the early training stage,  but as training progresses, Mixup  becomes over-fitting to the noise in the synthetic data. Extensive experiments are performed on a variety of benchmark datasets, validating this explanation.
%However, recently from a series of experiments, we have found that the if we overtrain the networks with mixup, although the mixup training loss can be further reduced, the generalization of the trained networks measured by the their testing accuracy may significantly decrease. In this paper, we first verify the existence of this phenomenon from training experiments on the benchmarking datasets including CIFAR10, CIFAR100 and SVHN. Then we provide a theoretical explanation for this. 
%Although Mixup has been empirically shown to be very effective on 
%various classification tasks for neural network models,

Owning to Mixup's simplicity and its effectiveness in boosting the  accuracy and calibration of deep classification models, 
%its working mechanism, training characteristics, and possible limitations have not been well understood. In the recent years, some research works regarding Mixup have begun exploring the potential properties about Mixup. 
there has been a recent surge of interest  attempting to better understand Mixup's working mechanism, training characteristics, regularization potential, and possible limitations (see, e.g., \cite{thulasidasan2019mixup},  \cite{guo2019mixup}, \cite{zhang2021how}, \cite{zhang2022and}). 
% For example, 
%  \cite{thulasidasan2019mixup} empirically show that Mixup helps improve the calibration of the trained networks. \cite{guo2019mixup} identify the  manifold intrusion issue in Mixup, where  the synthetic data ``intrudes'' the data manifolds of the real data. %, resulting in the conflicts between the synthetic labels and the ground-truth labels of the synthetic data.
% \cite{zhang2021how} theoretically explain the effectiveness via analyzing an upper bound of loss function used in Mixup. 
% %yield an upper bound of the Rademacher complexity of the class of functions that the network fits, which in turn bounds the generalization error of the network.
% \cite{zhang2022and}  suggest that the calibration effect of Mixup is correlated with the capacity of the network.  
%In this paper, we are also motivated by 
In this work, 
%we carry out an exploration along these research lines. In this paper, 
we further investigate the  generalization properties of Mixup training.

%\textcolor{red}{Through extensive experiments on various benchmark datasets and deep neural networks, we observe that over-training the networks with Mixup may results in the degradation of the networks' generalization. In other words, the longer we train a network with Mixup, the lower its training loss can be reduced, bu also the worse its testing accuracy becomes. Also, along with the training epochs, the generalization of a network measure by its testing error may form a U-shaped curve. }

We first report a previously unobserved phenomenon in Mixup training. 
Through extensive experiments on various benchmarks, % and network architectures, 
we observe 
%the following behaviors in Mixup training. 
that 
over-training the networks with Mixup may result in significant degradation of the networks' generalization performance. 
%In particular, after certain number of epochs, the longer we train a network with Mixup, the worse its testing accuracy becomes, although the training loss continues to decrease.
As a result, along the training epochs, the generalization performance of the network measured by its testing error may exhibit a U-shaped curve. Figure~\ref{fig: CIFAR10 loss & acc curves: introduction} shows such a curve obtained from over-training   ResNet18~
% \citep{DBLP:journals/corr/HeZR016} 
with Mixup on CIFAR10. %~\citep{CIFARdataset}.  
As can be seen from  Figure~\ref{fig: CIFAR10 loss & acc curves: introduction},
after training with Mixup for a long time (200 epochs),  both  ERM and  Mixup keep decreasing their training loss, but the testing error of the Mixup-trained ResNet18 gradually increases, while that of the ERM-trained ResNet18 continues to decrease.

Motivated by this observation, we  conduct a theoretical analysis, aiming to better  understand the aforementioned behavior of Mixup training. We 
%first 
show theoretically that Mixup training may introduce undesired data-dependent label noises to the synthesized data. 
Then by  analyzing the gradient-descent dynamics of training a random feature model for a least-square regression problem, we explain why  noisy labels may cause the U-shaped curve to occur: under label noise, the early phase of training is primarily driven by the clean data pattern, which moves the model parameter closer to the correct solution. But as training progresses, the effect of label noise accumulates through iterations and gradually over-weighs that of the clean pattern and dominates the late training process. In this phase, the model parameter gradually moves away from the correct solution until it is sufficient apart and approaches a location depending on the noise realization.
\section{Related Work}
%\vspace{-2mm}
\paragraph{Mixup Improves Generalization}
% Mixup \citep{zhang2018mixup} is a data-dependent regularization technique that regularizes the deep neural networks by training them on the synthetic training data. It draws random pairs of the real training data and formulates the synthetic data by convexly combining both the two training examples and their one-hot labels in each pair. 
% Mixup  \citep{zhang2018mixup}  has been empirically shown to be effective in improving the networks generalization and robustness for various benchmark dataset including CIFAR10, CIFAR100, SVHN, \textit{etc}. In addition,
After the initial work of \cite{zhang2018mixup}, a series of the Mixup's variants have been proposed~\citep{guo2019mixup,verma19a,yun2019cutmix,kimICML20,greenawald2021kmixup,han2022g,sohn2022genlabel}. For example, \textit{AdaMixup} \citep{guo2019mixup} trains an extra network to dynamically determine the interpolation coefficient parameter $\alpha$. \textit{Manifold Mixup} \citep{verma19a} performs linear mixing on the hidden states of the neural networks. %Other versions of Mixup see \citep{yun2019cutmix,greenawald2021kmixup,sohn2022genlabel}. 
% \textit{CutMix} \citep{yun2019cutmix} mixes each pair of the training images by extracting a part from each of the two images and merging them together. \textit{\boldmath{$k$}-Mixup} \citep{greenawald2021kmixup} randomly draws two batches of $k$ training examples and pairs the examples in the way that minimizes the Wasserstein distance between the two batches.  \textit{GenLabel} \citep{sohn2022genlabel} reassigns the training targets for the synthetic data using an additionally trained generative model about the training data. 
Aside from its use in various applications, Mixup's working mechanism and it possible limitations are also being explored constantly. 
% and have not been fully understood. 
For example, \cite{zhang2021how} demonstrate that Mixup yields a generalization upper bound in terms of the Rademacher complexity of the function class that the network fits.  \cite{thulasidasan2019mixup}   show that Mixup helps to improve the calibration of the trained networks. % and that the networks.
 \cite{zhang2022and}  theoretically justify that the calibration effect of Mixup is correlated with the capacity of the network. Additionally,
%  \cite{guo2019mixup} introduce the concept of manifold intrusion. 
 %, which is an issue that may hurt the performance of Mixup.
  \cite{guo2019mixup}
% It
point out a ``manifold intrusion'' phenomenon in Mixup training where the synthetic data ``intrudes'' the data manifolds of the real data. %, resulting in the conflicts between the synthetic labels and the ground-truth labels of the training data. 

\vspace{-2mm}

\paragraph{Training on Random Labels, Epoch-Wise Double Descent and Robust Overfitting} The thought-provoking work of \cite{zhang2017understanding} highlights that neural networks are able to fit data with random labels. After that,  the generalization behavior on corrupted label dataset has been widely investigated \citep{arpit2017closer,liu2020early,feng2021phases,wang2022on,pmlr-v162-liu22w}. Specifically, \cite{arpit2017closer} observe that neural networks will learn the clean pattern first before fitting to data with random labels. This is further explained by \cite{arora2019fine} where they demonstrate that in the overparameterization regime, the convergence of loss depends on the projections of labels on the eigenvectors of some Gram matrix, 
% and these projections are  different for
where true labels and random labels have different projections. In a parallel line of research, an \textit{epoch-wise} double descent behavior of testing loss of deep neural networks is observed in \cite{Nakkiran2020Deep}, shortly after the observation of the \textit{model-wise} double descent  \citep{belkin2019reconciling,hastie2022surprises,mei2022generalization,Ba2020Generalization}. Theoretical works studying the \textit{epoch-wise} double descent are rather limited to date  \citep{heckel2021early,stephenson2021and,pezeshki2022multi}, among which \cite{advani2020high} inspires the theoretical analysis of the U-sharped curve of Mixup in this paper. Moreover, robust overfitting \citep{rice2020overfitting} is also another yet related research line,
% which is found by \cite{rice2020overfitting}. 
In particular, robust overfitting is referred to a phenomenon in adversarial training that robust accuracy will first increase then decrease after a long training time. \cite{dong2022label} show that robust overfitting is deemed to the early part of epoch-wise double descent due to the \textit{implicit label noise} induced by adversarial training. Since Mixup training has been connected to adversarial training or adversarial robustness in the previous works \citep{archambault2019mixup,zhang2021how}, the work of \cite{dong2022label} indeed motivates us to study the label noise induced by Mixup training. 
%\vspace{-2mm}
\section{Preliminaries}
%\vspace{-2mm}
%\subsection{Empirical Risk Minimization (ERM)}
% \vspace{-1mm}
Consider a $C$-class classification setting with input space ${\cal X}=\mathbb{R}^{d_0}$ and label space ${\cal Y}:=\{1, 2, \ldots, C\}$. 
%Denote by ${\cal P}({\cal Y})$ the space of distributions over ${\cal Y}$. 
Let $S=\{(\textbf{x}_i,\textbf{y}_i)\}_{i=1}^n$ be a training set, where each $\textbf{y}_i \in {\cal Y}$ may also be treated as a one-hot vector in ${\cal P}({\cal Y})$,  the space of distributions over ${\cal Y}$. 
Let $\Theta$ denote the model parameter space, and for each $\theta\in\Theta$, let $f_\theta:\mathcal{X}\rightarrow[0,1]^C$ denote the predictive function associated with $\theta$, which maps an input feature to a distribution in ${\cal P}({\cal Y})$. For any pair  $(\textbf{x},\textbf{y})\in {\cal X}\times {\cal P}({\cal Y})$, let
$\ell(\theta,\textbf{x},\textbf{y})$ denote the loss of the prediction $f_\theta(\textbf{x})$ with respect to 
%the target label 
$\textbf{y}$. 
% which can adapt any forms of defining the distance between two distributions, 
%(e.g., the cross-entropy loss, mean squared error (MSE), \textit{etc}). 
The empirical risk of 
%the predictor 
$\theta$ on $S$ is then
\[
    \hat{R}_S(\theta):=\frac{1}{n}\sum_{i=1}^{n}\ell(\theta,\textbf{x}_i,\textbf{y}_i).
\vspace{-0.3mm}
\]
When training with Empirical Risk Minimization (ERM), one sets out to find a $\theta^*$ to minimize this risk.
% \[\theta^*:=\mathop{\arg \min}\limits_{\theta\in\Theta}\hat{R}_S(\theta).\]
It is evident that if $\ell(\cdot)$ is taken as the cross-entropy loss, the empirical risk  $\hat{R}_S(\theta)$ is non-negative, where $\hat{R}_S(\theta)=0$ precisely when 
$f_\theta(\textbf{x}_i)=\textbf{y}_i$ for every $i=1, 2, \ldots, n$.

%\subsection{Mixup}
In Mixup, instead of using the original training set $S$, the training is performed on a synthetic dataset 
$\widetilde{S}$ 
obtained by interpolating training examples in $S$. 
%Without loss of generality, we consider the synthetic dataset for each 
For a given interpolating parameter $\lambda \in [0, 1]$, let synthetic training set $\widetilde{S}_\lambda$ be defined as
\begin{equation}
    \widetilde{S}_\lambda:= \{
(\lambda\textbf{x}+(1-\lambda)\textbf{x}',
        \lambda\textbf{y}+(1-\lambda)\textbf{y}'): ({\textbf{x}},{\textbf{y}})\in S, 
        ({\textbf{x}}',\textbf{y}')\in S\} 
\end{equation}

%Note that in practice, $\lambda\in[0,1]$ is drawn from some prescribed  distribution, independently across for all example pairs, which would correspond to sampling from $\widetilde{S}_\lambda$ for a random $\lambda$.  
The optimization objective, or the ``Mixup loss'',  is then  
\[
\mathbb E_\lambda\hat{R}_{\widetilde{S}_{\lambda}}(\theta
):=
\mathbb E_\lambda
\frac{1}{|\widetilde{S}_\lambda|}\sum_{(\tilde{\textbf{x}}, 
\tilde{\textbf{y}})\in \widetilde{S}_\lambda
}\ell(\theta,\tilde{\textbf{x}},\tilde{\textbf{y}})
\]
% Mixup training aims to find a $\theta^*$ that minimizes the Mixup loss.
where the interpolating parameter $\lambda$ is drawn from a symmetric Beta distribution, ${\rm Beta}(\alpha, \alpha)$. The default option is to take $\alpha=1$.  In this case, the following can be proved.

% . The general Mixup loss of a model $f_\theta$ on each $(\tilde{\textbf{x}},\tilde{\textbf{y}})$ can be denoted by $\ell(\theta,(\tilde{\textbf{x}},\tilde{\textbf{y}}))$. For a given $\lambda$, we define the empirical Mixup loss as $\hat{R}_{\widetilde{S}}(\theta,\lambda):=\frac{1}{n^2}\sum_{i=1}^{n}\sum_{j=1}^{n}\ell(\theta,(\tilde{\textbf{x}},\tilde{\textbf{y}}))$. Normally the $\lambda$'s are drawn randomly from a Beta distribution $Beta(\alpha,\alpha)$ with a prefixed $\alpha>0$. Thus, we take the expectation of $\hat{R}_{\widetilde{S}}(\theta,\lambda)$ and define it to be the Mixup loss: $\hat{R}_{\widetilde{S}}(\theta,\alpha):=\mathop{\mathbb{E}}_{\lambda\sim{Beta(\alpha,\alpha)}}\frac{1}{n^2}\sum_{i=1}^{n}\sum_{j=1}^{n}\ell(\theta,(\tilde{\textbf{x}},\tilde{\textbf{y}}))$. The optimization objective of Mixup training is defined as: $\theta^*:=\mathop{\arg \min}\limits_{\theta\in\Theta}\hat{R}_{\widetilde{S}}(\theta,\alpha)$. The following lemma shows the lower bound of the Mixup loss.
\begin{lem}
% Suppose that 
Let $\ell(\cdot)$ be the cross-entropy loss, and 
%$\{\lambda\}$ is drawn i.i.d. 
$\lambda$ is drawn 
from ${\rm Beta}(1, 1)$ (or the uniform distribution on $[0, 1]$). Then for all $\theta\in\Theta$ and 
for any given training set $S$ that is balanced,
% $S=\{(\textbf{x}_i,\textbf{y}_i)\}_{i=1}^n$ of a $C$-class classification task and a model $f_\theta$, if we adapt the definition of the cross-entropy loss on the general loss functions and particularly draw $\lambda$'s from $Beta(\alpha,\alpha)$, then we have:
\[
    {\mathbb E}_{\lambda}\hat{R}_{\widetilde{S}_\lambda}(\theta)\geq\dfrac{C-1}{2C},%\times0.5
\]
where the equality holds if and only if ${f}_\theta(\tilde{\textbf{x}})=\tilde{\textbf{y}}$ for each synthetic example $(\tilde{\textbf{x}},\tilde{\textbf{y}})\in \widetilde{S}_\lambda$.
\label{lem:mixup loss lower bound}
\end{lem}
% The lower bound %$\dfrac{C-1}{2C}$ 
% in the lemma 
%This lower bound enables us to understand the Mixup loss during training. For example, 
For 10-class classification tasks, the bound has value 0.45. Then only when the Mixup loss approaches this value, %we may conclude that %the model parameter is near an optimum 
the found solution is near a true optimum (for models with adequate capacity).
% an optimum is found (assuming the model has sufficient capacity). 

%\vspace{-2mm}
\section{Empirical Observations}
%\vspace{-2mm}
%In this section,
%We here  empirically show that over-training with  Mixup may hurt the networks' generalization.
%first introduce the outlines of the experiments. We then show the results from which we have observed the phenomenon that over-training the deep neural networks with Mixup may hurt the networks' generalization.
%\subsection{Experiments Setup}
%We are interested in observing whether over-training a deep neural network with Mixup makes any difference on the model's performance; if it does, we are also interested in whether the existence of the effect depend on the size of the training set.  

We conduct experiments using CIFAR10, CIFAR100 and SVHN using ERM and Mixup respectively. For each of the datasets, we have adopted both the original dataset and some balanced subsets obtained by downsampling the original data for certain proportions.
%For CIFAR10 and SVHN, we adopt both the original training set and a balanced subset containing 30\% of the original data. For CIFAR100, we only use the original training set, since downsampling  CIFAR100  appears to result in testing performance baselines with high variances. 
SGD with weight decay is used.
%due to the small sample size for each of the 100 classes. 
%We train ResNet networks on \textcolor{red}{these} 
%the three datasets 
%using both ERM and  Mixup, while adopting SGD with weight decay. 
%In order to remove the extra randomness besides that induced by random $\lambda$'s, 
% No data augmentation is used.
%in  the  experiments except otherwise specified. 
At each epoch, we record the minimum training loss up to that epoch, as well as the testing accuracy at the epoch achieving the minimum training loss. Results are obtained both for training with data augmentation and for training without. More experimental details are given in Appendix~\ref{appendix:overtrain experiments setups}.
%In each training trial, we record the achieved training loss and the corresponding epoch, as well as the network's testing accuracy at that epoch. %We show that by increasing the total number of the training epochs, the minimal Mixup training loss can be further reduced, while the corresponding testing accuracy of the model drops significantly.

% \vspace{-2mm}
\subsection{Results on Over-Training Without Data Augmentation}\label{subsec:results of cifar10}
% \vspace{-2mm}

\begin{figure}[htbp]
    \centering
    \subfloat[Train loss ($30\%$ {data})\label{subfig:0.3 cifar10 resnet18 trainloss}]{%
       \includegraphics[width=0.2625\linewidth]{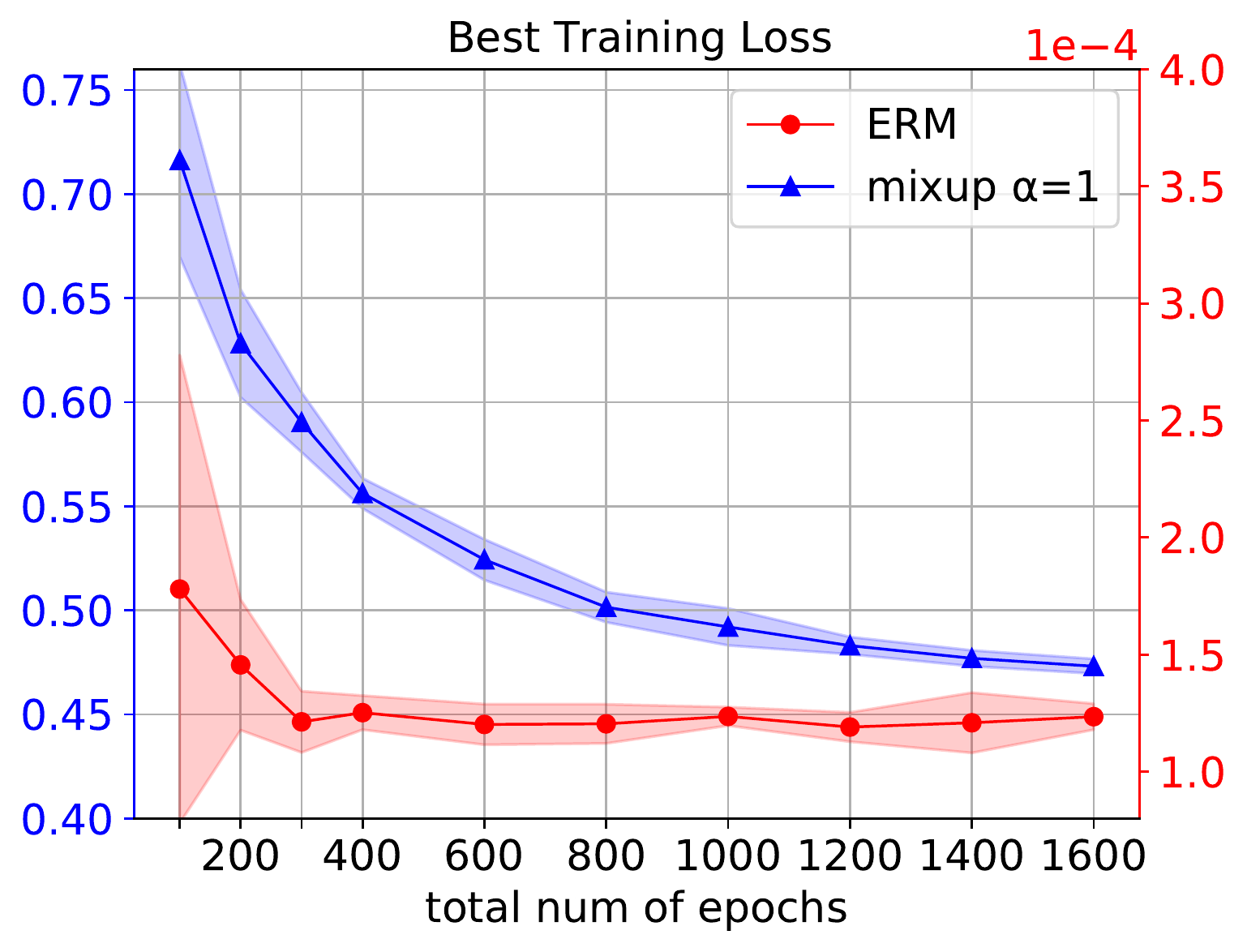}}
       \hfill
    \subfloat[Test acc ($30\%$ {data})\label{0.3 cifar10 resnet18 testacc}]{%
       \includegraphics[width=0.2375\linewidth]{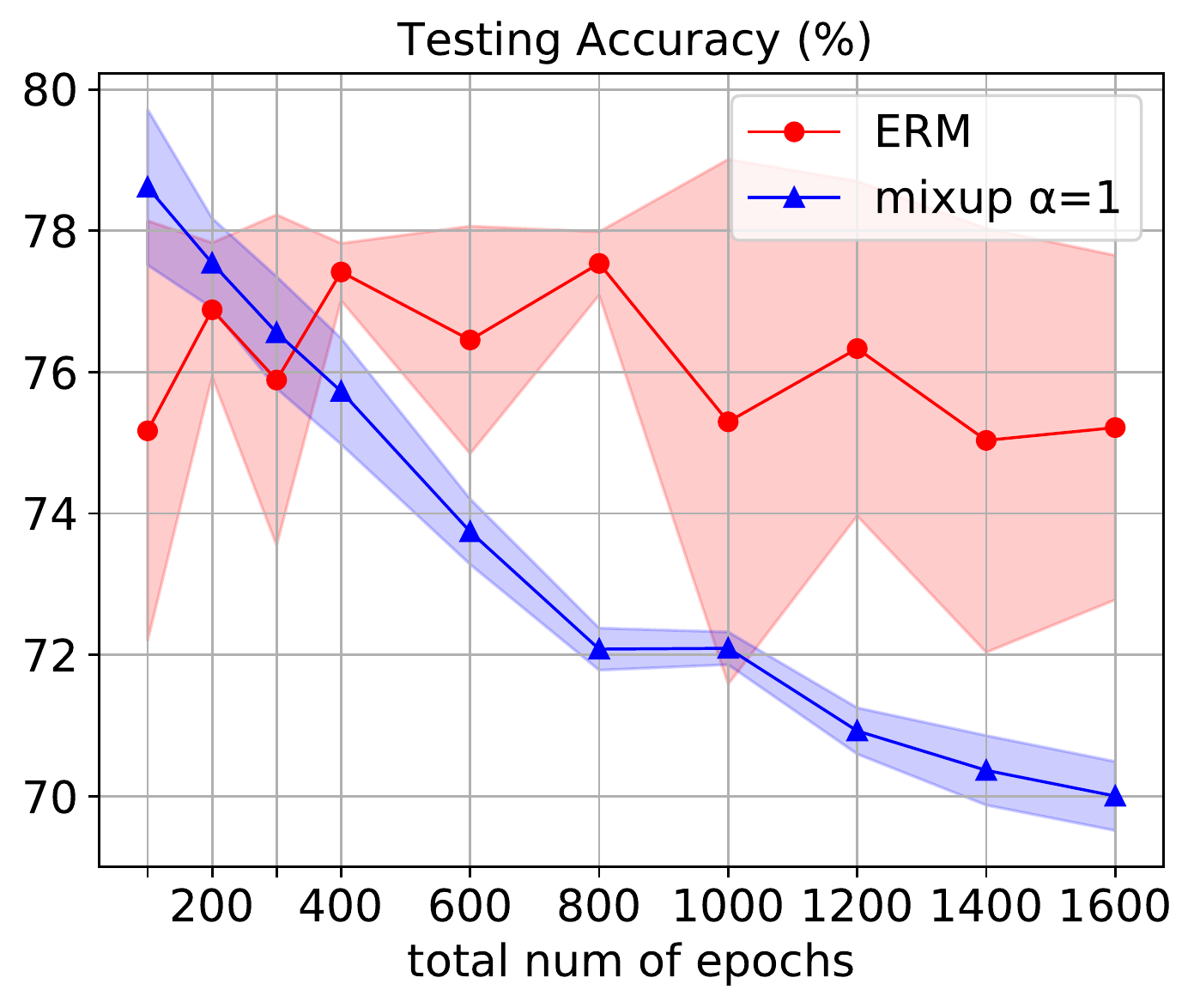}}
       \hfill
    \subfloat[Train loss ($100\%$ {data})\label{cifar10 resnet18 trainloss}]{%
       \includegraphics[width=0.2625\linewidth]{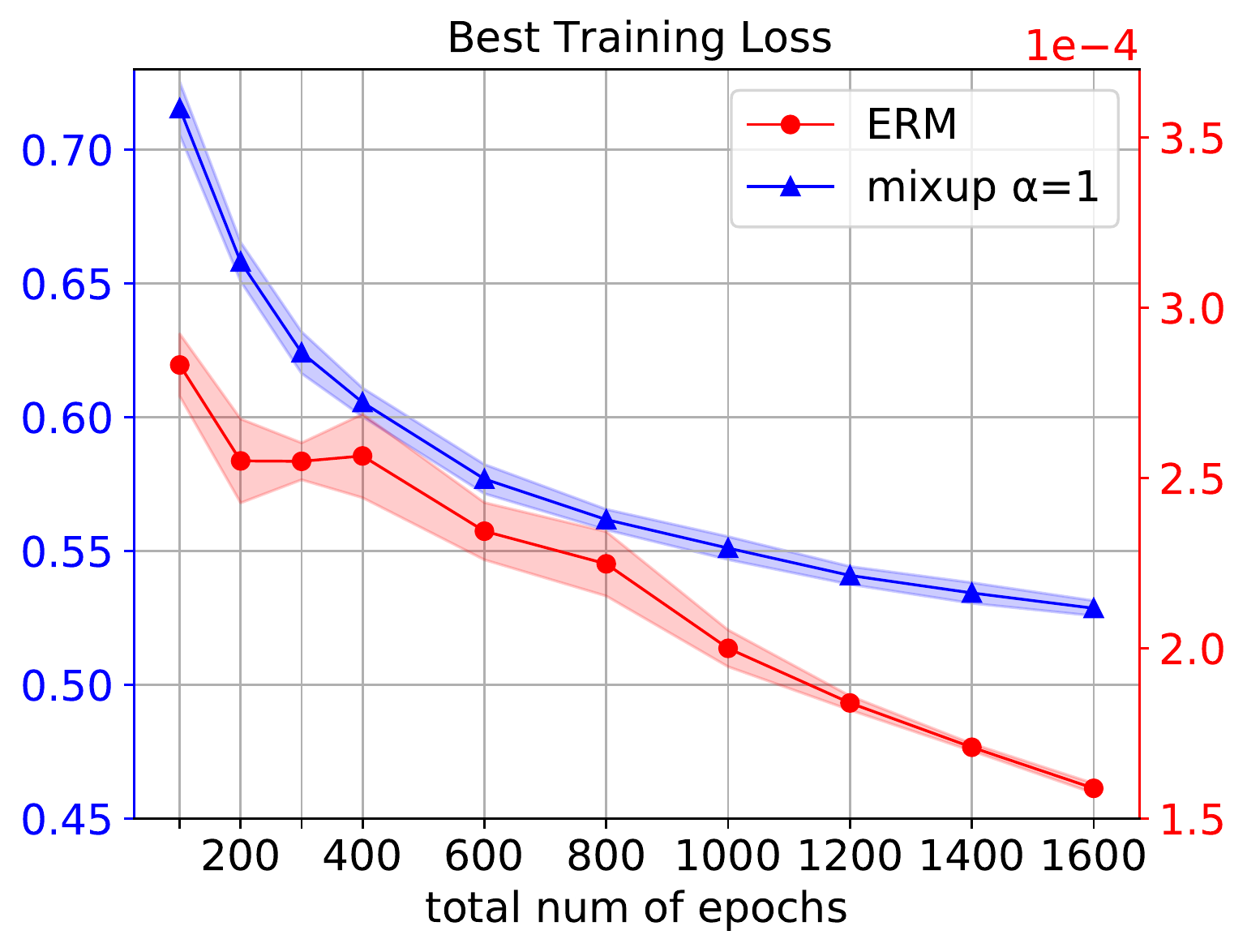}}
       \hfill
    \subfloat[Test acc ($100\%$ {data})\label{cifar10 resnet18 testacc}]{%
       \includegraphics[width=0.2375\linewidth]{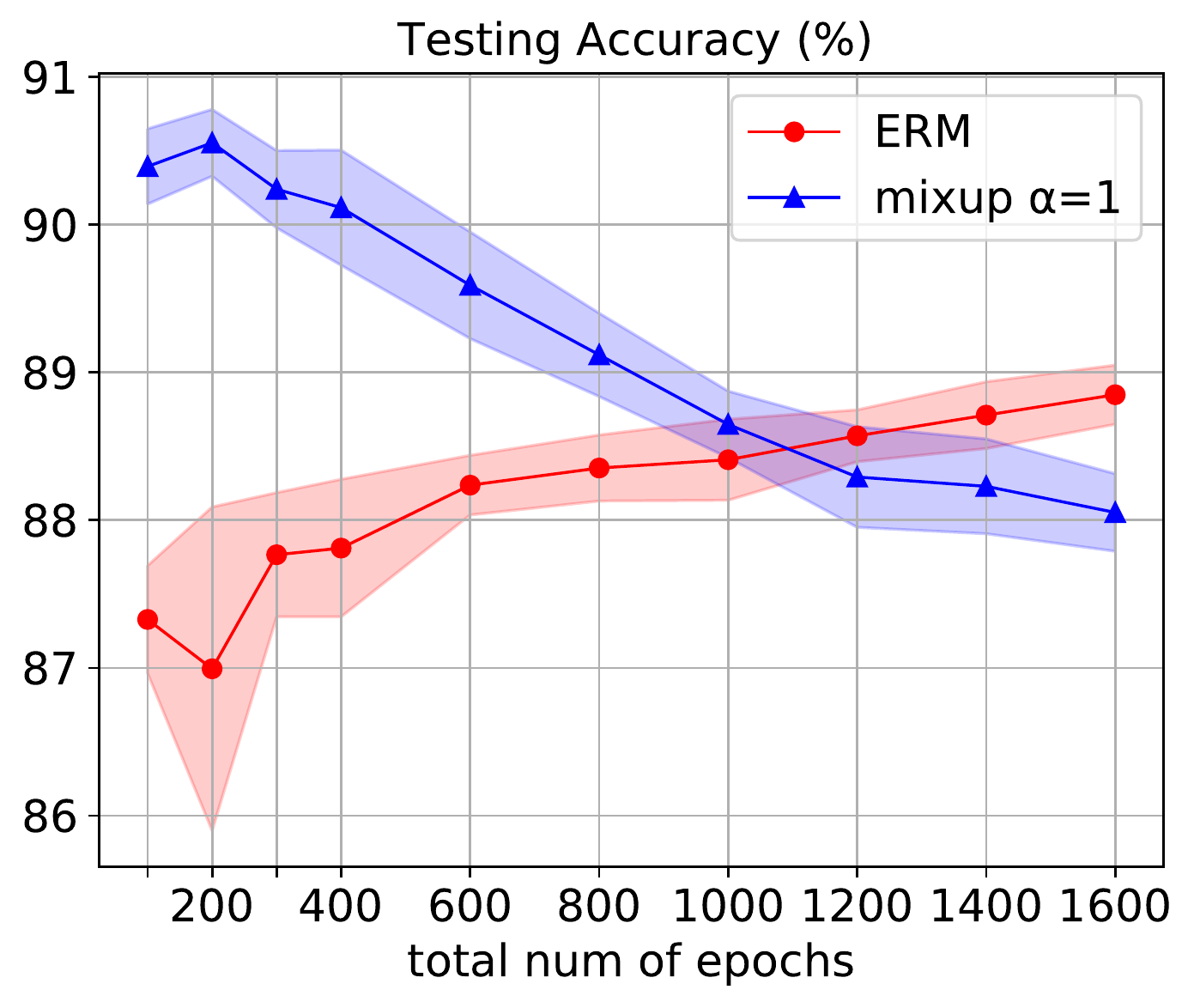}}\\
    \subfloat[200 epochs\label{0.3 cifar10 resnet18 landscape e200}]{%
       \includegraphics[width=0.16\linewidth]{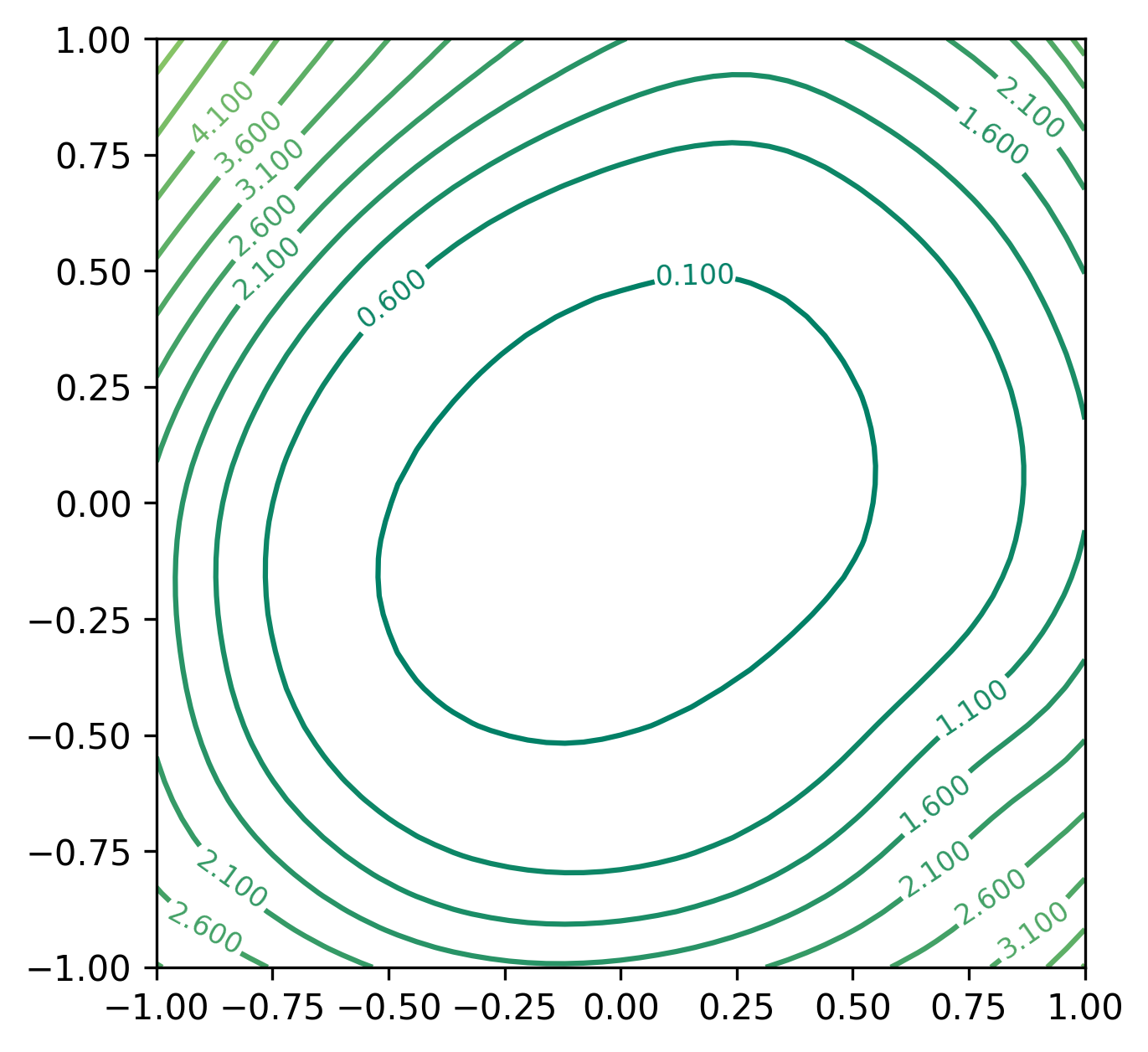}}
       \hfill
    \subfloat[400 epochs\label{0.3 cifar10 resnet18 landscape e400}]{%
       \includegraphics[width=0.16\linewidth]{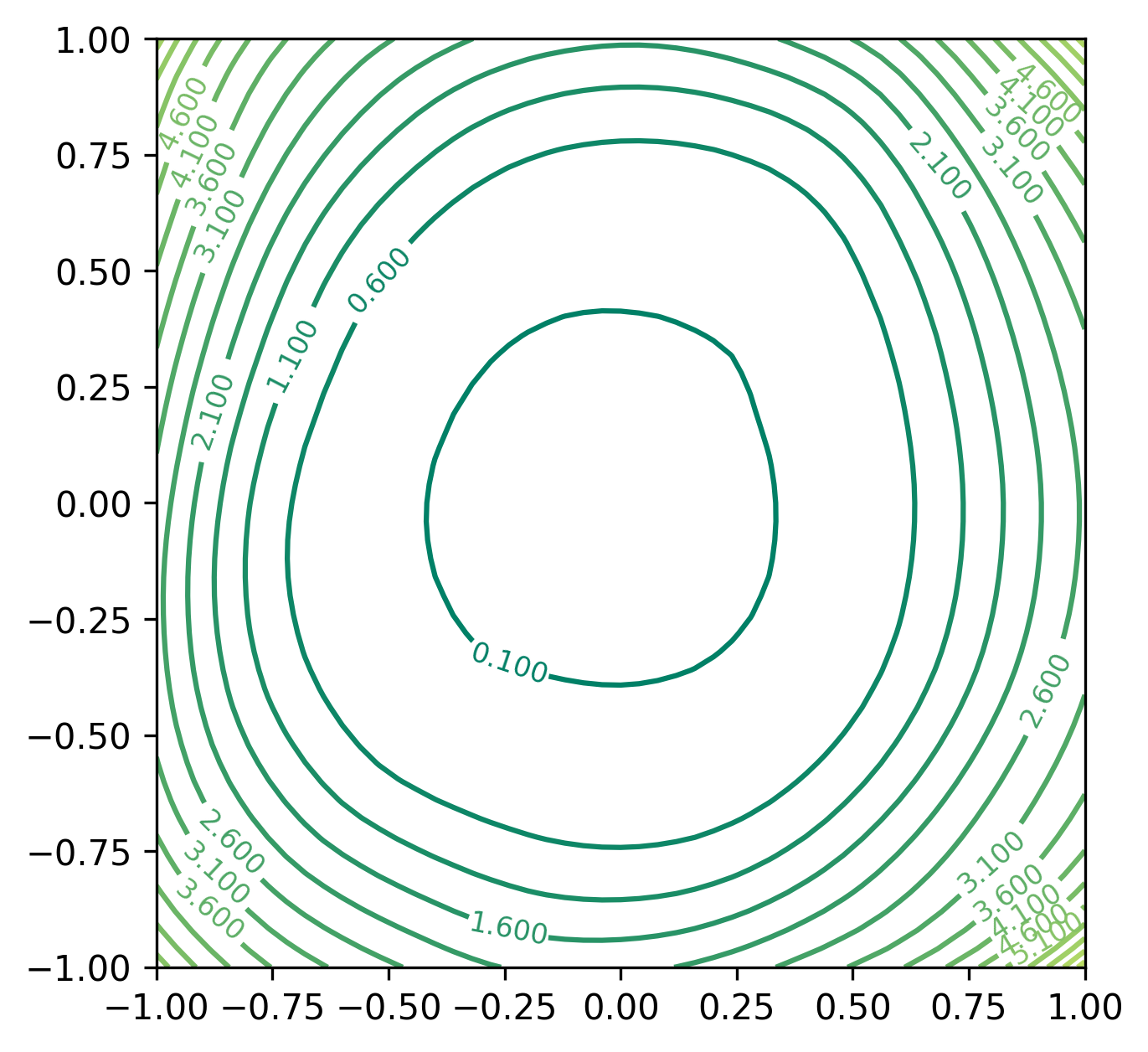}}
       \hfill
    \subfloat[800 epochs\label{0.3 cifar10 resnet18 landscape e800}]{%
       \includegraphics[width=0.16\linewidth]{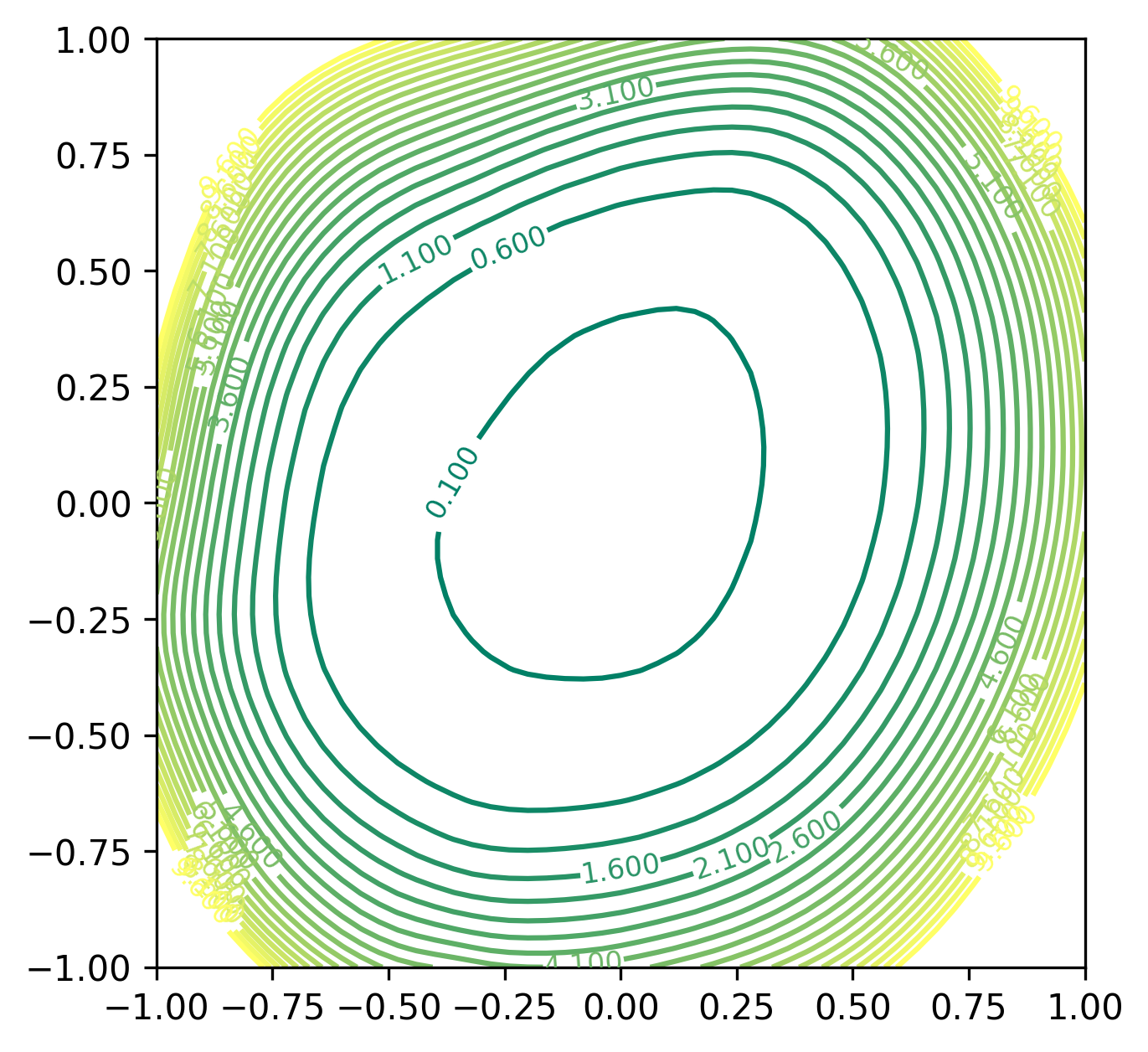}}
       \hfill
    \subfloat[200 epochs\label{cifar10 resnet18 landscape e200}]{%
       \includegraphics[width=0.16\linewidth]{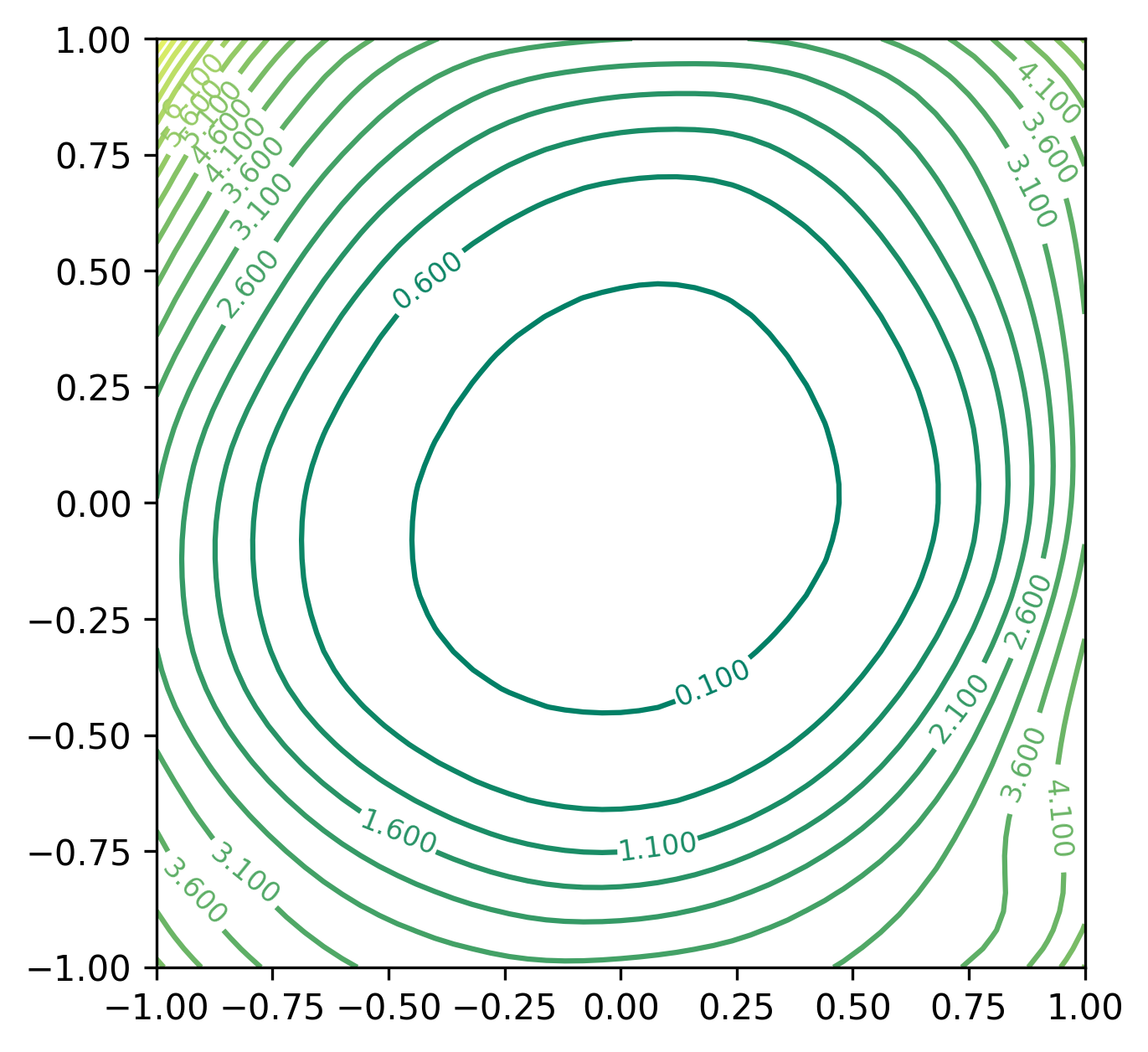}}
       \hfill
    \subfloat[400 epochs\label{cifar10 resnet18 landscape e400}]{%
       \includegraphics[width=0.16\linewidth]{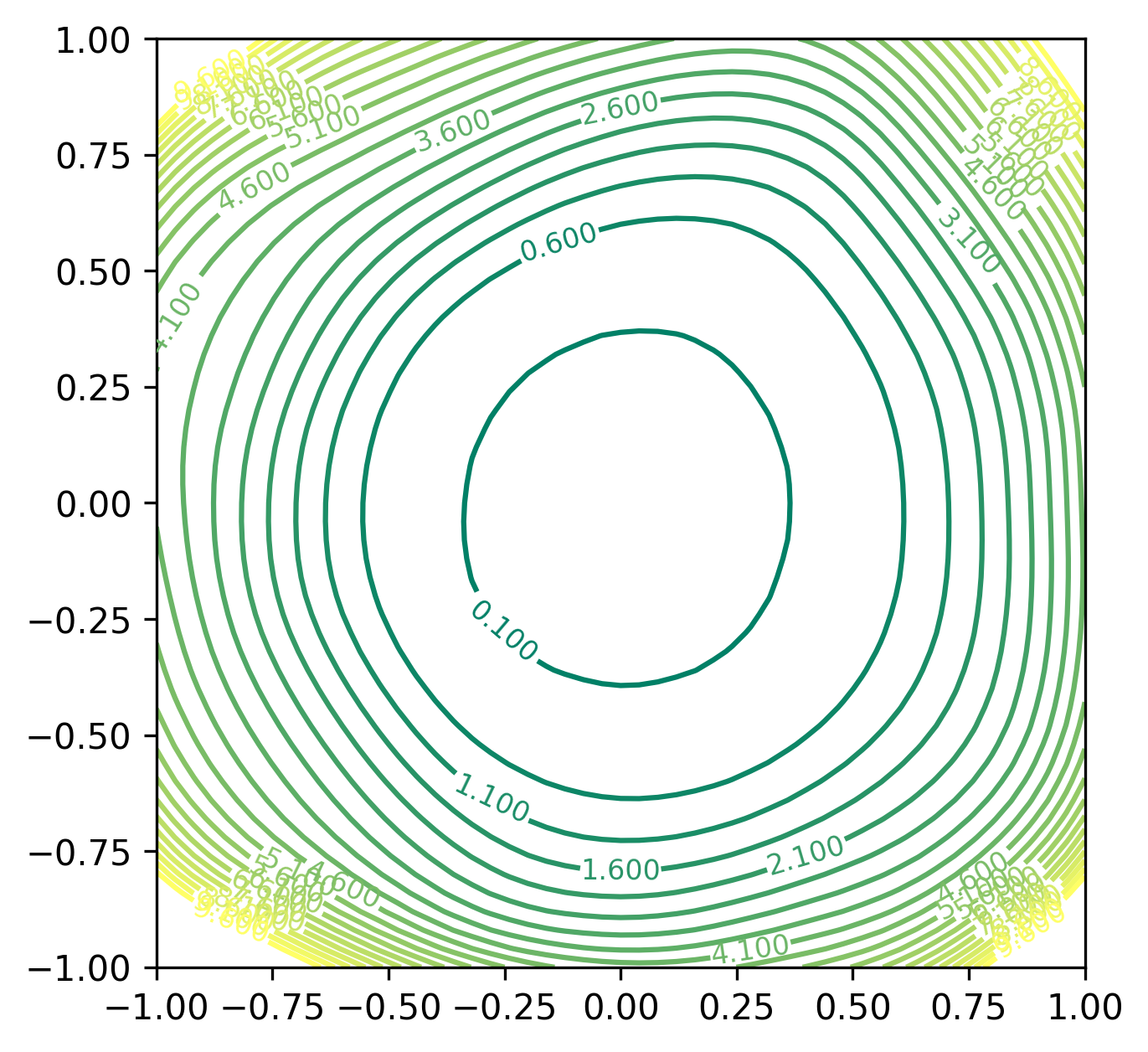}}
       \hfill
    \subfloat[800 epochs\label{cifar10 resnet18 landscape e800}]{%
       \includegraphics[width=0.16\linewidth]{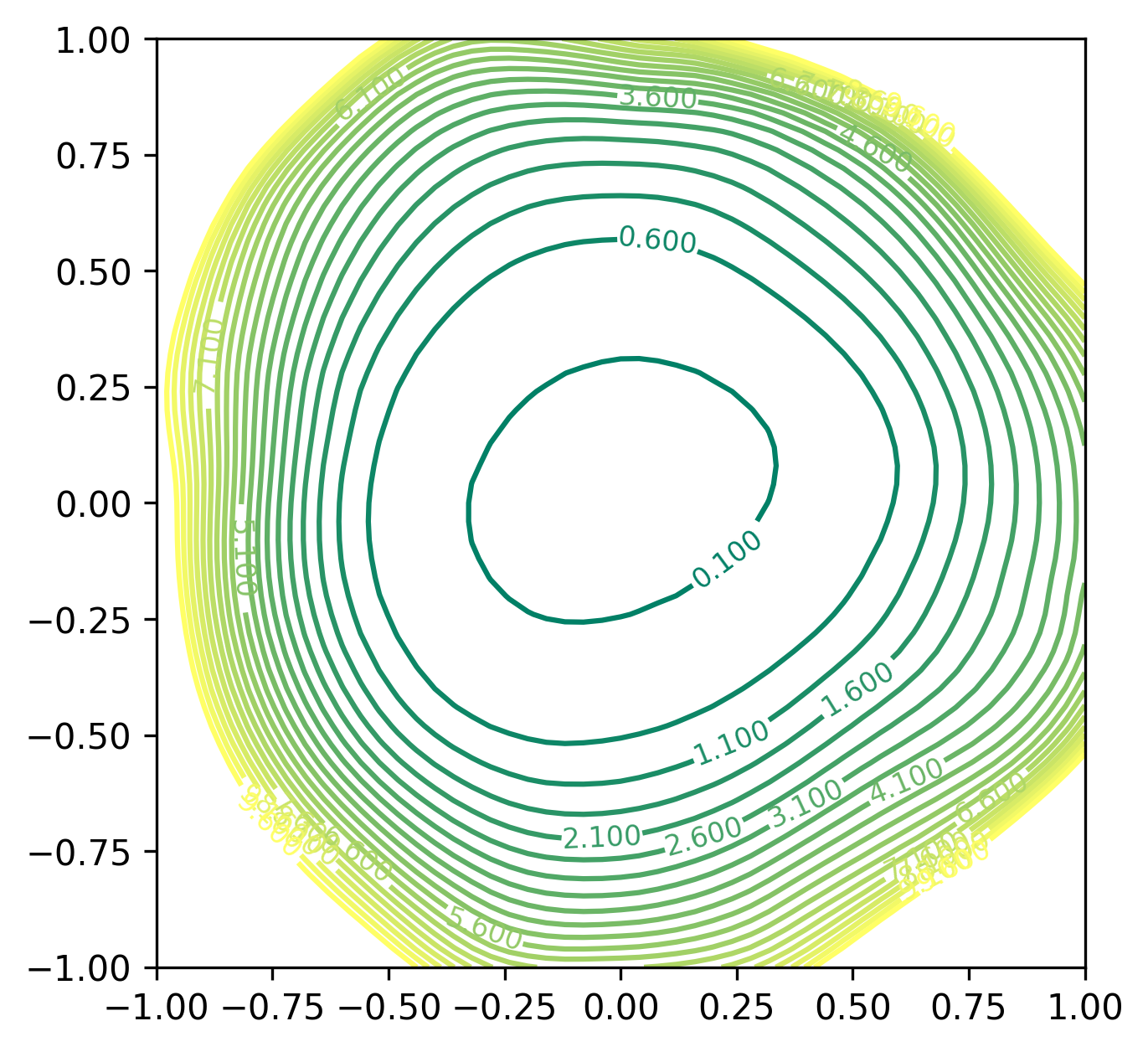}}
    \caption{Training ResNet18 on CIFAR10  {training set} (100\% data and 30\% data) without data augmentation. Top row: training loss and testing accuracy for ERM and Mixup.
    %Figures \ref{0.3 cifar10 resnet18 trainloss} to \ref{cifar10 resnet18 testacc}, 
    %the red and blue curves refer to the corresponding results of ERM training and Mixup training respectively. 
    %Figures \ref{0.3 cifar10 resnet18 trainloss} and \ref{0.3 cifar10 resnet18 testacc} respectively show how the minimal training loss and the corresponding testing accuracy change by the total number of epochs on the $30\%$ CIFAR10. We observe that over-training reduces both the ERM and the Mixup training loss, but the testing accuracy of the Mixup-trained ResNet18 gradually decreases, while that is not the case for the ERM-trained ResNet18. 
    %of the ERM-trained ResNet18 show no significant changes.
    Bottom row: 
    %Figures \ref{0.3 cifar10 resnet18 landscape e200} to \ref{0.3 cifar10 resnet18 landscape e800} show the 
    loss landscape of the Mixup-trained ResNet18 (where ``loss'' refers to the empirical risk on the real data) at various training epochs; left 3 figures are for the 30\% CIFAR10 dataset, and the right 3 are for the full CIFAR10 dataset; visualization follows~\cite{li2018visualizing}
    %with $200$, $400$ and $800$ training epochs in total. 
    %We observe that as the training epochs increases, the model tends to fall on the sharper local minima on the landscape of the training loss. Additionally, as shown in figures  \ref{cifar10 resnet18 trainloss}, \ref{cifar10 resnet18 testacc}, \ref{cifar10 resnet18 landscape e200}, \ref{cifar10 resnet18 landscape e400} and \ref{cifar10 resnet18 landscape e800}, training ResNet18 on the original CIFAR10 brings about the similar results. %The difference is that the training loss of the ERM-trained ResNet18 on the original CIFAR10 decrease with over-training, while its generalization still show no significant changes.
    }
    % \vspace{-5mm}
    \label{fig: CIFAR10 loss & acc curves}
\end{figure}

For CIFAR10 and SVHN, ResNet18 is used, and both the original training sets and subsets with 30\% of the original data are adopted. % and ResNet34 is deployed for the more challenging task CIFAR100. 
%on both the original training set and $30\%$ of it. We tune the total number of  training epochs intermittently from $200$ to $1600$. 
%Also, the local loss landscapes (where ``loss'' refers to the empirical risk on the real data) around the found solutions are visualized in 2D following~\cite{li2018visualizing}
 Training is performed for up to $1600$ epochs for CIFAR10, and the results  are shown in Figure \ref{fig: CIFAR10 loss & acc curves}. For both the 30\% dataset and the full dataset, we see clearly that after some number of epochs (e.g, epoch 200 for the full dataset), the testing accuracy of the Mixup-trained network starts decreasing and this trend continues. This confirms that over-training with Mixup hurts the network's generalization. One would observe a U-shaped curve, as shown in Figure \ref{fig: CIFAR10 loss & acc curves: introduction} (right),  if we were to plot testing error and include results from earlier epochs. Notably, this phenomenon is not observed in ERM. We also found that over-training with Mixup tends to force the network to learn a solution located at the sharper local minima on the loss landscape, a phenomenon correlated with degraded generalization performance
%The relationship between the two phenomenons is in accordance with a commonly accepted theorem that a network is generalized better if its learned solution corresponds to a flat local minima of the training loss 
\citep{hochreiter1997flat,keskar2016large}.  The results of training ResNet18 on SVHN is presented in Appendix~\ref{appendix:additional overtrain results}.

% \vspace{-4mm}

\begin{figure}[!h]
    \centering
    \subfloat[{Train loss}\label{cifar100 resnet34 trainloss}]{%
       \includegraphics[width=0.346\linewidth]{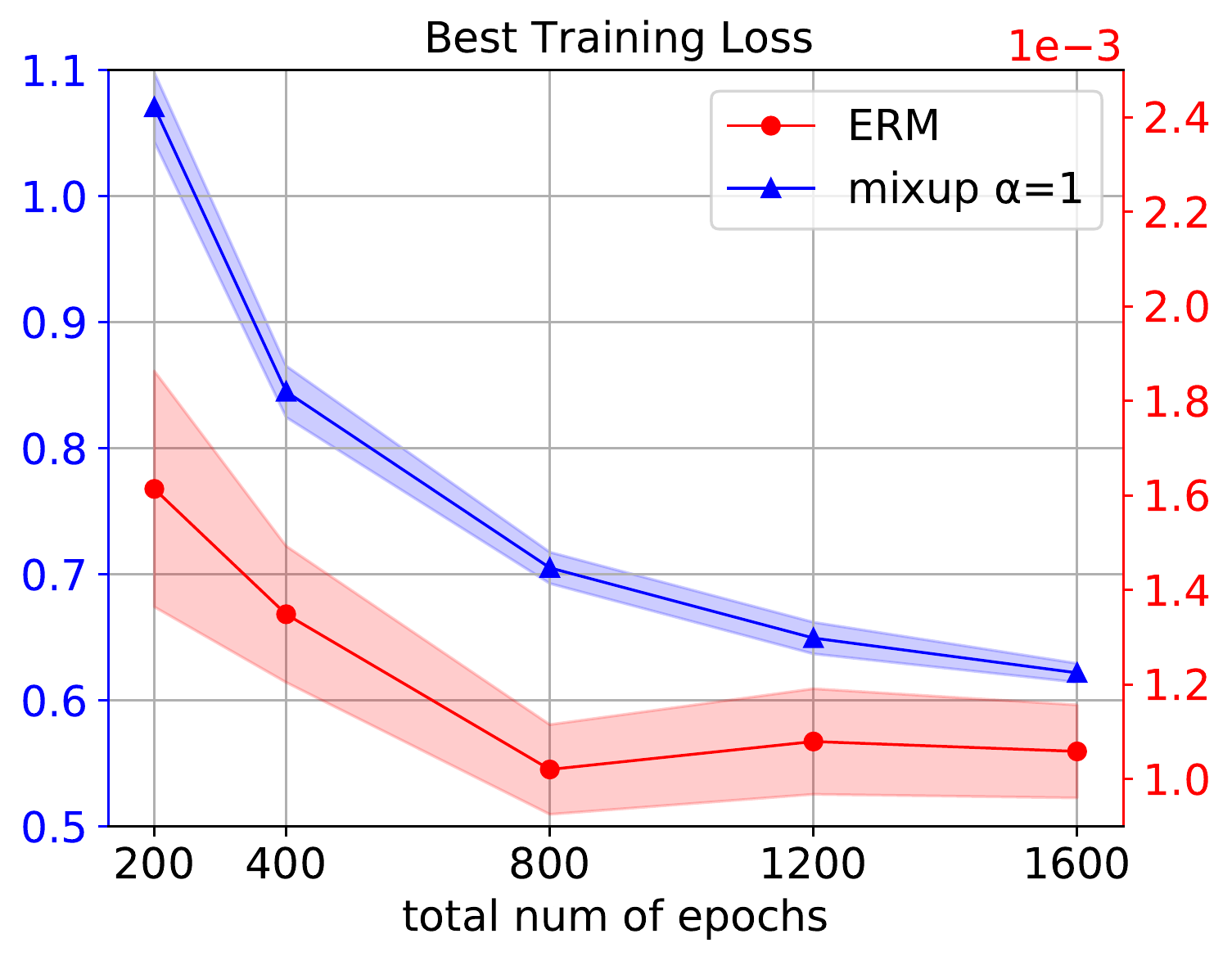}}
    \subfloat[{Test acc}\label{cifar100 resnet34 testacc}]{%
       \includegraphics[width=0.331\linewidth]{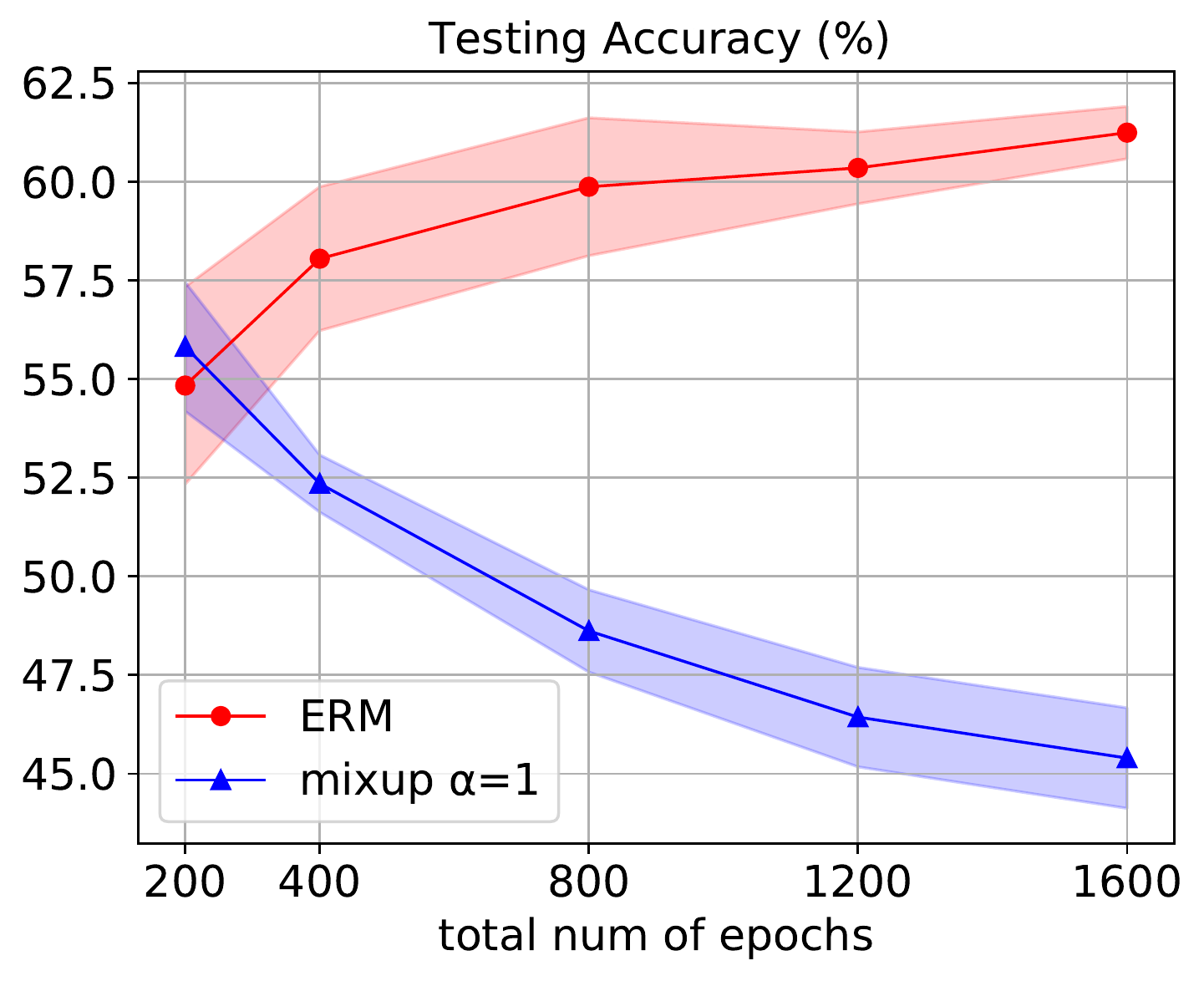}}
    \subfloat[U-shaped curve\label{subfig:cifar100 resnet34 u curve}]{%
       \includegraphics[width=0.321\linewidth]{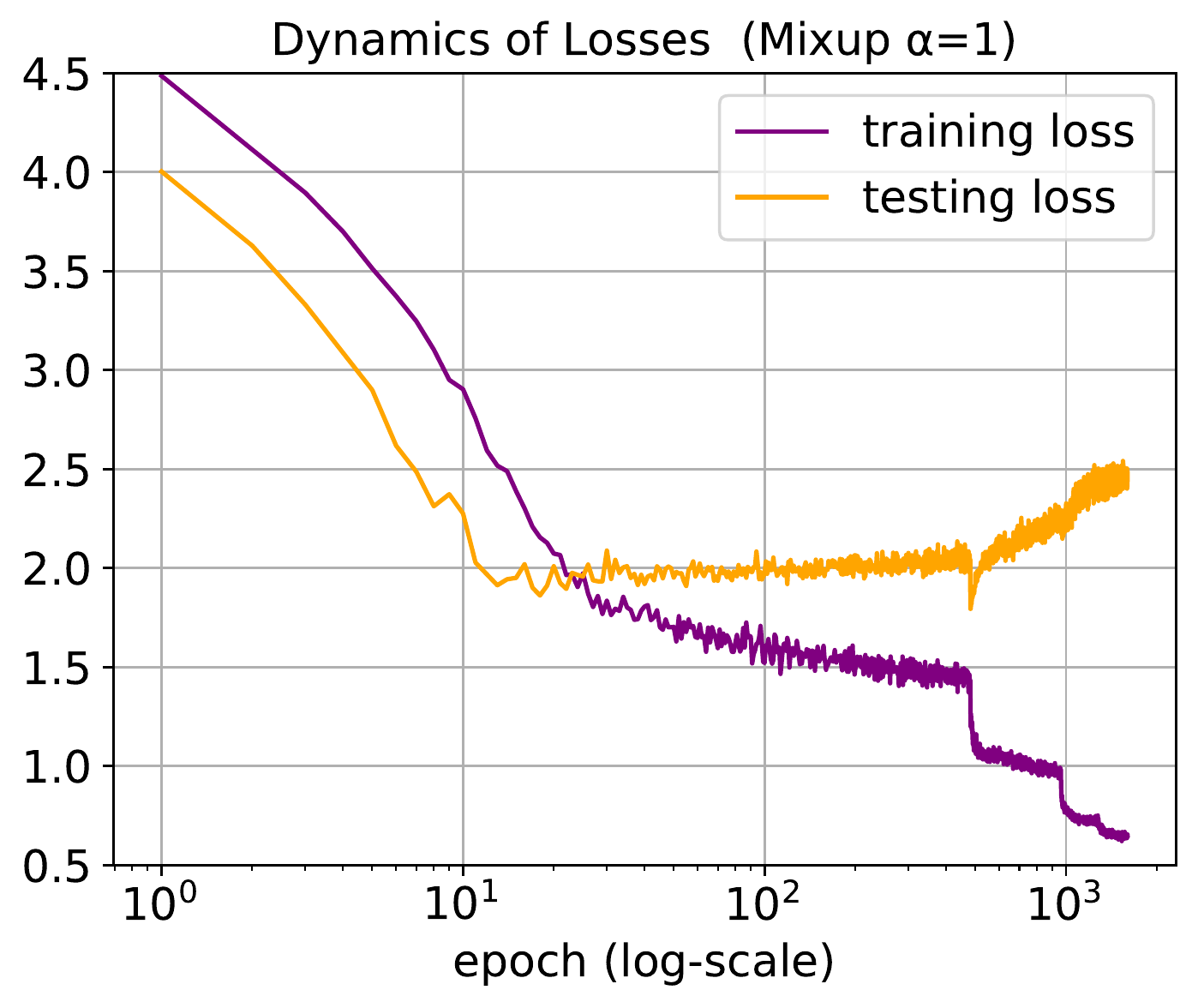}}
    \caption{Training loss, testing accuracy and a U-shaped testing loss curve (subfigure (c), yellow) of training ResNet34 on CIFAR100 (100\% training data) without data augmentation. %Figure \ref{cifar100 resnet34 u curve} show how the training loss and the testing loss change during a single trial of over-training ResNet34 on CIFAR100 with Mixup, where %. A remarkable a U-shaped curve of the testing loss is also observed.
    }
    % \vspace{-4mm}
    \label{fig: CIFAR100 loss & acc curves}
\end{figure}

ResNet34 is used for the more challenging task CIFAR100. This choice allows Mixup to drive its loss to lower values, closer to the lower bound given in Lemma~\ref{lem:mixup loss lower bound}. In this case, we only use the original training set, since downsampling CIFAR100 appears to result high variances in the testing performance.
%Unlike section \ref{subsec:results of cifar10} and \ref{subsec:results of svhn}, for CIFAR100 we only present our observations of over-training ResNet34. The reason is that the Mixup training loss of ResNet18 trained on CIFAR100 has high variance and in general cannot be reduced to be relatively close to its lower bound given by Lemma~\ref{lem:mixup loss loweround}.
Training is performed for up to $1600$ epochs. The results are plotted in Figure \ref{fig: CIFAR100 loss & acc curves}. The results again confirm that over-training with Mixup hurts the generalization capability of the learned model. 
%where the same over-fitting behavior when over-training the network with Mixup is observed. 
A U-shaped testing loss curve (obtained from a single trial) is also observed in Figure \ref{subfig:cifar100 resnet34 u curve}. Additional results of training ResNet34 on CIFAR10 and SVHN are provided in Appendix~\ref{appendix:additional overtrain results}.
%the loss curve in a single trial of over-training with a total of $1600$ epochs. This subfigure again shows that the Mixup training loss  decreases continuously, while the testing loss presents a U-shaped curve. 
%\subsection{Results for CIFAR100}

\subsection{Results on Over-Training With Data Augmentation}\label{subsec:results of cifar10 with data augmentation}
% \vspace{-2mm}

\begin{figure}[htbp]
    \centering
    \subfloat[Train loss (CIFAR10) \label{subfig: 0.1 cifar10 resnet18 aug loss curves}]{%
       \includegraphics[width=0.256\linewidth]{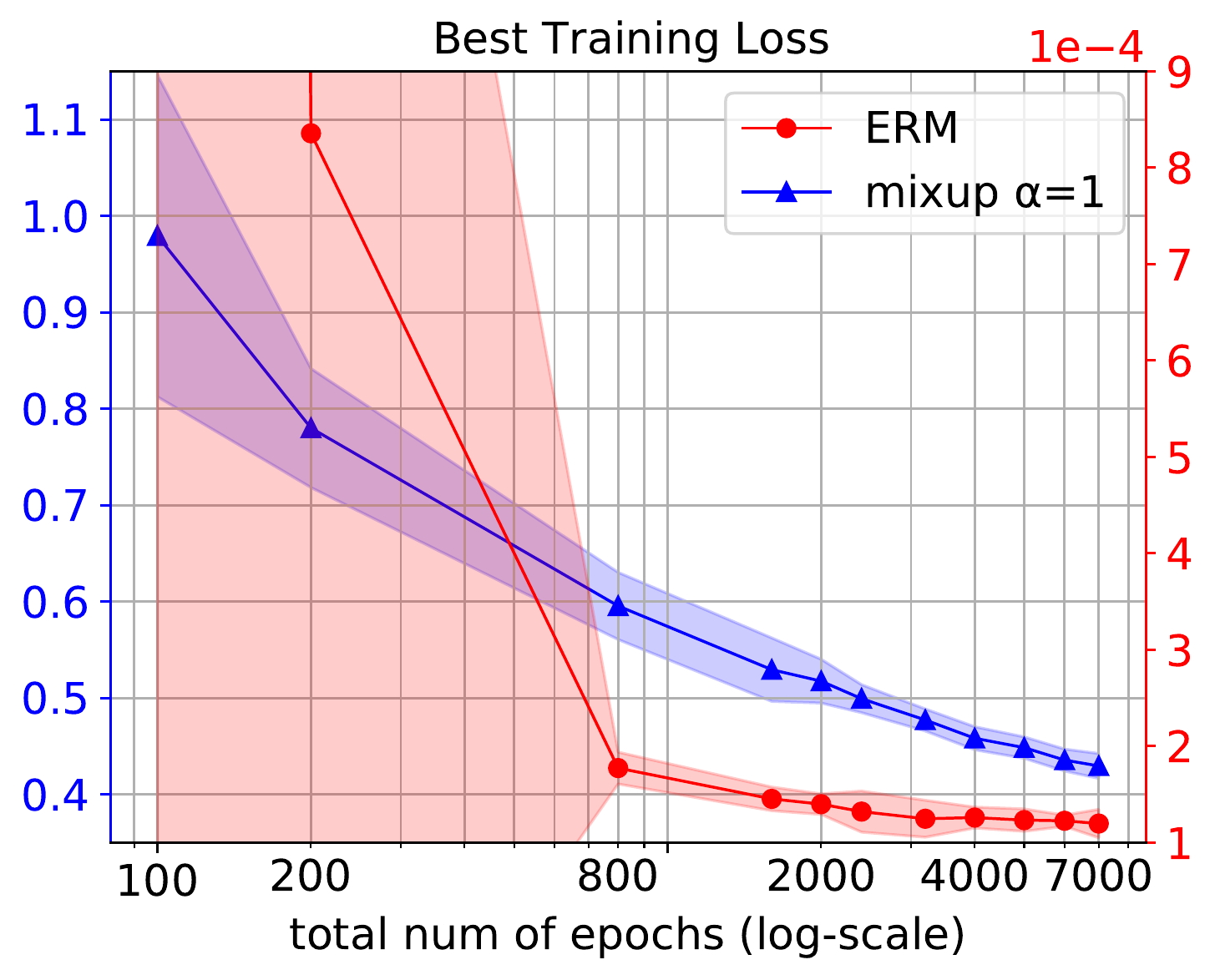}}
       % \hspace{14mm}
    \subfloat[Test error (CIFAR10) \label{subfig: 0.1 cifar10 resnet18 aug acc curves}]{%
       \includegraphics[width=0.244\linewidth]{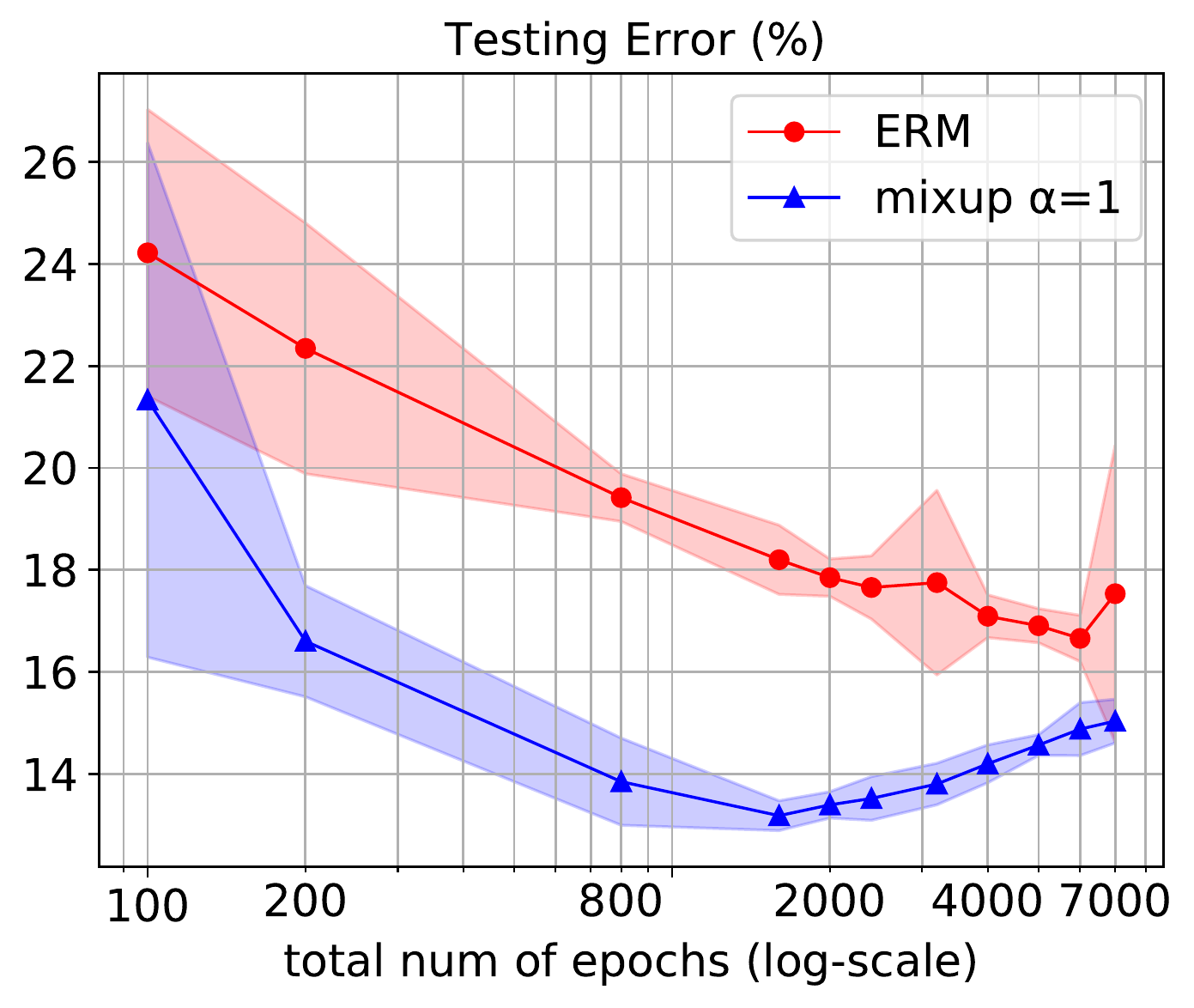}}
    \subfloat[Train loss (CIFAR100) \label{subfig: 0.1 cifar100 resnet34 aug loss curves}]{%
       \includegraphics[width=0.256\linewidth]{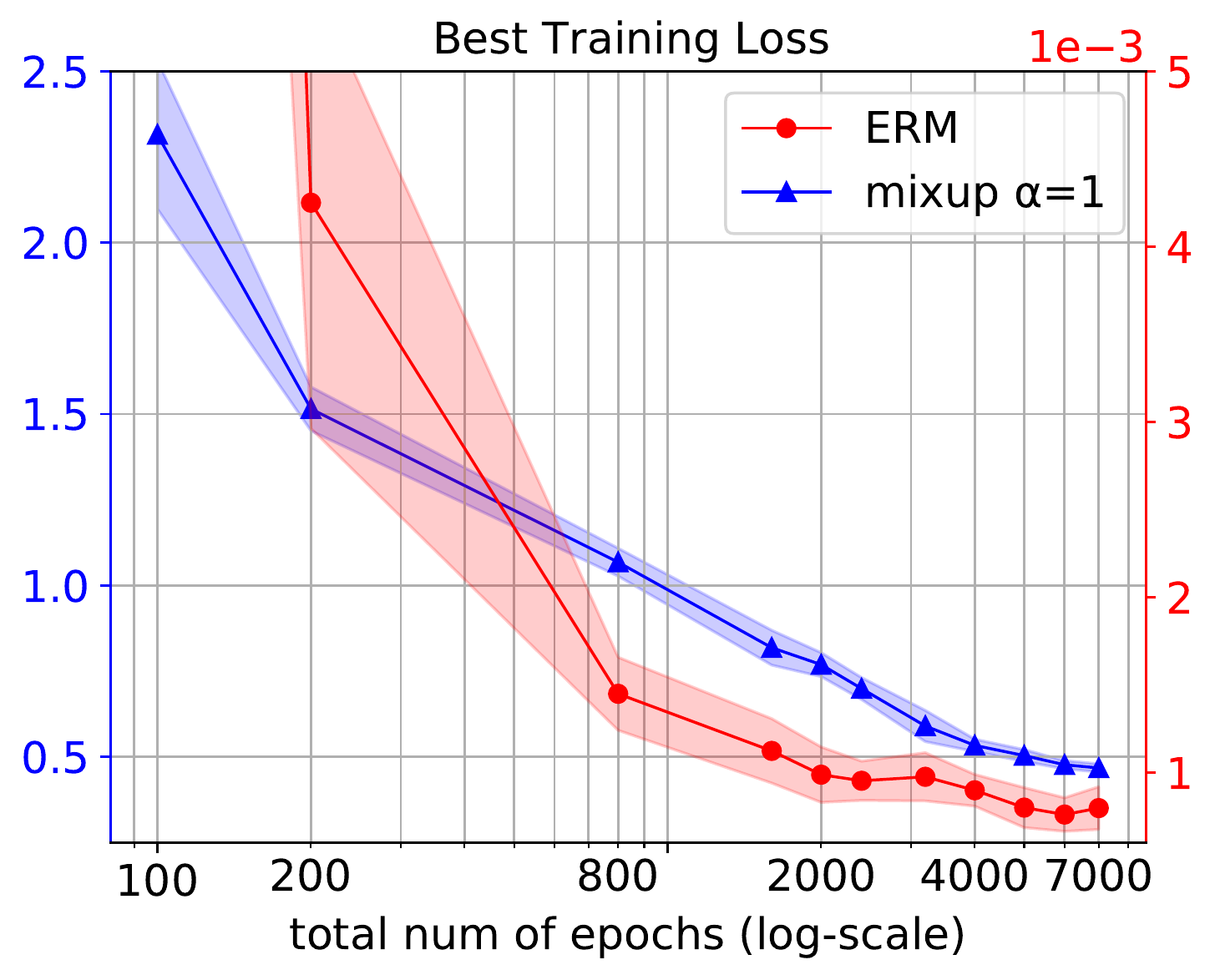}}
       % \hspace{14mm}
    \subfloat[Test error (CIFAR100) \label{subfig: 0.1 cifar100 resnet34 aug acc curves}]{%
       \includegraphics[width=0.244\linewidth]{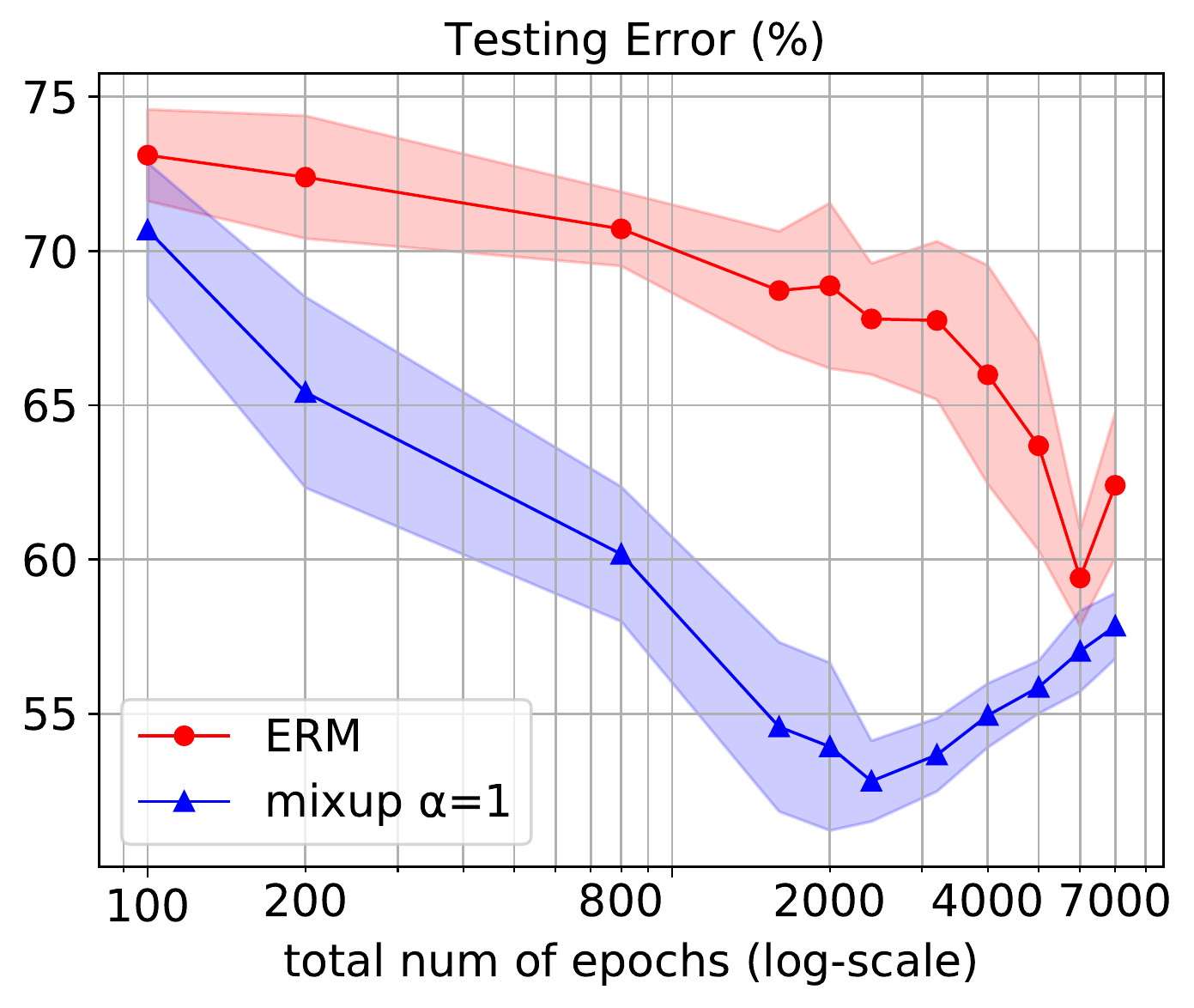}}
    \caption{(a),(b): Training losses and testing errors of over-training ResNet18 on 10\% of the CIFAR10 training set with data augmentation. (c),(d): Training losses and testing errors of over-training ResNet34 on 10\% of the CIFAR100 training set with data augmentation.}
    \label{fig: 0.1 aug loss & acc curves}
\end{figure}

%We have also carried out the over-training experiments with data augmentation applied on CIFAR10 and CIFAR100. 
The data augmentation methods include ``random crop'' and ``horizontal flip'' are applied to training on CIFAR10 and CIFAR100. 
We train ResNet18 on 10\% of the CIFAR10 training set 
for up to $7000$ epochs. The results are given in Figures~\ref{subfig: 0.1 cifar10 resnet18 aug loss curves} and \ref{subfig: 0.1 cifar10 resnet18 aug acc curves}. In this case, the Mixup-trained model also produces a U-shaped generalization curve. However, while the dataset is downsampled to a lower proportion, the turning point of the U-shaped curve nevertheless comes much later compared to the previous experiments where data augmentation is not applied on CIFAR10.

\vspace{-2mm}

% \vspace{-4mm}

The results of over-training ResNet34 on 10\% of the CIFAR100 training set for up to $7000$ epochs are given in Figures~\ref{subfig: 0.1 cifar100 resnet34 aug loss curves} and \ref{subfig: 0.1 cifar100 resnet34 aug acc curves}, where similar phenomenons  are observed.
% \vspace{-1mm}
% \begin{figure}[htbp]
%     \centering
%     \subfloat[Training Loss ($10\%$ data)]{%
%        \includegraphics[width=0.35\linewidth]{iclr2023/figures/CIFAR100_ResNet34_CEloss_aug_alpha_log_trainloss_p0.1_fill.pdf}}
%        \hspace{14mm}
%     \subfloat[Testing Error ($10\%$ data)]{%
%        \includegraphics[width=0.333\linewidth]{iclr2023/figures/CIFAR100_ResNet34_CEloss_aug_alpha_log_testerr_p0.1_fill.pdf}}
%     \caption{\textcolor{red}{Results of the recorded training losses and testing errors of over-training ResNet34 on 10\% of the CIFAR100 training set with data augmentation.}}
%     \label{fig: 0.1 cifar100 resnet34 aug loss & acc curves}
% \end{figure}
% \vspace{-3mm}

%\vspace{-1mm}
\section{Theoretical Explanation}
\label{analysisresult}
%\vspace{-1mm}
% In this section, we first demonstrate Mixup may induce undesired label noises. We then analyze a special case to illustrate why label noise will render the U-shaped %behaviour in its generalization dynamics.
% generalization curve.

%\vspace{-1mm}
\subsection{Mixup Induces Label Noise}
% \vspace{-2mm}
\label{mixupintroducenoise}
%Recall that the input feature space $\mathcal{X}\subset\mathbb{R}^{d_0}$. 
%We now let $\mathbf{y}$ be the hard label instead of one hot vector then the label space $\mathcal{Y}=\{1,2,\dots,C\}$. 

We will use the capital letters $X$ and $Y$ 
to denote the random variables representing the input feature and output label, while reserving the notations $\textbf{x}$ and $\textbf{y}$ to denote their respective realizations. In particular, we consider each true label $\textbf{y}$ is as a token, in ${\cal Y}$, not a one-hot vector in ${\cal P}({\cal Y})$.
Let $P(Y|X)$ be the ground-truth conditional distribution of the label $Y$ given input feature $X$.  For simplicity, we also express $P(Y|X)$ as a vector-valued function $f:\mathcal{X}\rightarrow \mathbb{R}^C$, where $f_j(\textbf{x})\triangleq P(Y=j|X=\textbf{x})$ for each dimension $j\in\mathcal{Y}$.
%and input $\textbf{x}$.
%Under this ground truth, the correct hard-assignment of label for $\textbf{x}$ is $\arg\max_{j\in\mathcal{Y}} f_j(\textbf{x})$. 

For simplicity, we consider Mixup with a fixed $\lambda\in[0,1]$; extension to random $\lambda$ is straight-forward. 
Let $\widetilde{X}$ and $\widetilde{Y}$ be the random variables 
corresponding to the synthetic feature and synthetic label respectively. Then $\widetilde{X}\triangleq \lambda X +  (1-\lambda) X'$.  Let $P(\widetilde{Y}|\widetilde{X})$ be the 
conditional distribution of the synthetic label conditioned on the synthetic feature, induced by Mixup, namely, $P(\widetilde{Y}=j|\widetilde{X})=\lambda f_j(X) + (1-\lambda) f_j(X')$ for each $j$. 
Then for a synthetic feature
$\widetilde{X}$, there are two ways to assign to it a hard label. The first is based on the ground truth, assigning
$\widetilde{Y}^*_{\rm h}\triangleq \arg\max_{j\in\mathcal{Y}}f_j(\widetilde{X})$. The second is based on the Mixup-induced conditional $P(\widetilde{Y}|\widetilde{X})$, assigning $\widetilde{Y}_{\rm h}\triangleq \arg\max_{j\in\mathcal{Y}}P(\widetilde{Y}=j|\widetilde{X})$.
When the two assignments disagree, or
$\widetilde{Y}_{\rm h}\neq \widetilde{Y}_{\rm h}^*$, we say that the Mixup-assigned label $\widetilde{Y}_{\rm h}$ is noisy.

%$\widetilde{X}$, namely $\widetilde{Y}_{\rm h}$,  is noisy. The following theorem gives a lower bound of such noise.

\begin{thm}
\label{thm: noise-lower-bound}
For any fixed $X$, $X'$ and $\widetilde{X}$ related by $\widetilde{X}= \lambda X + (1-\lambda) X'$ for a fixed $\lambda\in[0,1]$, the probability of assigning a noisy label  is lower bounded by
\[
P(\widetilde{Y}_{\rm h}\neq \widetilde{Y}_{\rm h}^*|\widetilde{X})\geq \mathrm{TV}(P(\widetilde{Y}|\widetilde{X}), P(Y|X))\geq\frac{1}{2}\sup_{j\in\mathcal{Y}}\left|f_{j}(\widetilde{X})-\left[(1-\lambda) f_{j}(X) + \lambda f_{j}(X')\right]\right|,
\]
where $\mathrm{TV}(\cdot,\cdot)$ is the total variation (see Appendix~\ref{sec:proofs}). 
% Further, if there exists some $\tilde{j}\in\mathcal{Y}$ such that $f_{\tilde{j}}$ is strongly convex with parameter $\rho>0$, then
% \[
% P(\widetilde{Y}\neq \widetilde{Y}^*|\widetilde{X})\geq \frac{\rho}{2}\lambda(1-\lambda)||X-X'||_2^2.
% \]
\end{thm}
\begin{rem}
This lower bound hints that the label noise induced by Mixup training depends on the distribution of original data $P_X$, the convexity of $f(X)$ and the value of $\lambda$. Clearly, Mixup will create noisy labels with non-zero probability (at least for some $\lambda$) unless $f_j$ is linear for each $j$.
\end{rem}
\begin{rem}
We often consider that the real data are labelled with certainty, i.e.,  $\max_{j\in\mathcal{Y}} f_j(X) = 1$ and $\sum_{j=1}^C f_j(X) = 1$. Then the probability of assigning noisy label to a given synthetic data can be discussed in three situations: i) if $\widetilde{Y}_{\rm h}^*\notin\{Y,Y'\}$,  where $Y$ could be the same with $Y'$, then $\widetilde{Y}$ is a noisy label with probability one; ii) if $\widetilde{Y}_{\rm h}^*\in\{Y,Y'\}$ where $Y\neq Y'$, then the probability of assigning a noisy label is non-zero and depends on $\lambda$;
% is assigned with probability at least $\lambda$ or $1-\lambda$; 
iii) if $\widetilde{Y}_{\rm h}^*=Y=Y'$, then $\widetilde{Y}_{\rm h}^*=\widetilde{Y}$.
\end{rem}

As shown in  \citep{arpit2017closer,arora2019fine}, when neural networks are trained with a fraction of random labels, they will first learn the clean pattern and then overfit to noisy labels. In Mixup training, we in fact create much more data, possibly with noisy labels, than traditional ERM training ($n^2$ for a fixed $\lambda$). Thus, %\textcolor{blue}{
one may expect an improved performance (relative to ERM) in the early training phase, due to the clean pattern in the enlarged training set, 
but a performance impairment in the later phase due to noisy labels.
%Mixup training %more clean data
%will give higher testing performance in the first stage of learning (where neural networks learn the true pattern of data \citep{arpit2017closer}), and then \textcolor{red}{the} performance will be impaired due to overfitting to noisy data. 
Specifically if $\widetilde{Y}^*_{\rm h}\notin\{Y,Y'\}$ happens with a high chance,  a phenomenon known as ``manifold intrusion'' \citep{guo2019mixup}, then the synthetic dataset contains too many noisy labels, causing Mixup to perform inferior to ERM.

Theorem 5.1 has implied that, in classification problems, Mixup training induces label noise. Next, we will provide a theoretical analysis using a regression setup to explain that such label noise may result in the U-shape learning curve. The choice of a regression setup in this analysis is due to the difficulty in directly analyzing classification problems (under the cross-entropy loss).  Such a regression setting may not perfectly explain the U-shaped curve in classification tasks, we however believe that they give adequate insight illuminating such scenarios as well. 
Such an approach has been taken in most analytic works that study the behaviour of deep learning. 
%Indeed, due to the difficulty of analyzing the cross-entropy loss, most of the analytic works for deep learning study regression problems under the square-error loss and use insights obtain this way to explain the behaviour of deep neural net in classification settings. 
For example,  ~\cite{DBLP:conf/icml/AroraDHLW19} uses a regression setup to analyze the optimization and generalization property of overparameterized neural networks. ~\cite{DBLP:conf/icml/YangYYSM20} theoretically analyze the bias-variance trade-off in  deep network generalization using a regression problem.

% \vspace{-2mm}
\subsection{Regression Setting With Random Feature Models}
% \vspace{-2mm}
\label{theoryRegression}

Consider a simple least squares regression problem. Let $\mathcal{Y}=\mathbb{R}$ and let $f:\mathcal{X}\rightarrow \mathcal{Y}$ be the ground-truth labelling function. 
Let $(\widetilde{X}, \widetilde{Y})$ be a synthetic pair obtained by mixing $(X, Y)$
and $(X', Y')$. Let $\widetilde{Y}^*=f(\widetilde{X})$ and $Z\triangleq \widetilde{Y}- \widetilde{Y}^* $. Then $Z$ can be regarded as noise introduced by Mixup, which may be data-dependent. 
For example, if $f$ is strongly convex with some parameter $\rho>0$, then $Z\geq \frac{\rho}{2}\lambda(1-\lambda)||X-X'||_2^2$. Given a synthesized training dataset $\widetilde{S}=\{(\widetilde{X}_i,\widetilde{Y}_i)\}_{i=1}^m$, consider a random feature model, $\theta^T\phi(X)$, where $\phi:\mathcal{X}\rightarrow\mathbb{R}^d$ and $\theta\in \mathbb{R}^d$. We will consider $\phi$ fixed and only learn the model parameter $\theta$ using gradient descent on the MSE loss
%Notice that although $\phi$ may have some fixed parameters, we only update $\theta$ during training. We use MSE as the loss, so the empirical loss is
\[
\hat{R}_{\widetilde{S}}(\theta)\triangleq\frac{1}{2m}\left|\left|\theta^T\widetilde{\Phi}-\widetilde{\mathbf{Y}}^T\right|\right|_2^2,
\]
where $\widetilde{\Phi}=[\phi(\widetilde{X}_1), \phi(\widetilde{X}_2), \dots,\phi(\widetilde{X}_m)]\in\mathbb{R}^{d\times m}$ and $\widetilde{\mathbf{Y}}=[\widetilde{Y}_1,\widetilde{Y}_2, \dots, \widetilde{Y}_m]\in\mathbb{R}^m$. 
%Notably, we will consider $\phi$ as fixed and only update $\theta$.

For a fixed $\lambda$, Mixup can create $m=n^2$ synthesized examples. Thus it is reasonable to assume $m>d$ (e.g., under-parameterized regime) in Mixup training. For example, ResNet-50 has less than $30$ million parameters while the square of CIFAR10 training dataset size is larger than $200$ million without using other data augmentation techniques. Then the gradient flow, as shown in \cite{liao2018dynamics}, is 
\begin{align}
    \label{eq:gradient-flow}
    \dot{\theta}=-\eta\nabla\hat{R}_{\widetilde{S}}(\theta)=\frac{\eta}{m}\widetilde{\Phi}\widetilde{\Phi}^T\left(\widetilde{\Phi}^{\dagger}\widetilde{\mathbf{Y}}-\theta\right),
\end{align}
where $\eta$ is learning rate and $\widetilde{\Phi}^{\dagger}=(\widetilde{\Phi}\widetilde{\Phi}^T)^{-1}\widetilde{\Phi}$ is the Moore–Penrose inverse of $\widetilde{\Phi}^T$ (only possible when $m>d$).
% Further, the ordinary differential equation above (Newton's law of cooling) has the following closed form solution
% \begin{align}
% \label{eq:ode-solution}
%     \theta_t = \widetilde{\Phi}^{\dagger}\widetilde{\mathbf{Y}} + (\theta_0-\widetilde{\Phi}^{\dagger}\widetilde{\mathbf{Y}})e^{-\frac{\eta}{m}\widetilde{\Phi}\widetilde{\Phi}^T t}.
% \end{align}
Thus, we have the following important lemma.

\begin{lem}
\label{lem:model-dynamic}
Let $\theta^*=\widetilde{\Phi}^{\dagger}\widetilde{\mathbf{Y}}^*$ and $\theta^{\mathrm{noise}}=\widetilde{\Phi}^{\dagger}\mathbf{Z}$ wherein $\mathbf{Z}=[Z_1,Z_2, \dots, Z_m]\in\mathbb{R}^m$, 
% Recall that $\widetilde{\mathbf{Y}}^*=\widetilde{\mathbf{Y}}+\mathbf{Z}$ wherein $\mathbf{Z}=[Z_1,Z_2, \dots, Z_m]\in\mathbb{R}^m$, 
the ODE in Eq.~(\ref{eq:gradient-flow}) has the following closed form solution
\begin{align}
\label{eq:ode-solution-2}
    \theta_t-\theta^* =   (\theta_0-\theta^*)e^{-\frac{\eta}{m}\widetilde{\Phi}\widetilde{\Phi}^T t} + (\mathbf{I}_d - e^{-\frac{\eta}{m}\widetilde{\Phi}\widetilde{\Phi}^T t})\theta^{\mathrm{noise}}.
\end{align}
\end{lem}
\begin{rem}
%Lemma~\ref{lem:model-dynamic} indicates that the dynamics of $\theta$ gives a U-shaped curve in each dimension, and the increasing behavior results from the second term that contains the noise $\mathbf{Z}$. More precisely, 
Notably, $\theta^*=\widetilde{\Phi}^{\dagger}\widetilde{\mathbf{Y}}^*$ may be seen as the ``clean pattern'' of the training data.
The first term in Eq.~(\ref{eq:ode-solution-2}) is  decreasing (in norm) 
and vanishes at $t\rightarrow \infty$. Thus its role is moving $\theta_t$ towards the clean pattern $\theta^*$, allowing the model to generalize to unseen data. But it only dominates the dynamics of $\theta_t$ in the early training phase.  The second term, initially 0,  increases with $t$ and converges to $\theta^{\rm noise}$ as $t\rightarrow \infty$. Thus its role is moving $\theta_t$ towards the ``noisy pattern" $\theta^*+\theta^{\mathrm{noise}}$. It dominates the later training phase and hence hurts generalization.
%Remarkably $\theta^*=\widetilde{\Phi}^{\dagger}\widetilde{\mathbf{Y}}^*$ may be understood as the ``clean pattern'' of the training data. 
%Then we see that the model, in the early phase, is learning the ``clean pattern'',  which generalizes to the unseen data (i.e. $(X,Y)$). 
%In the later training phase, the second term in  Eq.~(\ref{eq:ode-solution-2}) gradually dominates the trajectory of $\theta_t$, and the model learns the``noisy pattern'', namely converging to $\theta^*+\theta^{\mathrm{noise}}$. This then hurts generalization. 
It is noteworthy that  $\theta^*+\theta^{\mathrm{noise}}$ is also the closed-form solution for the regression problem (under Mixup labels). This suggests that the optimization problem associated with the Mixup loss has a ``wrong'' solution, but it is possible to  benefit from only solving this problem partially, using gradient descent without over-training.  
\end{rem}
\vspace{-2mm}
Noting that the population risk at time step $t$ is 
\[
R_t \triangleq \mathbb{E}_{\theta_t,X,Y}\left|\left|\theta_t^T\phi(X)-Y\right|\right|_2^2,
\]
and the true optimal risk is 
\[R^*=\mathbb{E}_{X,Y}\left|\left|Y-\theta^{*T}\phi(X)\right|\right|_2^2,\]
we have the following result. 

% \[
% R_t \triangleq \mathbb{E}_{\theta_t,X,Y}\left|\left|\theta_t^T\phi(X)-Y\right|\right|_2^2\leq 2\mathbb{E}_{\theta_t,X,Y}\left|\left|\theta_t^T\phi(X)-\theta^{*T}\phi(X)\right|\right|_2^2 + R^*,
% \]
% where $R^*=2\mathbb{E}_{X,Y}\left|\left|Y-\theta^{*T}\phi(X)\right|\right|_2^2$.

% As suggested in Theorem~(\ref{thm: noise-lower-bound}), the probability of $\widetilde{Y}\neq f(\widetilde{X})$ depends on the data distribution $P_X$ and the value of $\lambda$.

%The following theorem shows the dynamics of the population risk under mild assumptions.
\begin{thm}[Dynamics of Population Risk]
\label{thm:mixup-dynamic}
Given a synthesized dataset $\widetilde{S}$, assume $\theta_0\sim\mathcal{N}(0,\xi^2 \mathrm{I}_d)$,
% $||\phi(X)||_2$ has zero mean 
$||\phi(X)||^2\leq {C_1/2}$ for some constant $C_1>0$ 
and $|Z|\leq \sqrt{C_2}$ for some constant $C_2>0$, then we have the following upper bound
\[
R_t - R^*\leq  C_1\sum_{k=1}^d
{\Biggl [\Biggr.}
\left(\xi^2_k+\theta^{*2}_k\right)e^{-2\eta {\mu}_k t} +\frac{C_2}{\mu_k}\left(1 - e^{-{\eta}{\mu}_k t}\right)^2
{\Biggl.\Biggr]}
+ 2\sqrt{C_1R^*\zeta},
\]
where  $\zeta=\sum_{k=1}^d\max\{\xi^2_{k}+\theta_k^{*2},\frac{C_2}{\mu_k}\}$ and $\mu_k$ is the $k^{\rm th}$ eigenvalue of the matrix $\frac{1}{m}\widetilde{\Phi}\widetilde{\Phi}^T$. 
% where $\mathrm{Var}$ denotes the variance.
\end{thm}
\begin{rem}
The additive noise $Z$ is usually assumed as a zero mean Gaussian in the literature of generalization dynamics analysis \citep{advani2020high,pezeshki2022multi,heckel2021early}, but this would be hardly justifiable in this context. The boundness assumption of $Z$  in the theorem can however be easily satisfied as long as the output of $f$ is bounded.
\end{rem}

\begin{rem}
If we further let $\xi=0$ (i.e. using zero initialization) and assume that the eigenvalues of the matrix $\frac{1}{m}\widetilde{\Phi}\widetilde{\Phi}^T$ are all equal to $\mu$, then the summation part in the bound above can be re-written as
$
C_1\left|\left|\theta^{*}\right|\right|^2e^{-2\eta {\mu} t}+(C_2/\mu)\left(1 - e^{-{\eta}{\mu} t}\right)^2
$, then it is clear that the magnitude of the curve is controlled by the norm of $\theta^{*}$, the norm of the representation, the noise level and $\mu$.
\end{rem}

% \begin{wrapfigure}{r}{0.34\textwidth} %this figure will be at the right
% % \vspace{-10pt}
%     \vspace{-6mm}
%     \centering
%     \includegraphics[scale=0.3]{iclr2023/figures/toy.pdf}
%     \vspace{-4mm}
%     \caption{Dynamics of testing loss in the toy example. }
% \label{fig:toy-simulation}
% % \vspace{-20pt}
%     \vspace{-5mm}
% \end{wrapfigure}

Theorem~\ref{thm:mixup-dynamic} indicates that the population risk will first
% in each dimension will first %convexly 
decrease due to the first term (i.e.
$\left(\xi^2_k+\theta^{*2}_k\right)e^{-2\eta {\mu}_k t}$)
% $e^{-2\eta {\mu}_k t}$)
then it will
% concavely 
grow due to the 
% second term (i.e. $\left(1 - e^{-{\eta}{\mu}_k t}\right)^2$)
existence of label noises (i.e. $\frac{C_2}{\mu_k}\left(1 - e^{-{\eta}{\mu}_k t}\right)^2$)
. Overall, the population risk will be endowed with a U-shaped curve. Notice that the quantity $\eta\mu_k$ plays a key role in the upper bound, the larger id $\eta\mu_k$, the earlier comes the turning point of ``U''.
This may have an interesting application, justifying a multi-stage training strategy where the learning rate is reduced at each new stage. Suppose that with the initial learning rate, at epoch $T$, the test error has dropped to the bottom of the U-curve corresponding to this learning rate. If the learning rate is decreased at this point, then the U-curve corresponding to the new learning rate may have a lower minimum error and its bottom shifted to the right. In this case, the new learning rate allows the testing error to move to the new U-curve and further decay.

%\textcolor{red}{In particular, this explains why reducing the learning rate during training may let the population risk again decrease in a certain interval. Specifically, let the decreasing phase of risk be the stage one and let the increasing phase of risk be stage two. At the beginning of training, $\eta$ is large so the risk quickly goes to the second stage and begins to increase. If $\eta$ is reduced during this period, the risk can be decreased again. This is because smaller $\eta$ forces the risk to jump from the second stage of the original U-shaped curve to the first stage of a new U-shaped curve with a latter inflection point.}

% \vspace{-1mm}
\section{Empirical Verification}
% \vspace{-1mm}
%In this section, we present empirical evidences to validate our theoretical results in Section~\ref{analysisresult}. 

% \vspace{-2mm}

\subsection{A Teacher-Student Toy Setup}

\begin{wrapfigure}{r}{0.335\textwidth} %this figure will be at the right
% \vspace{-10pt}
    \vspace{-5.5mm}
    \centering
    \includegraphics[scale=0.3]{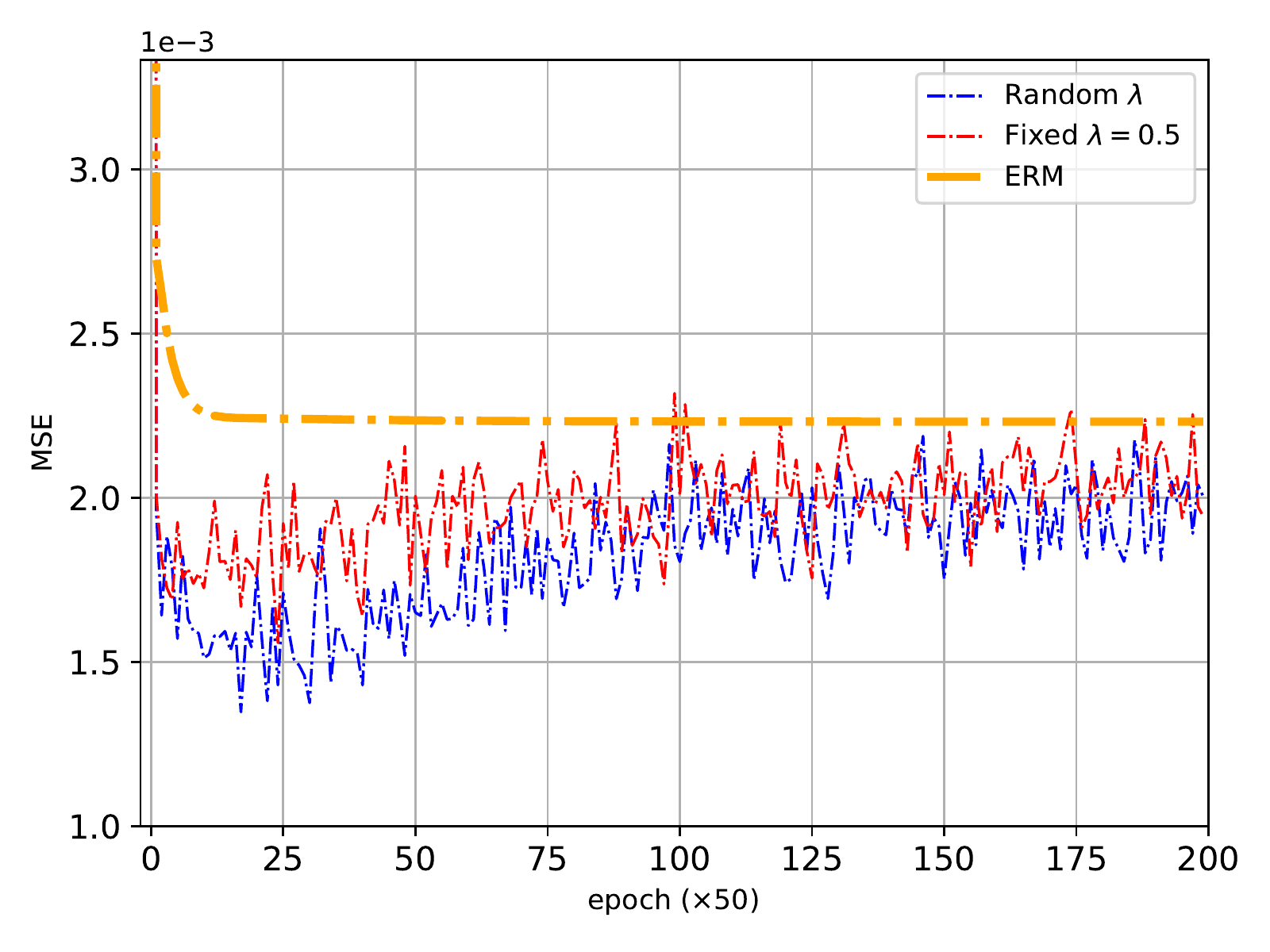}
    \vspace{-8.5mm}
    \caption{Dynamics of testing loss in the toy example. }
\label{fig:toy-simulation}
% \vspace{-20pt}
    \vspace{-5mm}
\end{wrapfigure}

To empirically verify our  theoretical results discussed in Section~\ref{theoryRegression}, we construct a simple teacher-student regression setting. The teacher network is a two-layer neural networks with \textbf{Tanh} activation and random weights. It only serves to create training data for the student network. Specifically, the training data is created by drawing $\{X_i\}_{i=1}^n$ i.i.d. from a standard Gaussian $\mathcal{N}(0,\mathrm{I}_{d_0})$ and passing them to teacher network to obtain labels $\{Y_i\}_{i=1}^n$.
%Consider the original data $\{X_i\}_{i=1}^n$ are drawn i.i.d. from a standard Gaussian $\mathcal{N}(0,\mathrm{I}_{d_0})$. 

 The student network is also a two-layer neural network with \textbf{Tanh} activation and hidden layer dimension $d=100$. We fix the parameters in the first layer and only train the second layer using the generated training data. %The output dimension of the first layer is $100$ (i.e. $d=100$).  
 Full-batch gradient descent on the MSE loss is used. For the value of $\lambda$, we consider two cases: a fixed value with $\lambda=0.5$ and  random values drawn from ${\rm Beta}(1, 1)$  at each epoch. As a comparison, we also present the result of ERM training in an over-parameterized regime (i.e., $n<d$).
% All the testing loss dynamics are presented in Figure~\ref{fig:toy-simulation}.

The testing loss dynamics are presented in Figure~\ref{fig:toy-simulation}. %From Figure~\ref{fig:toy-simulation}, 
We first note  that Mixup still outperforms ERM in this regression problem, but clearly, only Mixup training has a U-shaped curve while the testing loss of ERM training  converges to a constant value. Furthermore, the testing loss of Mixup training is endowed with a U-shaped behavior for both fixed $\lambda=0.5$ and random $\lambda$ drawn from ${\rm Beta}(1, 1)$. This suggests that our analysis of Mixup in Section~\ref{theoryRegression} based on a fixed $\lambda$ is also indicative for more general settings of $\lambda$. 
% Moreover, 
Figure~\ref{fig:toy-simulation} also indicates that when $\lambda$ is fixed to $0.5$, the increasing stage of the U-shaped curve comes earlier than that of  $\lambda$ with ${\rm Beta}(1, 1)$.  
%, we argue that this is because $\lambda=0.5$ provides larger noise level.
 This is consistent with our  theoretical results in Section~\ref{theoryRegression}. That is, owning to the fact that $\lambda$ with the constant value 0.5 for $\lambda$ represents the largest noise level in Mixup, the noise-dominating effect in Mixup training comes earlier.

% \textcolor{red}{
% \begin{enumerate}
%     \item This observation holds for the Data-Augmentation case (require longer training time)
%     \item Unlike ERM training, Gradient norm does not converge to zero in Mixup training, indeed it will increase to a maximum value.
%     \item This observation also holds for MSE without Data-Augmentation
%     \item This observation holds for different learning rate decay scheme.
% \end{enumerate}}

% \vspace{-2mm}

\subsection{Using Mixup Only in the Early Stage of Training} 
% \vspace{-3mm}
%We here aim to empirically verify that Mixup can induce  label noises as discussed in Section~\ref{mixupintroducenoise}.

\begin{figure}[!h]
    \centering
    \subfloat[CIFAR10 ($30\%$)\label{fig:cifar10-30-switch}]{%
       \includegraphics[width=0.25\linewidth]{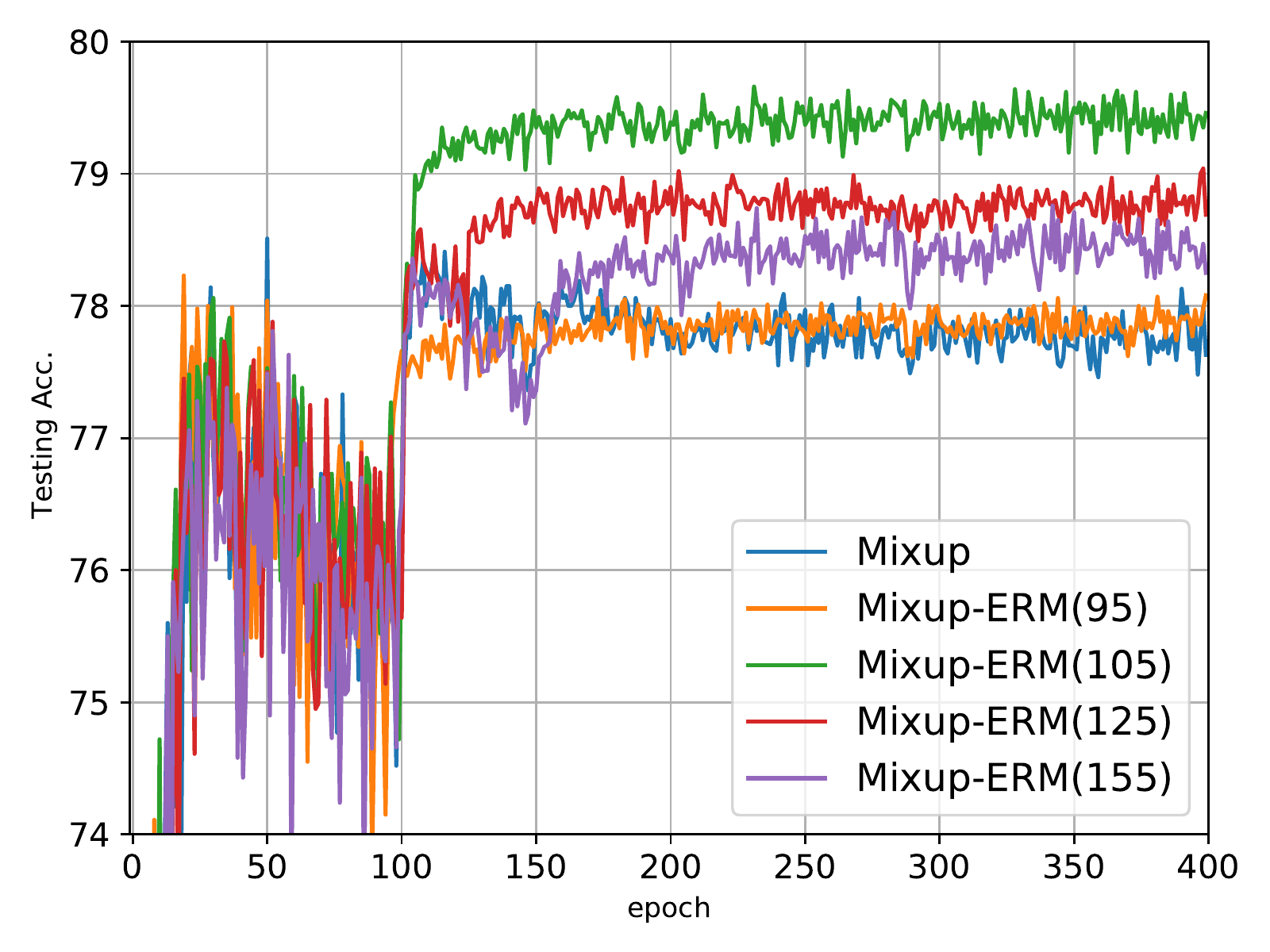}}
    \subfloat[CIFAR10 ($100\%$)\label{fig:cifar10-100-switch}]{%
       \includegraphics[width=0.25\linewidth]{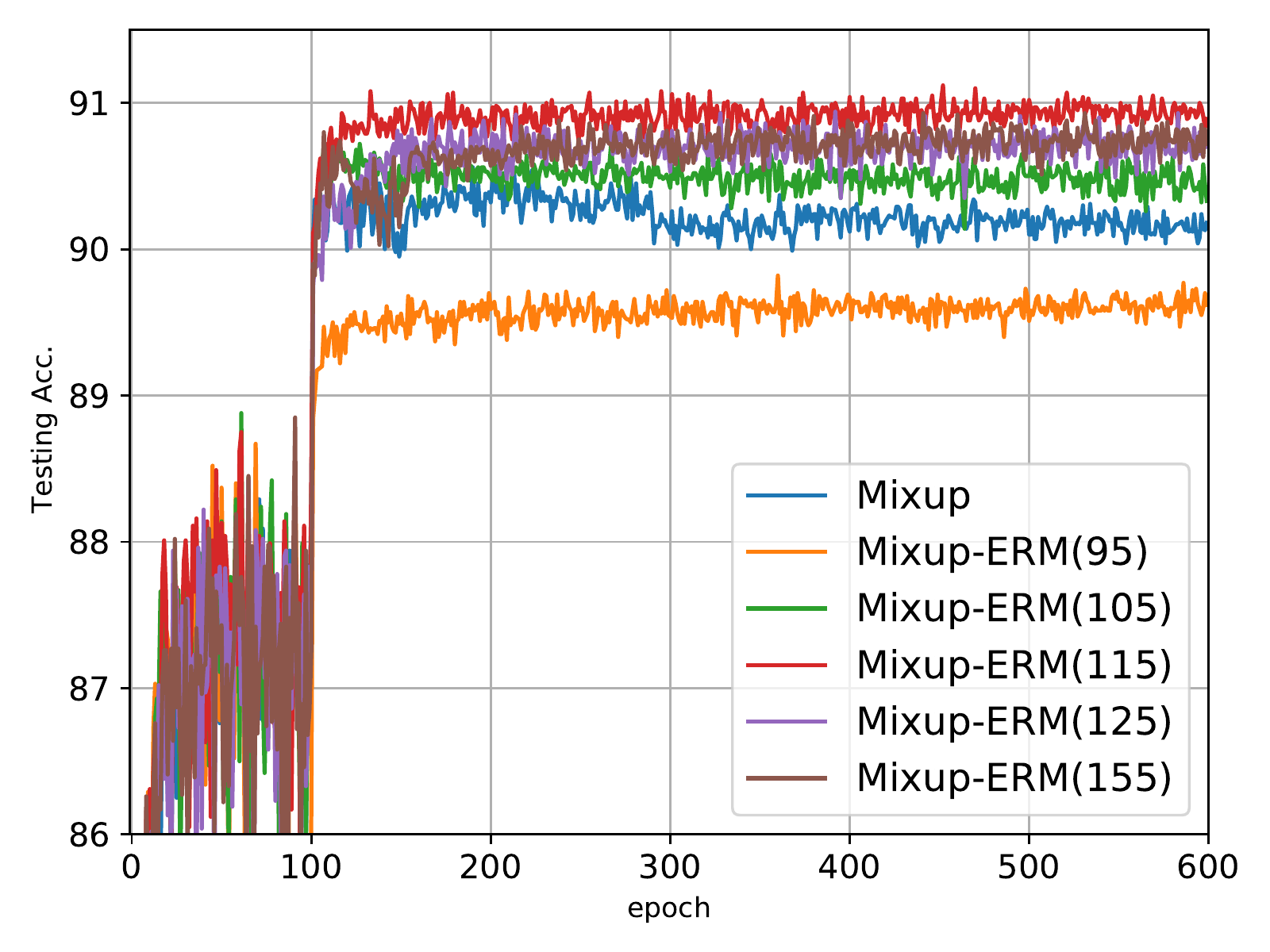}}
    \subfloat[SVHN ($30\%$)\label{fig:svhn-30-switch}]{%
       \includegraphics[width=0.25\linewidth]{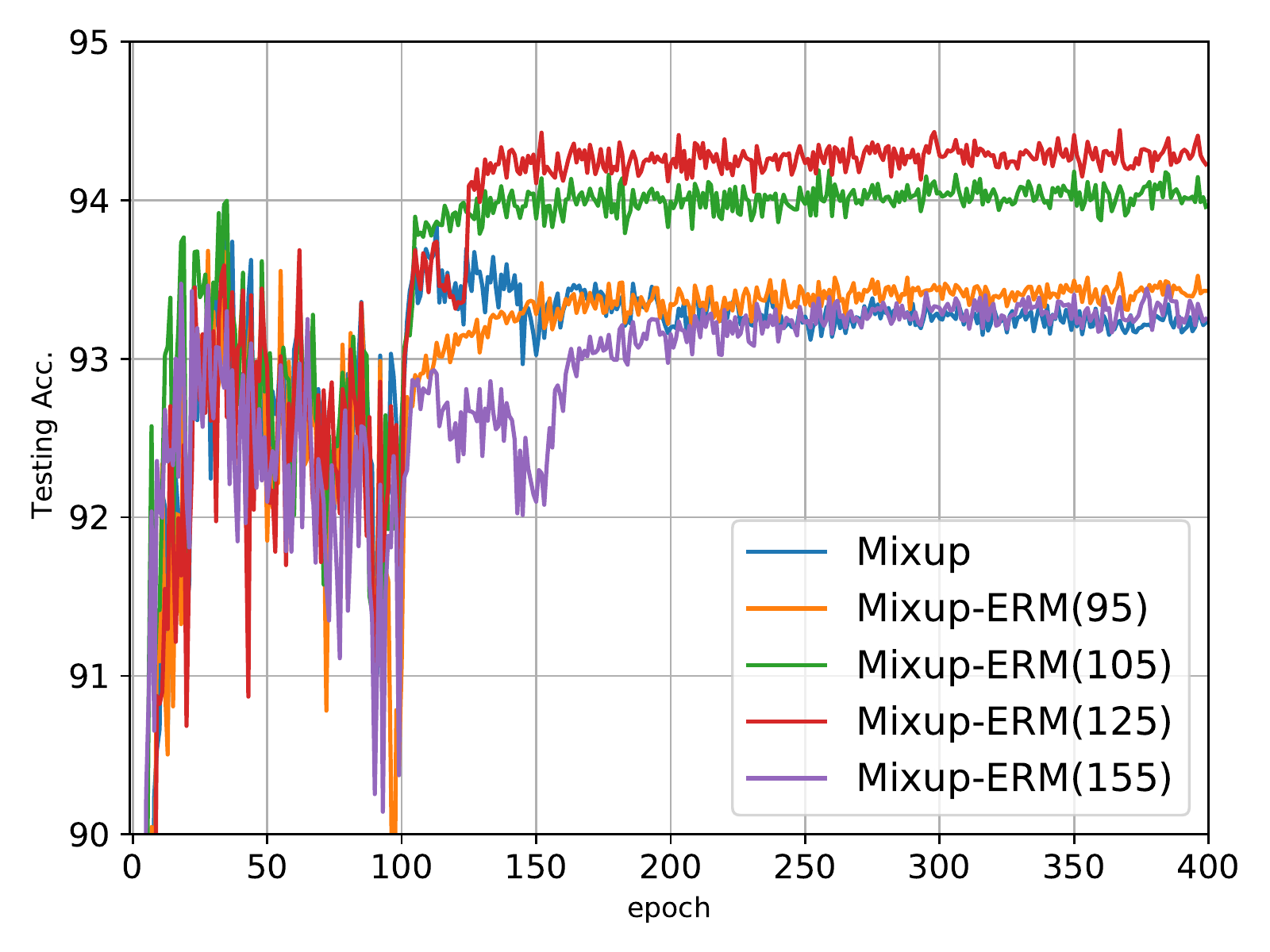}}
    \subfloat[SVHN ($100\%$)\label{fig:svhn-100-switch}]{%
       \includegraphics[width=0.25\linewidth]{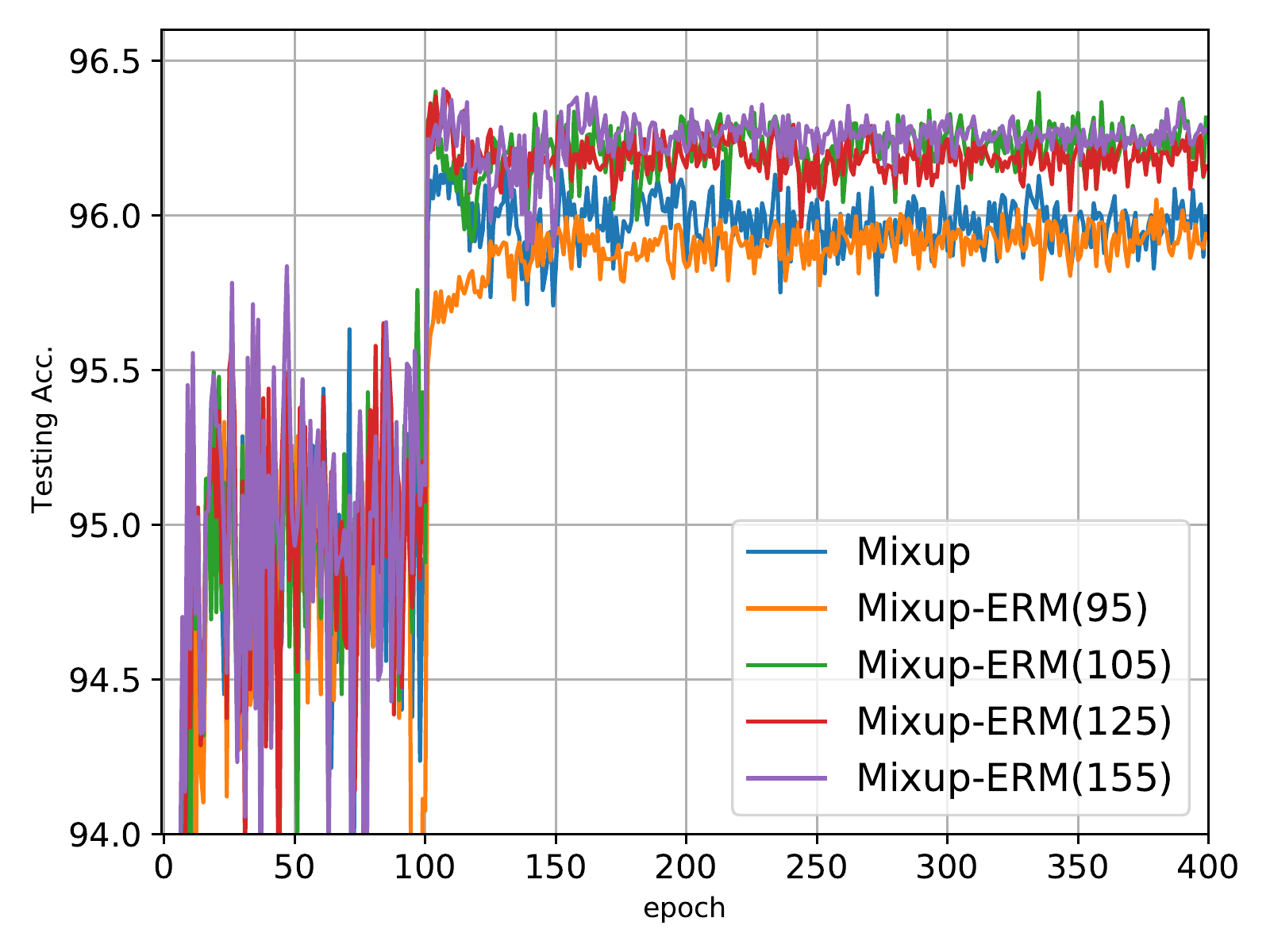}}
       \vspace{-3mm}
    \caption{Switching from Mixup training to ERM training. The number in the bracket is the epoch number where we let $\alpha=0$ (i.e.  Mixup training becomes ERM training).}
    % \vspace{-2.5mm}
    \label{fig:Mixup-ERM switch}
\end{figure}

In the previous section, we have argued that Mixup training learns ``clean patterns'' in the early stage of  the training {process} and then overfits the ``noisy patterns'' in the later stage.  Such a conclusion implies that turning of Mixup after a  certain number of epochs and returning to standard ERM training may prevent the training from overfitting the noises induced by Mixup. %In other words, we can change Mixup training to ERM training that does not contain induced noisy labels. 
%In this case, we utilize the advantages of both methods, that is, more informative data created by Mixup and less noisy labels of ERM. 
We now present results obtained from such a training scheme on both CIFAR10 and SVHN in Figure~\ref{fig:Mixup-ERM switch}.

% \vspace{-3.5mm}

Results in Figure~\ref{fig:Mixup-ERM switch}  clearly indicate that  switching from Mixup to ERM  at an appropriate time will successfully avoid the generalization degradation. 
%verifies this simple idea. 
Figure~\ref{fig:Mixup-ERM switch} also suggests that 
%It is  worth noting that, 
%Notice that
switching Mixup to ERM too early may not boost the model performance.
%it may not boost the model performance if we change Mixup to ERM before the  clean samples created by Mixup have large effect. 
In addition, if the switch is too late, 
%if we change Mixup to ERM too late 
memorization of noisy data may already taken effect, which impact generalization negatively. We  note that our results here can be regarded as a complement to \citep{golatkar2019time}, where the authors show that regularization techniques only matter during the early phase of learning.

\vspace{-2mm}
% \vspace{-1.5mm}
\section{Further Investigation}
\vspace{-2mm}
%\vspace{-1.5mm}
%\subsection{More Data Alleviate Overfitting} 

\begin{wrapfigure}{l}{0.66\textwidth}
    \centering
    \vspace{-3mm}
    \subfloat[ CIFAR10\label{fig:cifar10-number-test}]{%
      \includegraphics[width=0.33\textwidth]{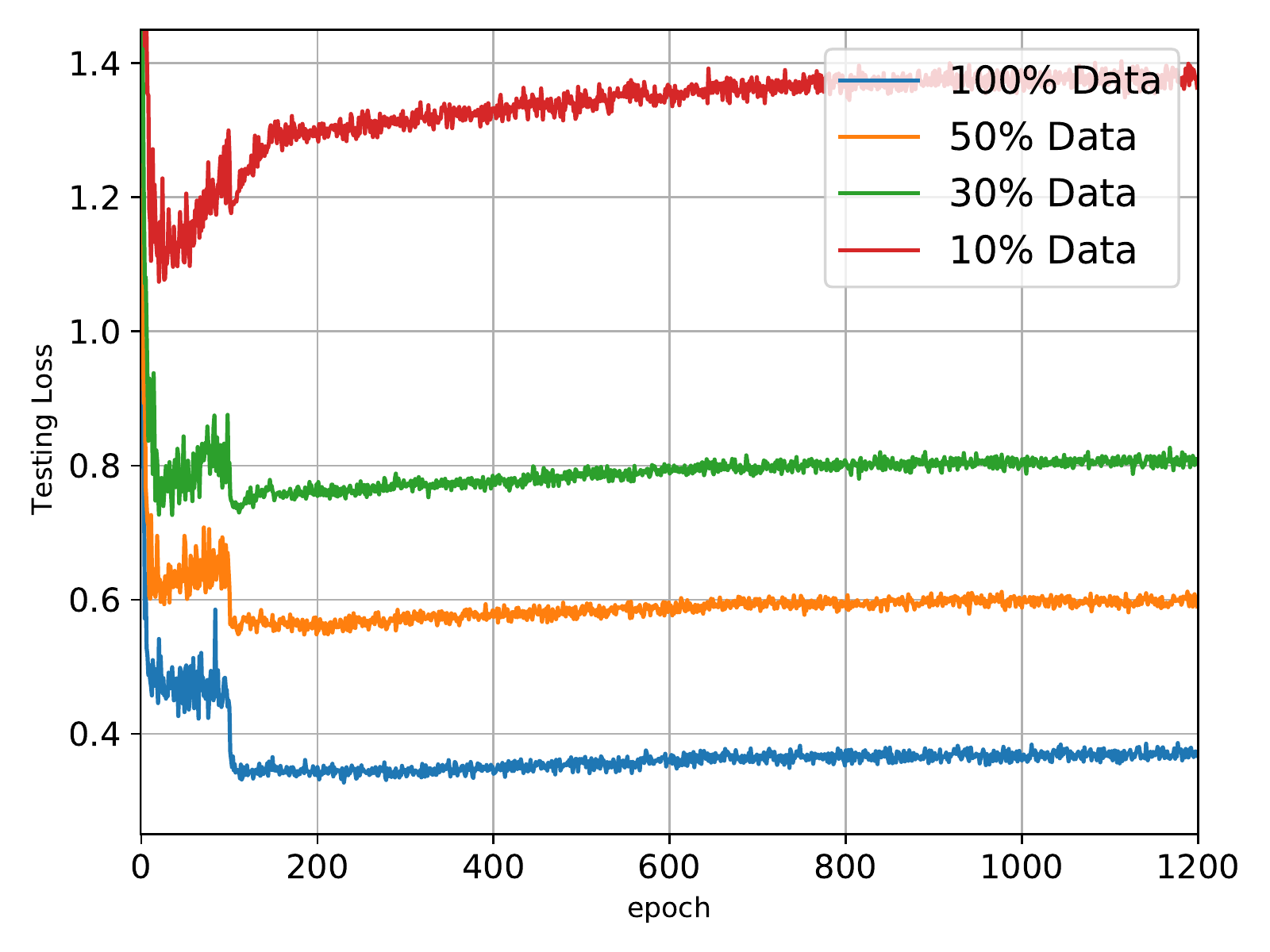}}
    %   \hfill
    \subfloat[SVHN \label{fig:svhn-number-test}]{%
      \includegraphics[width=0.33\textwidth]{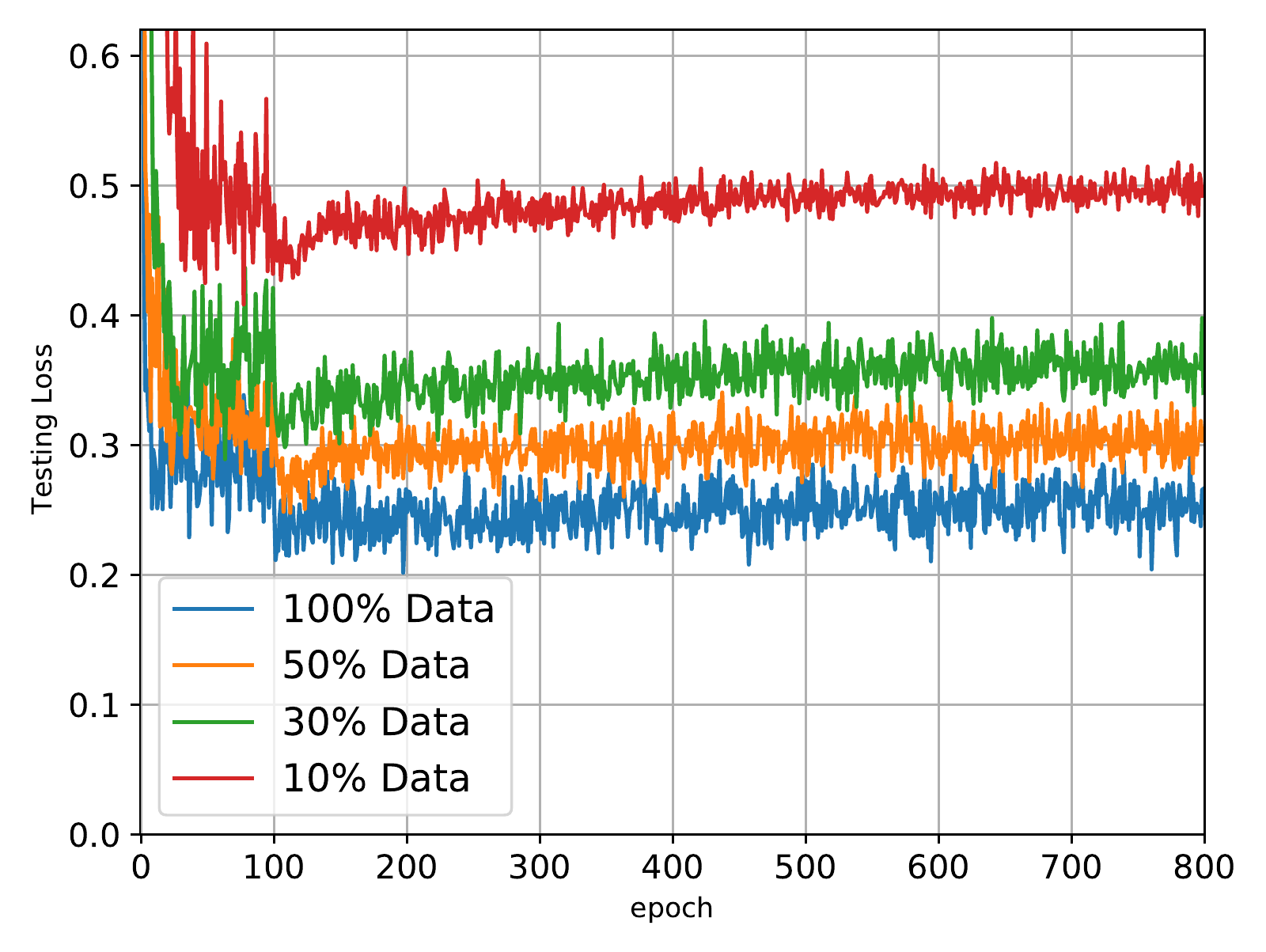}}
      \vspace{-3mm}
    \caption{Over-training on different number of samples.}
    \label{fig: LossOnNumberOfSamples}
\vspace{-4.5mm}
\end{wrapfigure}

\textbf{Impact of Data Size on U-shaped Curve} 
In the over-training experiments without data augmentation, although the U-shaped behavior occurs on both $100\%$ and $30\%$ of the original training data for both CIFAR10 and SVHN, we notice that smaller size %of (original)
datasets  appear to enable the turning point of the U-shaped curve to arrive earlier. We now corroborate this phenomenon with more experimental results, as shown in Figure~\ref{fig: LossOnNumberOfSamples}. 
In this context, 
an appropriate data augmentation can be seen as simply expanding the training set with additional clean data. Then the impact of data augmentation on the over-training dynamics of Mixup is arguably via increasing the size of the training set. This explains our observations in Section~\ref{subsec:results of cifar10 with data augmentation} where the turning points in training with data augmentation arrive much later compared to those without data augmentation.  Those observations are  also consistent with the results in Figure~\ref{fig: LossOnNumberOfSamples}.

% \vspace{-1mm}

 It may be tempting to consider the application of the usual analysis of generalization dynamics from the existing literature \citep{liao2018dynamics,advani2020high,stephenson2021and} to the training of Mixup.
%  , where they utilize some tools from random matrix theory. 
 For example, one can analyze the distribution of the eigenvalues in Theorem~\ref{thm:mixup-dynamic}. Specifically, if entries in ${\Phi}$ are independent identically distributed  with zero mean, then 
 in the limit of $d,m\rightarrow \infty$ with $d/m=\gamma\in(0,+\infty)$,
 the eigenvalues $\{\mu_k\}_{k=1}^d$ follow the Marchenko-Pasteur (MP) distribution \citep{marvcenko1967distribution}, which is defined as
 \vspace{-1mm}
\[
P^{MP}(\mu|\gamma)=\frac{1}{2\pi}\frac{\sqrt{(\gamma_+-\mu)(\mu-\gamma_-)}}{\mu\gamma}\mathbf{1}_{\mu\in [\gamma_-,\gamma_+]},
\]
where $\gamma_{\pm}=(1\pm \gamma)^2$. Note that the $P^{MP}$ are only non-zero when $\mu=0$ or $\mu\in [\gamma_-,\gamma_+]$. 
When $\gamma$ is close to one, the probability of
extremely small eigenvalues is immensely increased. 
From Theorem~\ref{thm:mixup-dynamic}, when $\mu_k$ is small, the second term, governed by the noisy pattern,  will badly dominate the behavior of population risk and converge to a larger value. Thus, letting $d\ll m$ will alleviate the domination of the noise term in Theorem~\ref{thm:mixup-dynamic}. However, it is important to note that such analysis lacks rigor since the columns in $\Phi$ are not independent (two columns might result from linearly combining  the same pair of original instances). To apply a similar analysis here, one  need to remove or relax the independence conditions on the entries of ${\Phi}$,
%for the MP distribution to hold, 
for example, by invoking some techniques similar to that developed in \cite{bryson2021marchenko}. This is beyond the scope of this paper,  and we will to leave it for future study.

\begin{wrapfigure}{r}{0.66\textwidth} %this figure will be at the left
 \vspace{-5mm}
    \centering
        \subfloat[Gradient norm on CIFAR10\label{fig:cifar10-gradient-norm}]{%
      \includegraphics[width=0.33\textwidth]{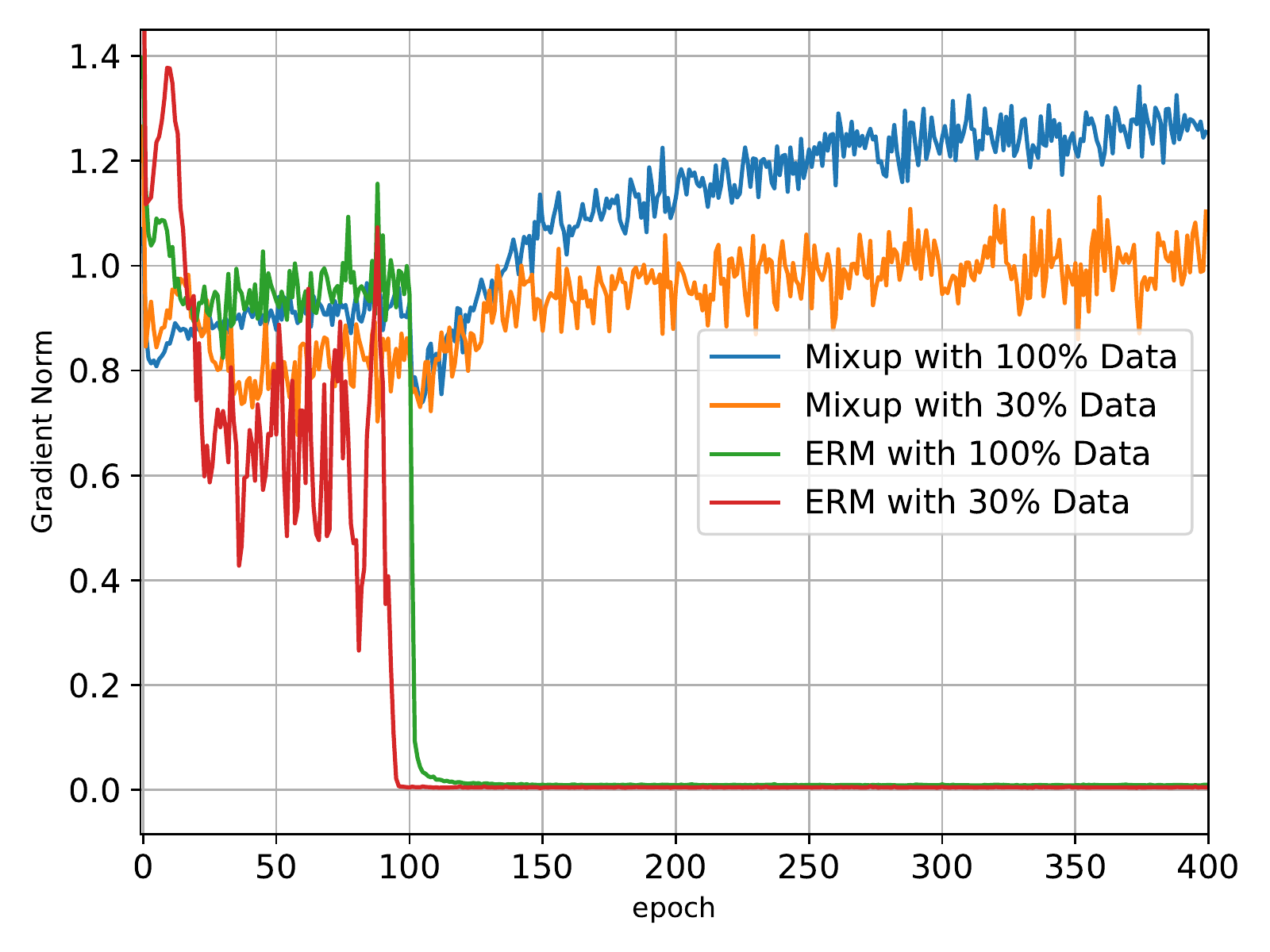}}
    \subfloat[ Gradient norm on SVHN\label{fig:svhn-gradient-norm}]{%
      \includegraphics[width=0.33\textwidth]{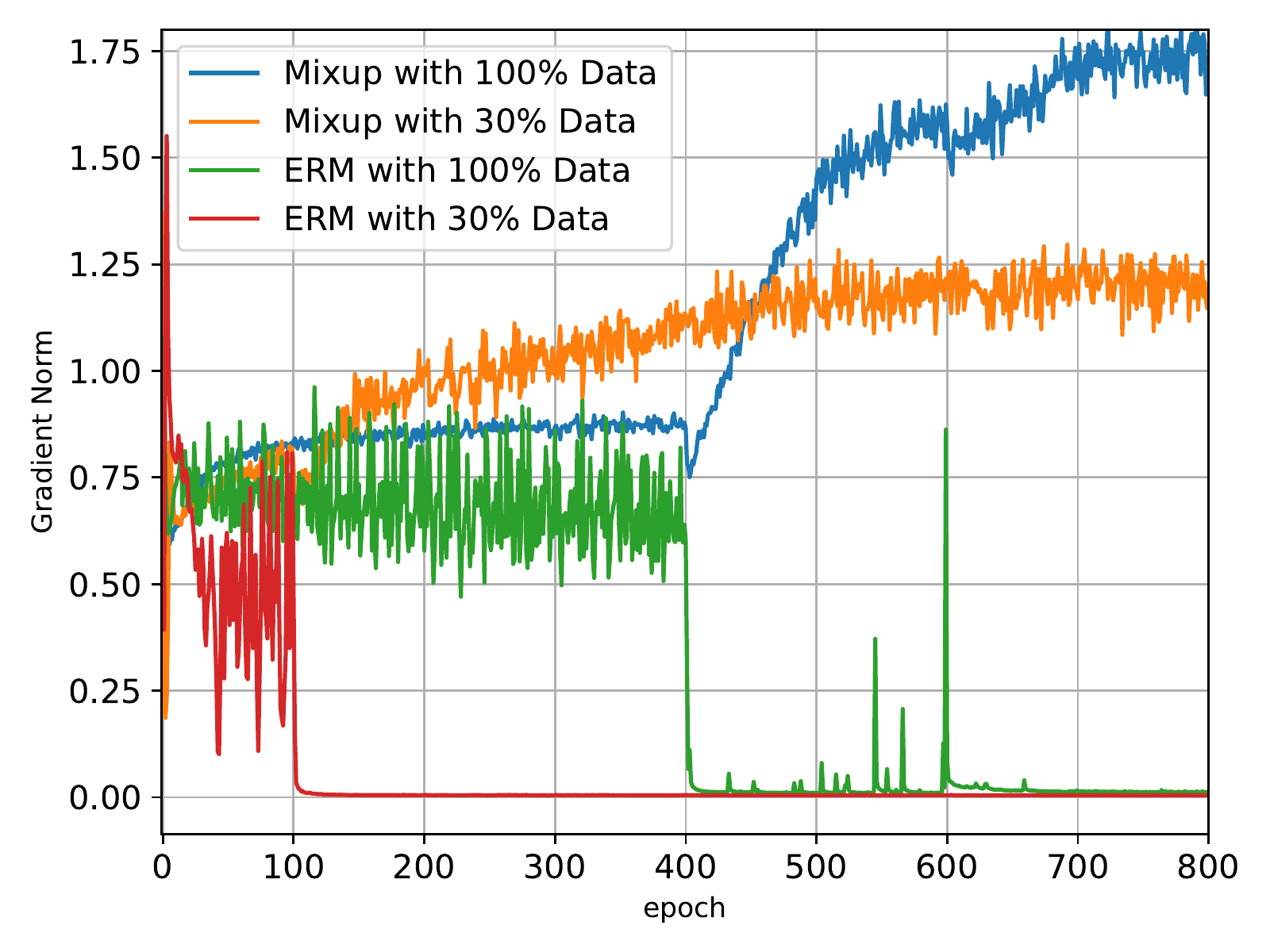}}
     \vspace{-3mm}
    \caption{Dynamics of gradient norm.}
\label{fig:gradient-norm}
\vspace{-4mm}
\end{wrapfigure}

\textbf{Gradient Norm in Mixup Training Does Not Vanish} Normally, ERM training obtains zero gradient norm at the end of training, which indicates that SGD finds a local minimum. However, We observe that the gradient norm of Mixup training does not converge to zero, as shown in Figure~\ref{fig:gradient-norm}. 

In fact, gradient norm in the Mixup training even increases until converging to a maximum value, as opposed to zero. When models are trained with ERM on random labels, this increasing trend of gradient norm is also observed in the previous works \citep{feng2021phases,wang2022on}. Specifically, in \cite{wang2022on}, such increasing behavior is interpreted as a sign that the training of SGD enters a ``memorization regime'', and after the overparameterized neural networks memorize all the noisy labels, the gradient norm (or gradient dispersion in \cite{wang2022on}) will decrease again until it converges to zero. In Mixup training, since the size of synthetic dataset is usually larger than the number of parameters (i.e., $m>d$), neural networks may not be able to memorize all the noisy labels in this case. Notice that $m$ is much larger than $n^2$ in practice since $\lambda$ is not fixed to a constant.

Notably, although ERM training is able to find a local minimum in the first 130 epochs on CIFAR10, Figure~\ref{fig: CIFAR10 loss & acc curves: introduction} indicates that Mixup training outperforms ERM in the first 400 epochs. Similar observation also holds for SVHN. This result in fact suggests 
%be view as conveying a message 
that Mixup can generalize well without converging to any stationary points. Notice that there is a related observation in the recent work of \cite{zhang2022neural}, where they show that large-scale neural networks generalize well without having the gradient norm vanish during training.
Additionally, by switching Mixup training to ERM training, as what we did in Figure~\ref{fig:Mixup-ERM switch}, the gradient norm will instantly become zero (see Figure~\ref{fig:Mixup-switch-norm} in Appendix~\ref{sec:mixup-erm-gn}). This further justifies that the ``clean patterns'' are already learned by Mixup trained neural networks at the early stage of training, and the original data may no longer provide any useful gradient signal. 



%\vspace{-4mm}
\section{Concluding Remarks}
%\vspace{-4mm}
We discovered a novel phenomenon in Mixup: over-training with Mixup may give rise to a  U-shaped generalization curve. 
We theoretically show that 
this is due to the data-dependent label noises introduced to the synthesized data, and suggest that Mixup improves generalization through fitting the clean patterns at the early training stage,  but  over-fits the noise as training proceeds. 
%We also provided experimental indications that unlike ERM training, Mixup can generalize well without converging to any stationary points. 
%This work points to several promising directions worth further study. For example,
%How can Mixup be improved  leveraging our discovery?
%the U-shaped generalization behavior identified  to devise a training paradigm for Mixup to automatically optimize its regularization effect would be  beneficial. 
The effectiveness of Mixup and the fact it works by only partially optimizing its loss function without reaching convergence, as are validated by our analysis and experiments, seem to suggest that the dynamics of the iterative learning algorithm and an appropriate criteria for terminating the algorithm might be more essential than the loss function or the solutions to the optimization problem. Exploration in the space of iterative algorithms (rather than the space of loss functions) may lead to fruitful discoveries.

%Second, unifying Mixup's generalization behavior here with that being pointed out by previous works such as~\cite{arpit2017closer,liu2020early,feng2021phases,wang2022on,pmlr-v162-liu22w} would be  useful.  Finally, theoretically verifying that Mixup generalizes well without converging to any stationary points would help improve our understanding on Mixup's generalization capabilities.  
\subsubsection*{Acknowledgments}
This work is supported in part by a National Research Council of Canada (NRC) Collaborative R\&D grant (AI4D-CORE-07). 
Ziqiao Wang is also supported in part by the NSERC CREATE program through the Interdisciplinary Math and Artificial Intelligence (INTER-MATH-AI) project.
% Ziqiao Wang is supported partly by the Interdisciplinary Math and Artificial Intelligence (INTER-MATH-AI) project which is awarded by an NSERC CREATE Program. 

% \newpage
\bibliography{ref}
\bibliographystyle{iclr2023_conference}

\newpage
\appendix

\section{Experimental Setups of Over-training}\label{appendix:overtrain experiments setups}
For any experimental setting (
% For any set of experimental setup 
such as the training dataset and its size, whether ERM or Mixup is used,  whether other data augmentation is used, etc.), we define a training ``trial''  as a training process starting from random initialization to a certain epoch $t$. In each trial, we record the minimum training loss obtained during the entire training process. The testing accuracy of the model's intermediate solution that gives rise to that minimum training loss is also recorded. For different trials we gradually increase $t$ so as to gradually let the model be over-trained. For each $t$, we repeat the trial for $10$ times with different random seeds and  collect all the recorded results (minimum training losses and the corresponding testing accuracies). We then compute their averages and standard deviations for all $t$'s. These results are eventually used to plot the line graphs for presentation.

% In each training trial, we train the network for in total a fixed number of epochs. We record the minimal training loss achieved by the network during the training process, and we also record the network's testing accuracy of the epoch at which the minimal training loss is achieved. Additionally, we visualize the local loss landscape around the solution found by the network at the aforementioned training epoch. We gradually increase the total number of the training epochs in different trials of training so as to gradually over-train the network. 
% We repeat each training trial for $10$ times (using $10$ different random seeds) and we average the recorded training losses and testing accuracies.
For example, Figure~\ref{subfig:0.3 cifar10 resnet18 trainloss} illustrates the results of training ResNet18 on $30\%$ CIFAR10 data without data augmentation. The total number of training epochs $t$, as shown on the horizontal axis, is increased from 100 to 1600. For each $t$, each point on its vertical axis represents the average of the recorded training losses from the 10 repeats. The width of the shade beside each point reflects the corresponding standard deviation.

\section{Additional Experimental Results}
\subsection{Additional Results of Over-training Without Data Augmentation}\label{appendix:additional overtrain results}
{
Besides CIFAR10, ResNet18 is also used for the SVHN dataset. }

\begin{figure}[!ht]
    \centering
    \subfloat[Train loss ($30\%$ {data})\label{0.3 svhn resnet18 trainloss}]{%
       \includegraphics[width=0.262\linewidth]{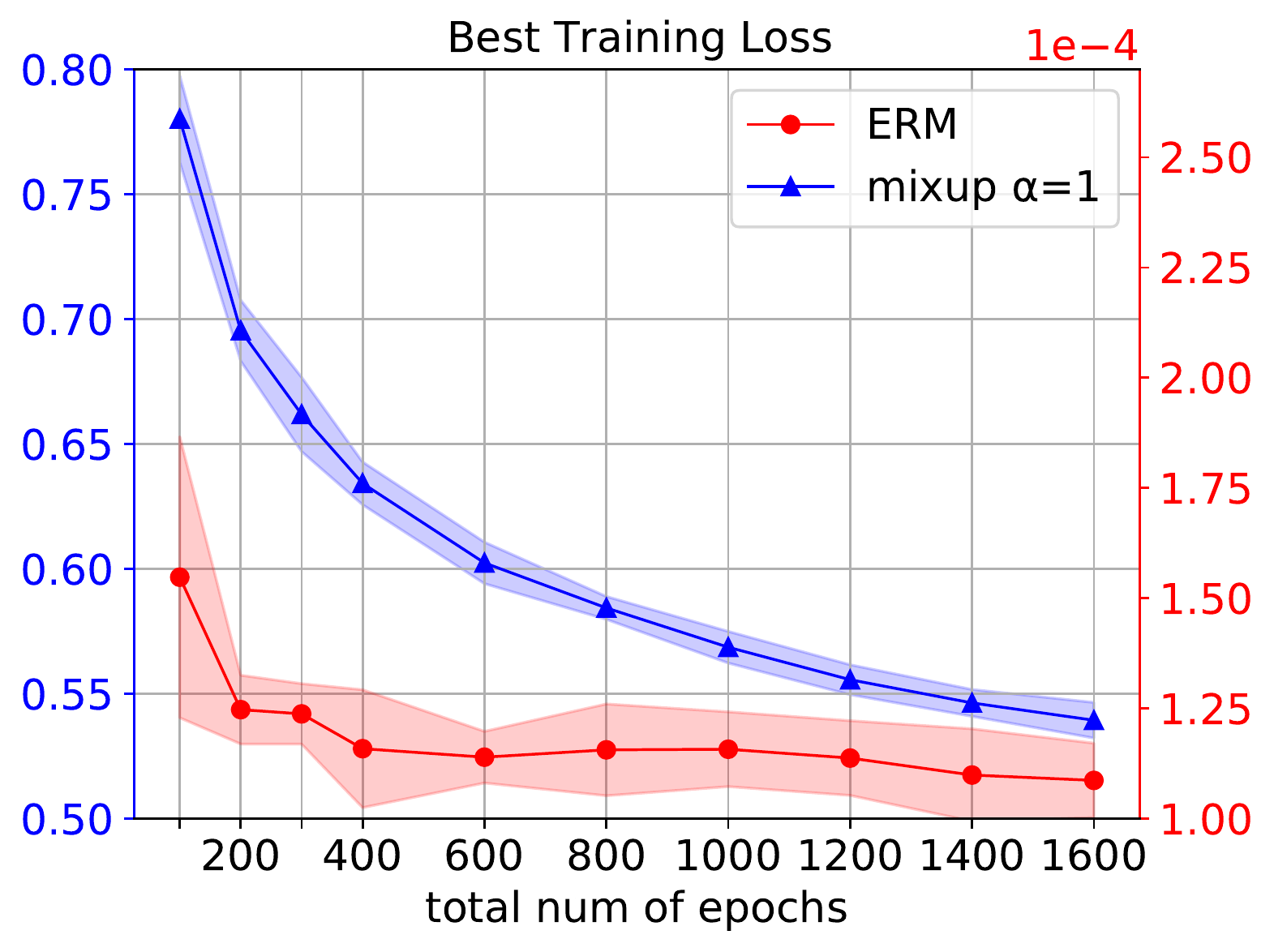}}
       \hfill
    \subfloat[Test acc ($30\%$ {data})\label{0.3 svhn resnet18 testacc}]{%
       \includegraphics[width=0.232\linewidth]{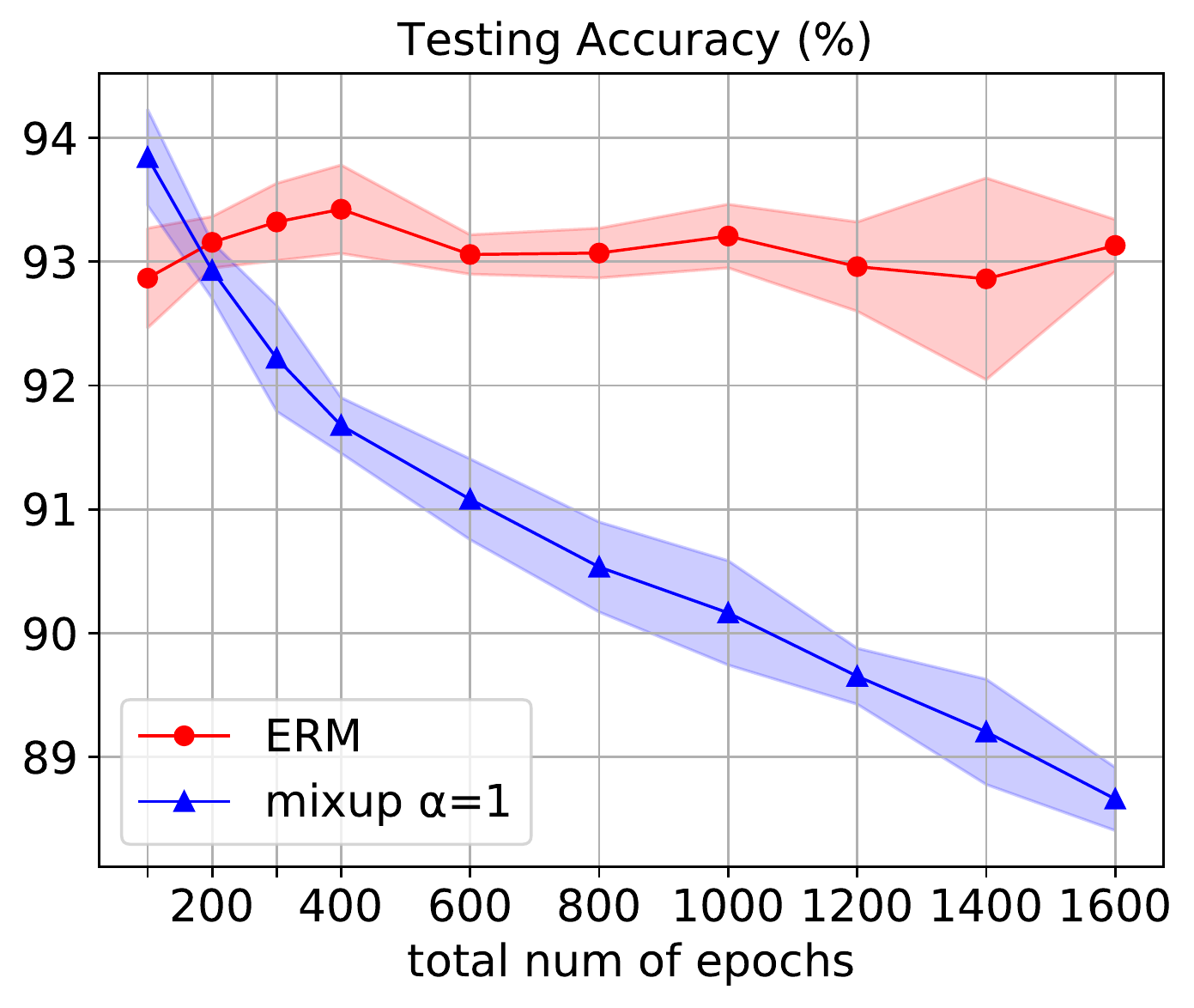}}
       \hfill
    \subfloat[Train loss ($100\%$ {data})\label{svhn resnet18 trainloss}]{%
       \includegraphics[width=0.261\linewidth]{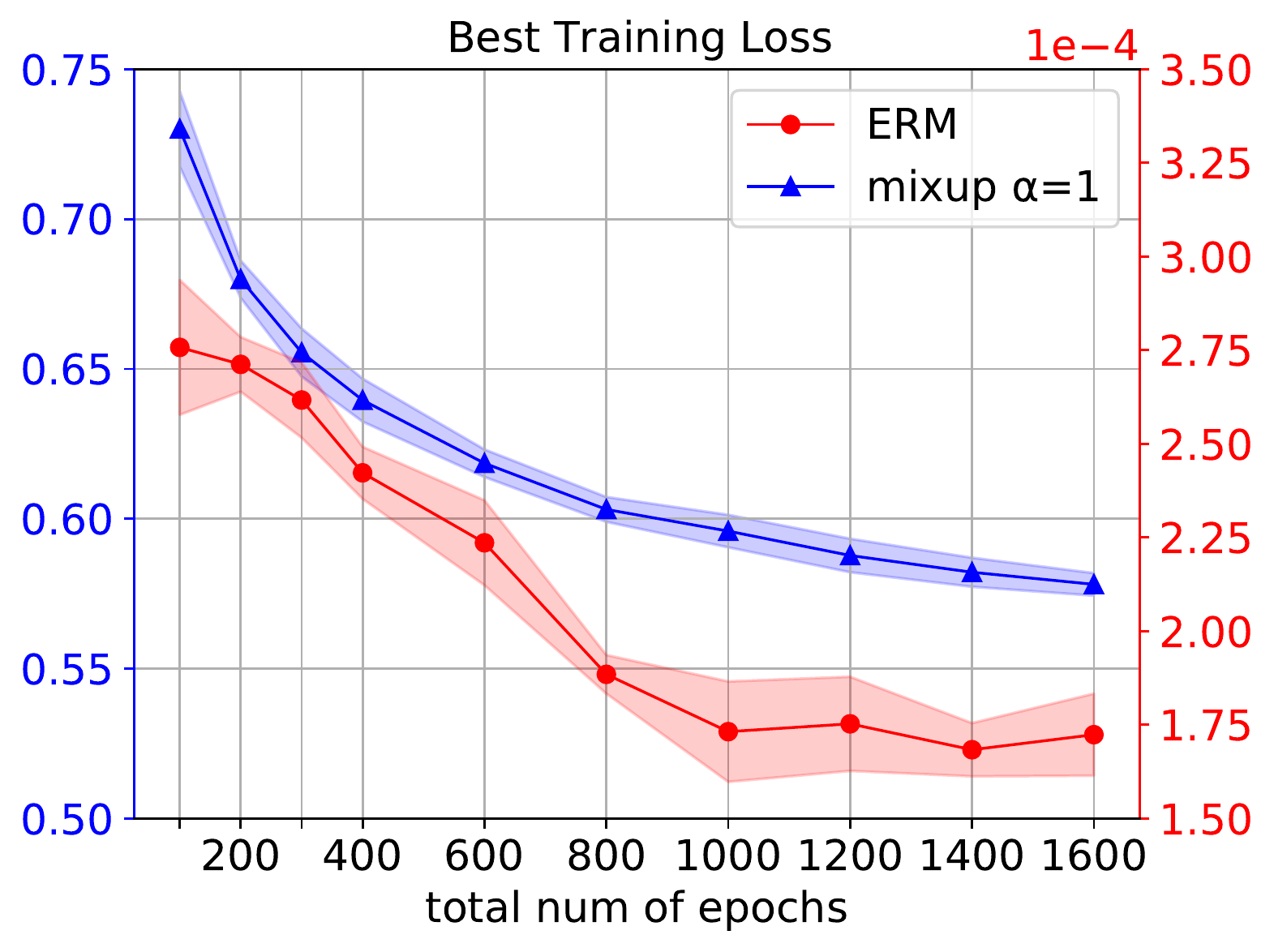}}
       \hfill
    \subfloat[Test acc ($100\%$ {data})\label{svhn resnet18 testacc}]{%
       \includegraphics[width=0.237\linewidth]{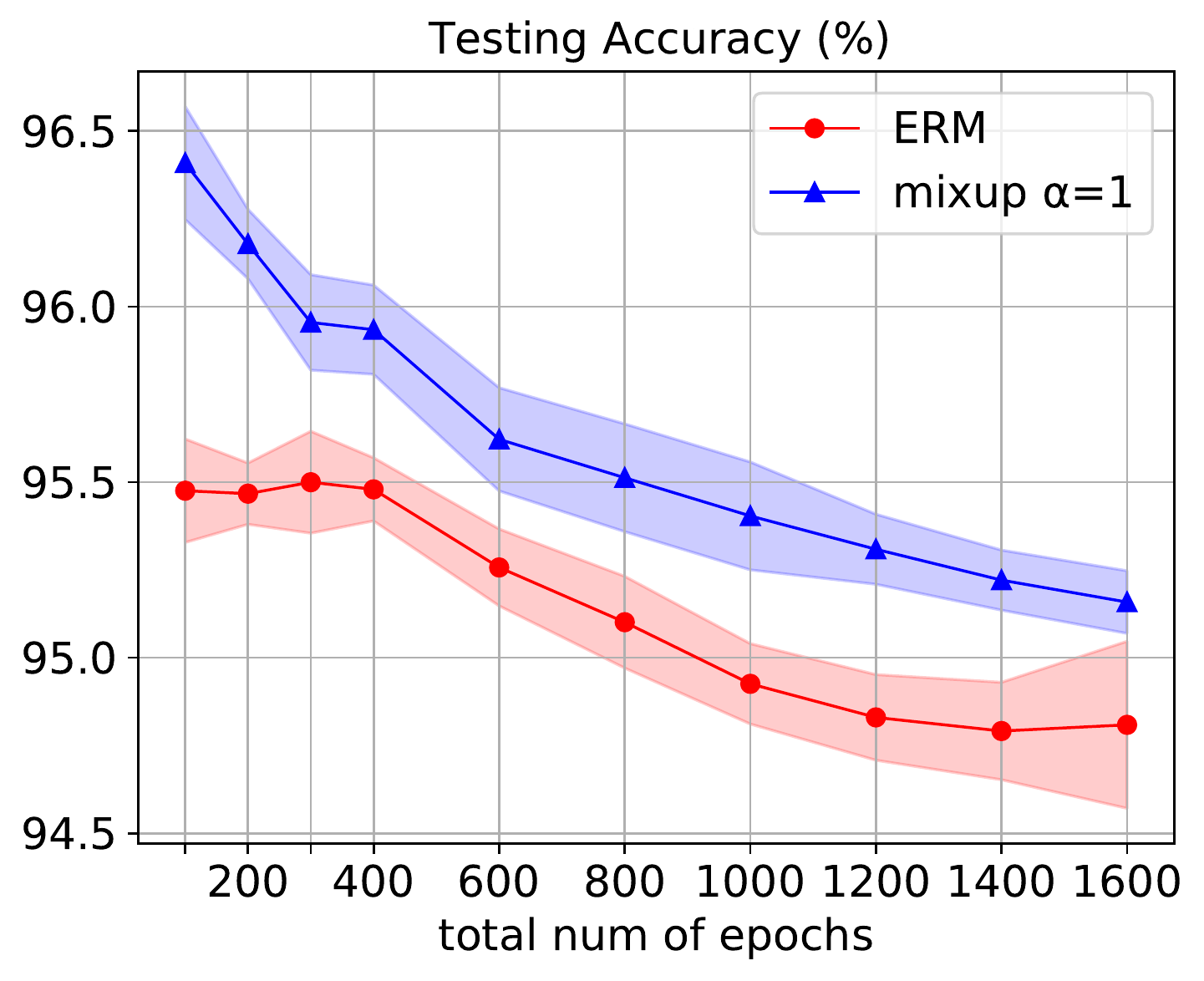}}\\
    \subfloat[200 epochs\label{0.3 svhn resnet18 landscape e200}]{%
       \includegraphics[width=0.16\linewidth]{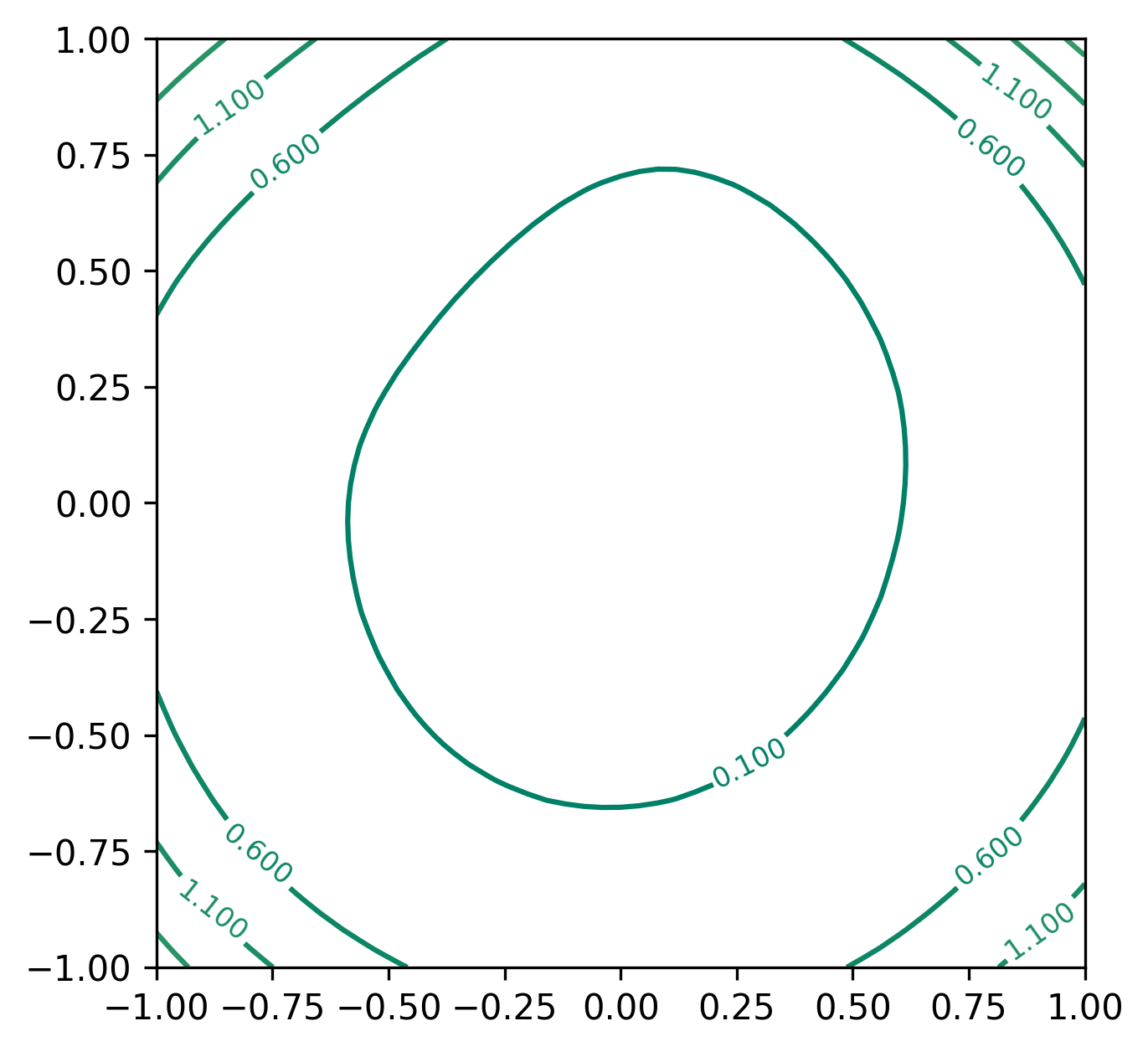}}
       \hfill
    \subfloat[400 epochs\label{0.3 svhn resnet18 landscape e400}]{%
       \includegraphics[width=0.16\linewidth]{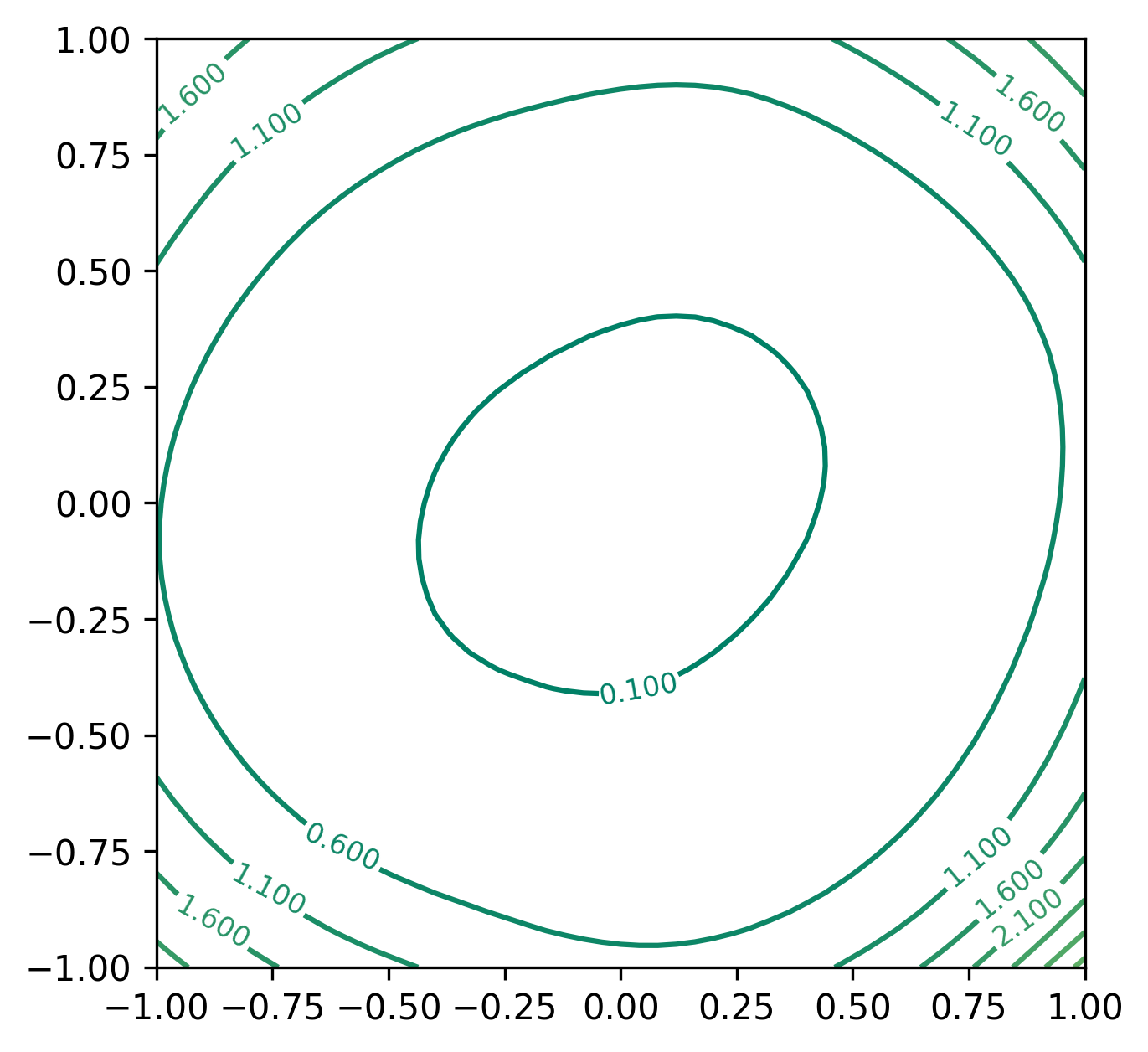}}
       \hfill
    \subfloat[800 epochs\label{0.3 svhn resnet18 landscape e800}]{%
       \includegraphics[width=0.16\linewidth]{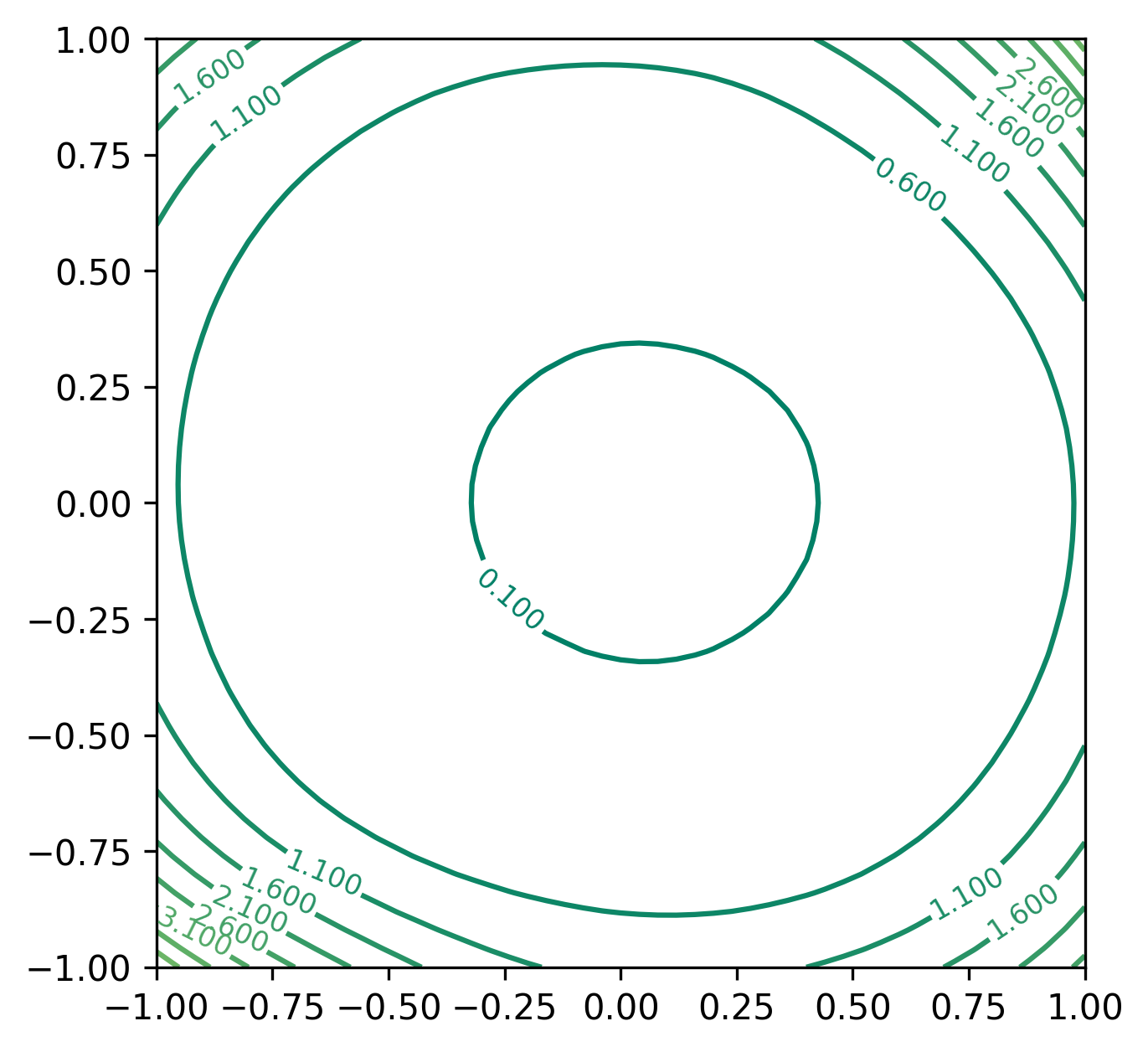}}
       \hfill
    \subfloat[200 epochs\label{svhn resnet18 landscape e200}]{%
       \includegraphics[width=0.16\linewidth]{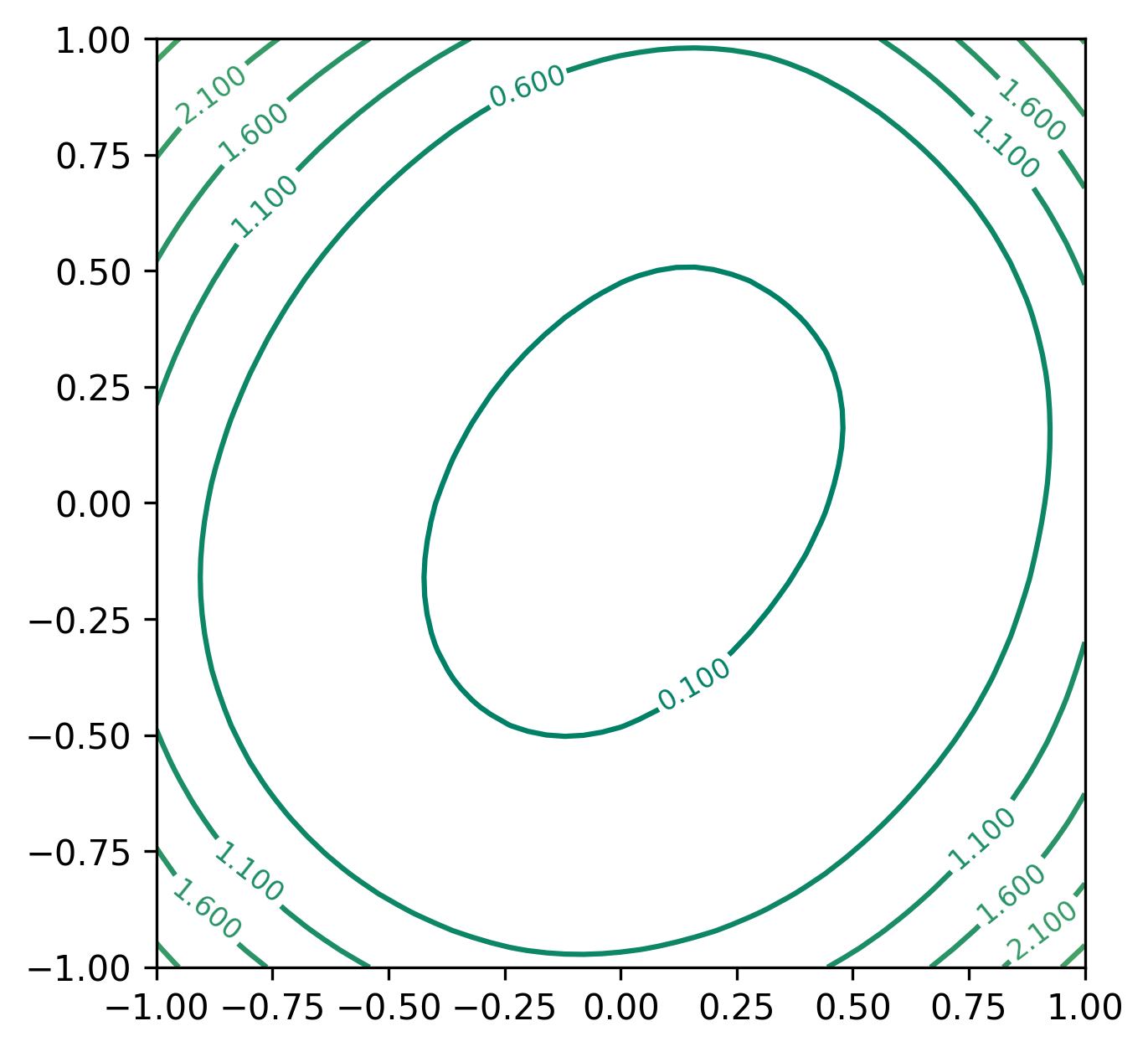}}
       \hfill
    \subfloat[400 epochs\label{svhn resnet18 landscape e400}]{%
       \includegraphics[width=0.16\linewidth]{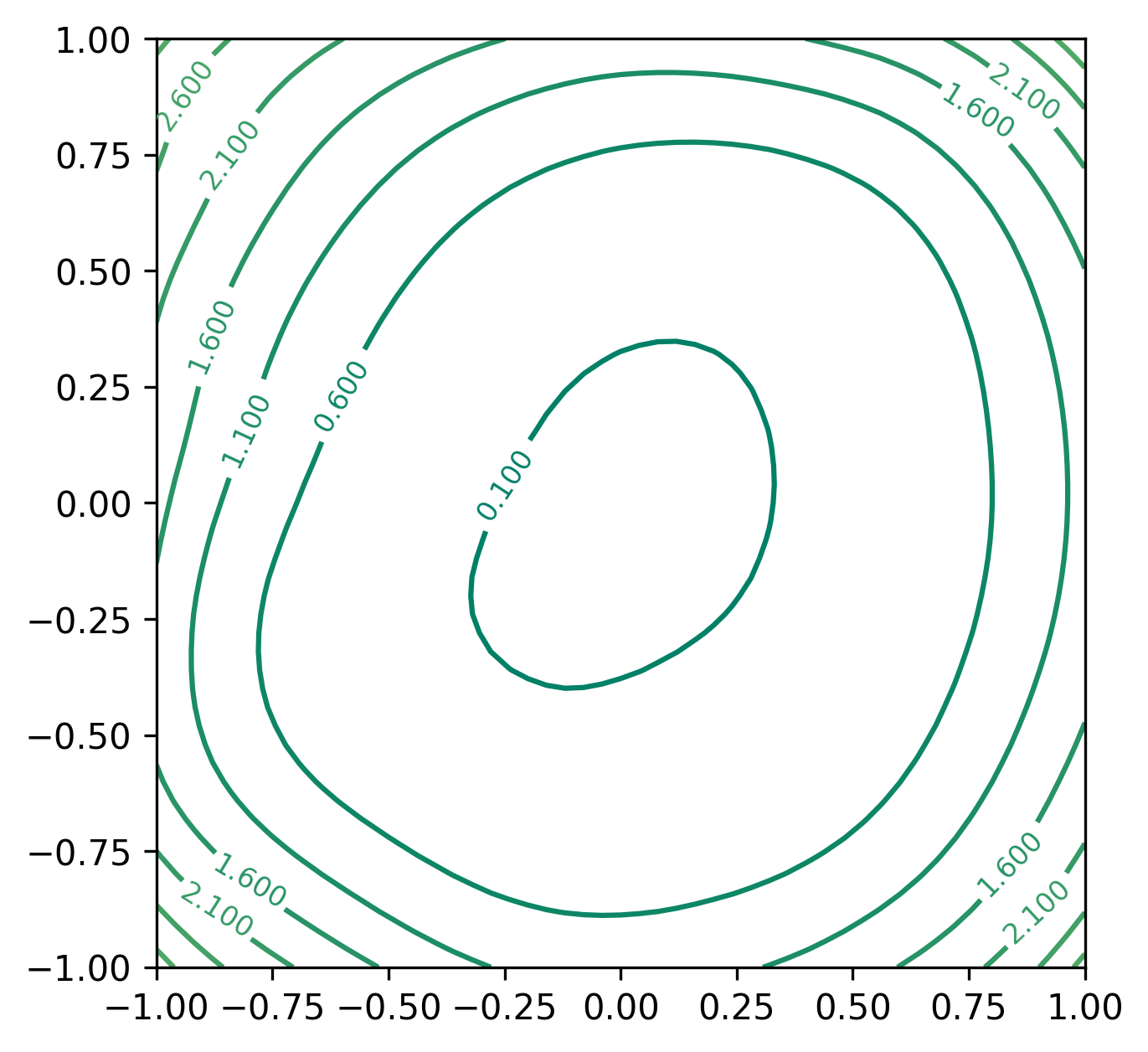}}
       \hfill
    \subfloat[800 epochs\label{svhn resnet18 landscape e800}]{%
       \includegraphics[width=0.16\linewidth]{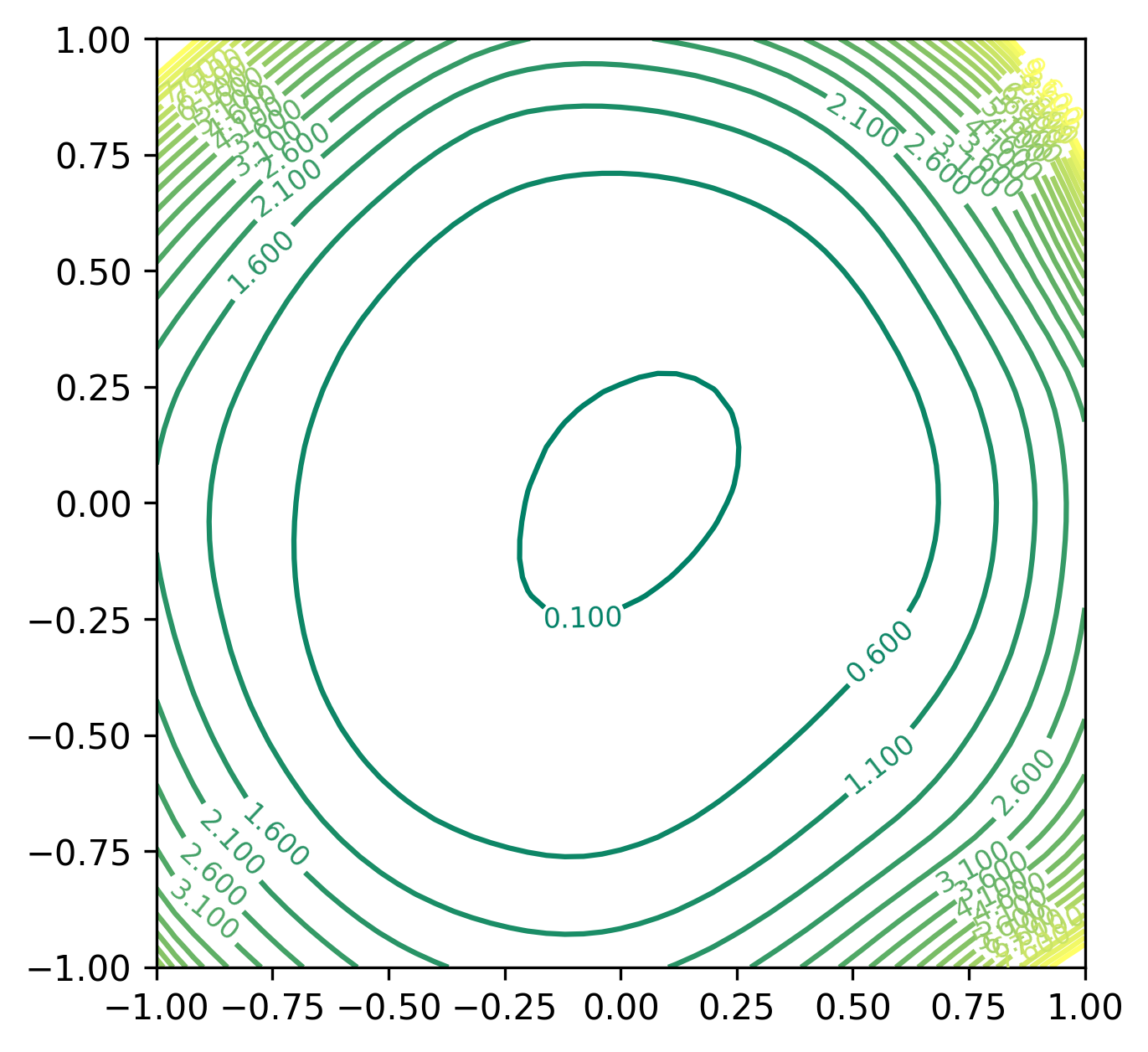}}
    \caption{{
    Results of training ResNet18 on SVHN  {training set} ($100\%$ data and $30\%$ data) without data augmentation. Top raw: training loss and testing accuracy for ERM and Mixup. Bottom raw: loss landscape of the Mixup-trained ResNet18 at various training epochs: the left $3$ figures are for the $30\%$ SVHN dataset, and the right $3$ are for the full SVHN dataset.
    }}
    % \vspace{-5mm}
    \label{fig: SVHN loss & acc curves}
\end{figure}

{
Training is performed for up to $1000$ epochs for SVHN, since we notice that if we continue training ResNet18 on SVHN after $1000$ epochs, the variance of the testing accuracy severely increases. The 
results are presented in Figure~\ref{fig: SVHN loss & acc curves}. 
Mixup exhibits a similar phenomenon as it does for CIFAR10. 
What differs notably is that  over-training with ERM on the original SVHN training set appears to also lead to worse test accuracy.
However, this does not occur on the $30\%$ SVHN training set\footnote{This might be related to the epoch-wise double descent behavior of  ERM training. That is, when over-training ResNet18 on the whole training set with a total of $1000$ epochs, the network is still in the first stage of over-fitting the training data, while when over-training the network on $30\%$ of the training set, the network learns faster on the training data due to the smaller sample size,
thus it passes the turning point of the double descent curve earlier.}.
}

{As for ResNet34, besides CIFAR100, it is also used for {both} the CIFAR10 and the SVHN datasets. }

\begin{figure}[!ht]
    \centering
    \subfloat[Train loss ($30\%$ {data})]{%
       \includegraphics[width=0.2635\linewidth]{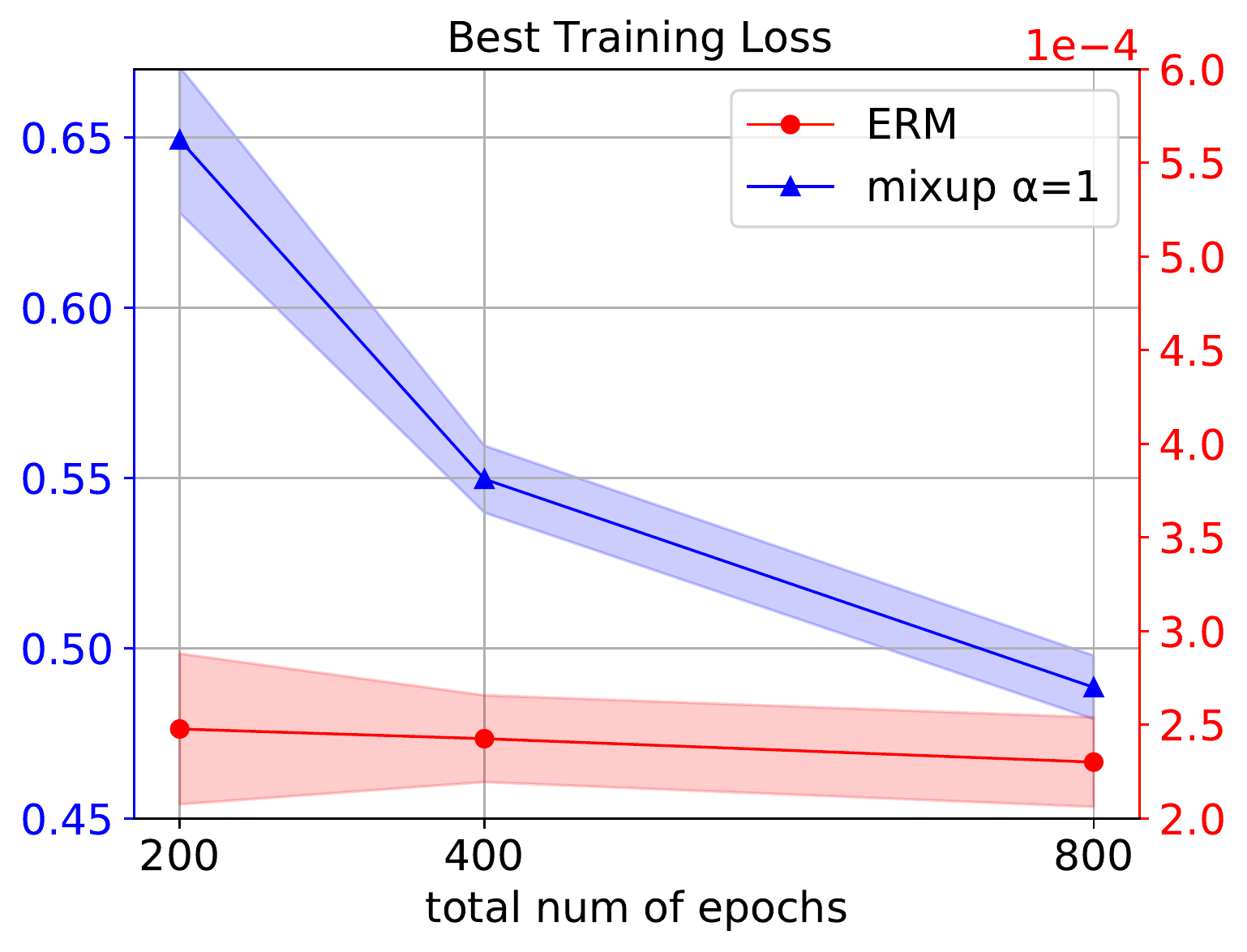}}
       \hfill
    \subfloat[Test acc ($30\%$ {data})]{%
       \includegraphics[width=0.2365\linewidth]{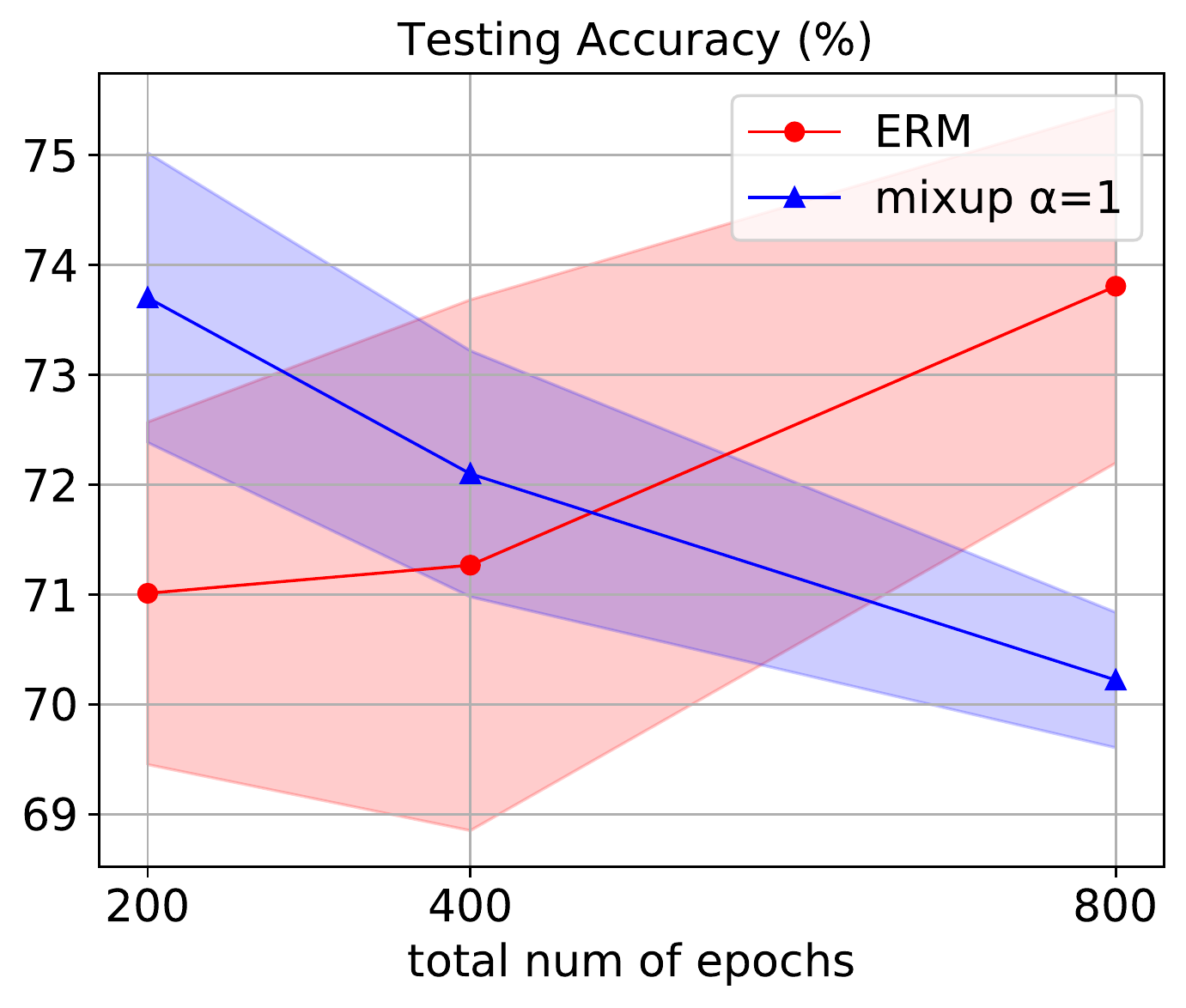}}
       \hfill
    \subfloat[Train loss ($100\%$ {data})]{%
       \includegraphics[width=0.2635\linewidth]{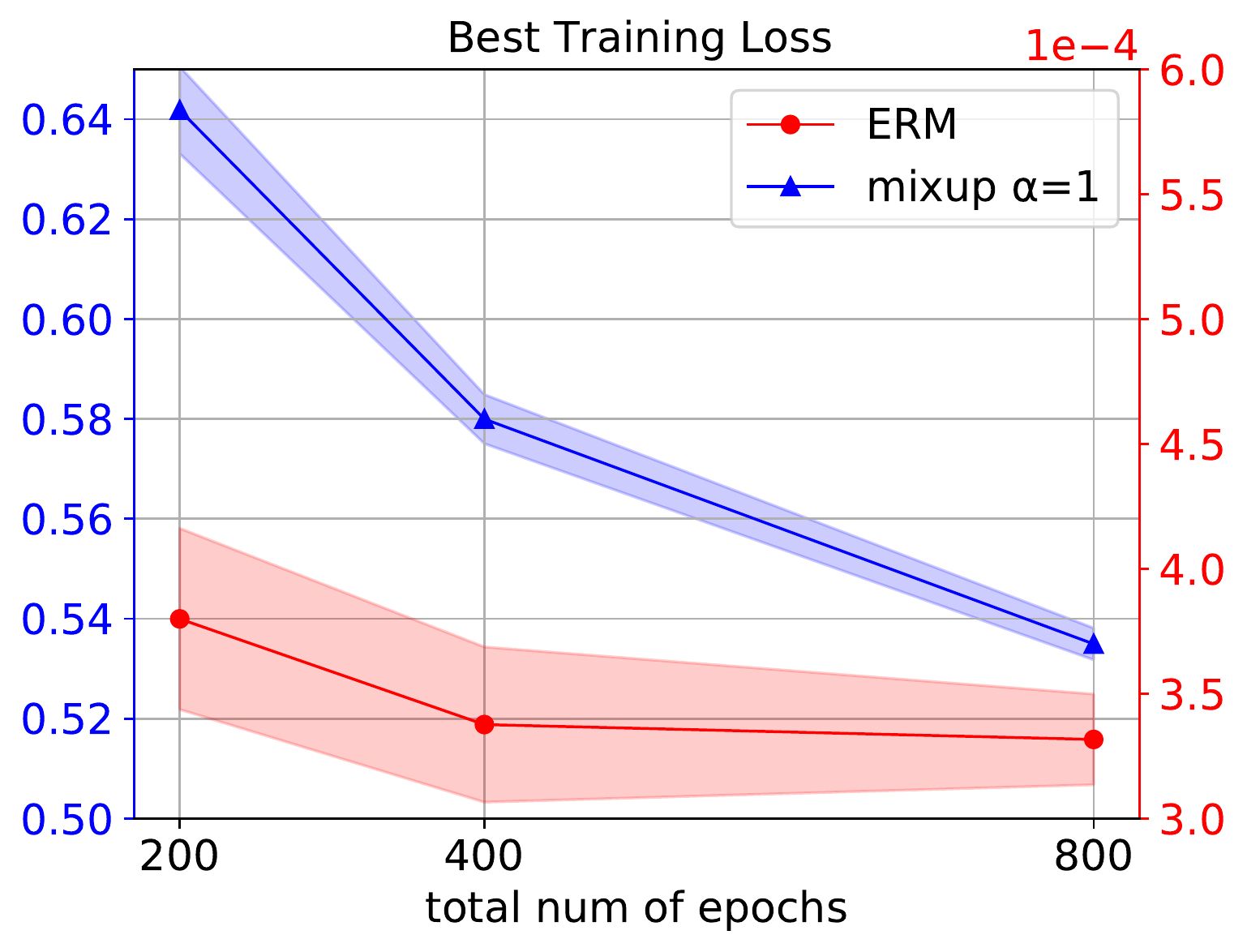}}
       \hfill
    \subfloat[Test acc ($100\%$ {data})]{%
       \includegraphics[width=0.2365\linewidth]{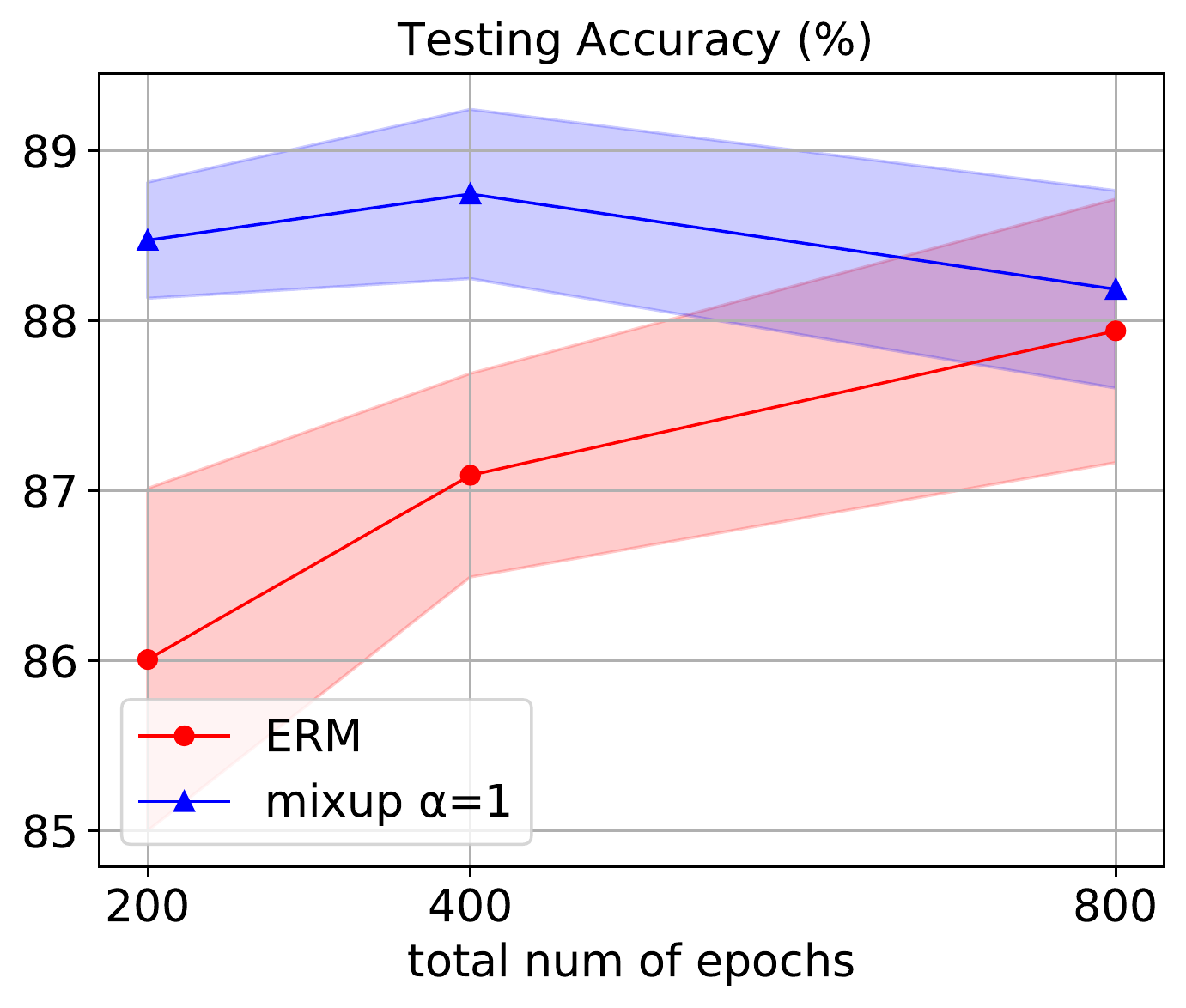}}
    \caption{
    Results of the recorded training losses and testing accuracies of training ResNet34 on CIFAR10 {training set} ($100\%$ data and $30\%$ data) without data augmentation.
    }
    \label{fig: cifar10 resnet34 loss & acc curves}
\end{figure}

Training is performed on both {datasets} for in total $200$, $400$ and $800$ epochs respectively. The results for CIFAR10 are shown in Figure~\ref{fig: cifar10 resnet34 loss & acc curves}. For both the 30\% dataset and the original dataset, Mixup exhibits a similar phenomenon as it does in training ResNet18 on CIFAR10. The difference is that over-training ResNet34 with ERM let the testing accuracy gradually increase on both the 30\% dataset and the original dataset.

\begin{figure}[!ht]
    \centering
    \subfloat[Train loss ($30\%$ {data})]{%
       \includegraphics[width=0.262\linewidth]{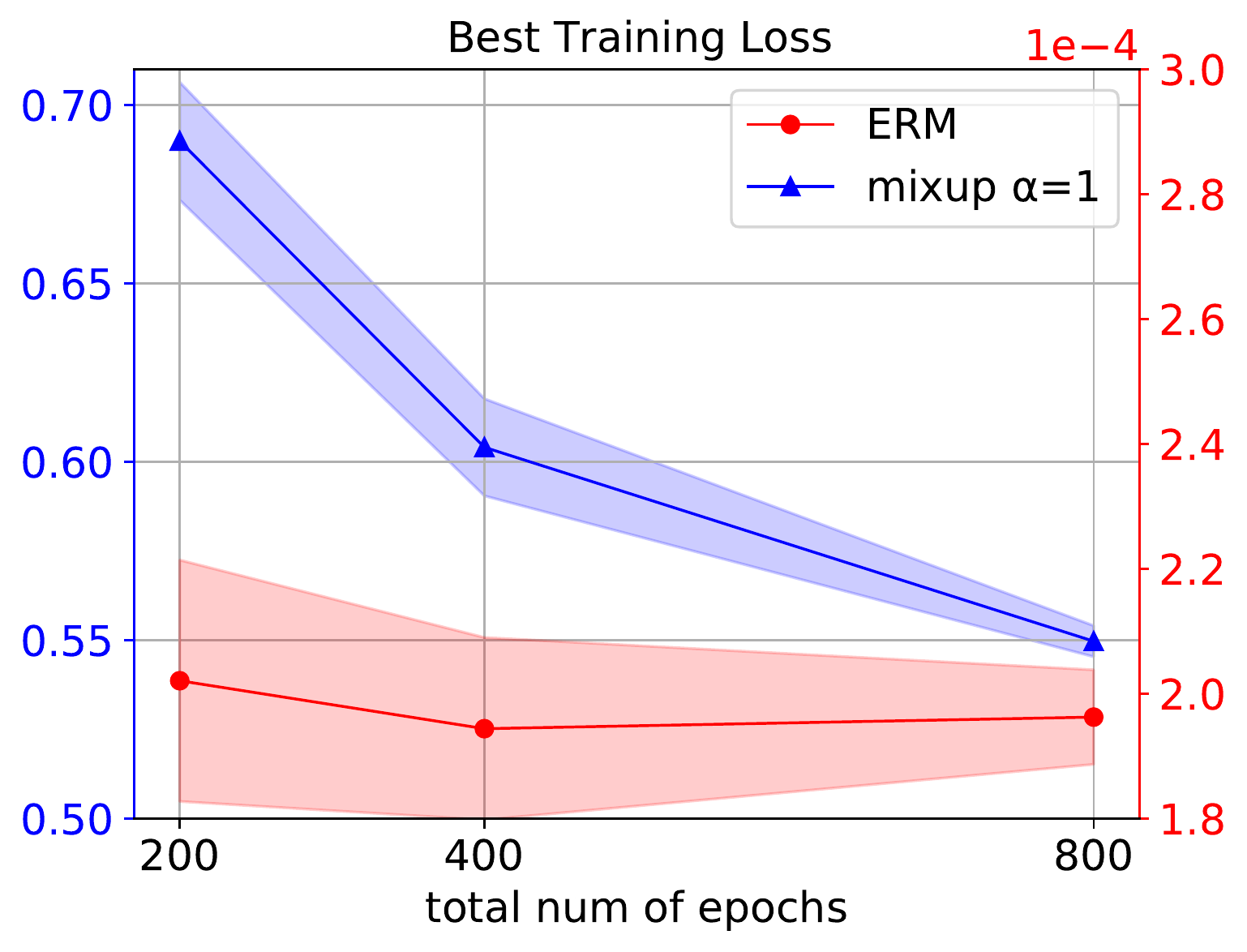}}
       \hfill
    \subfloat[Test acc ($30\%$ {data})]{%
       \includegraphics[width=0.235\linewidth]{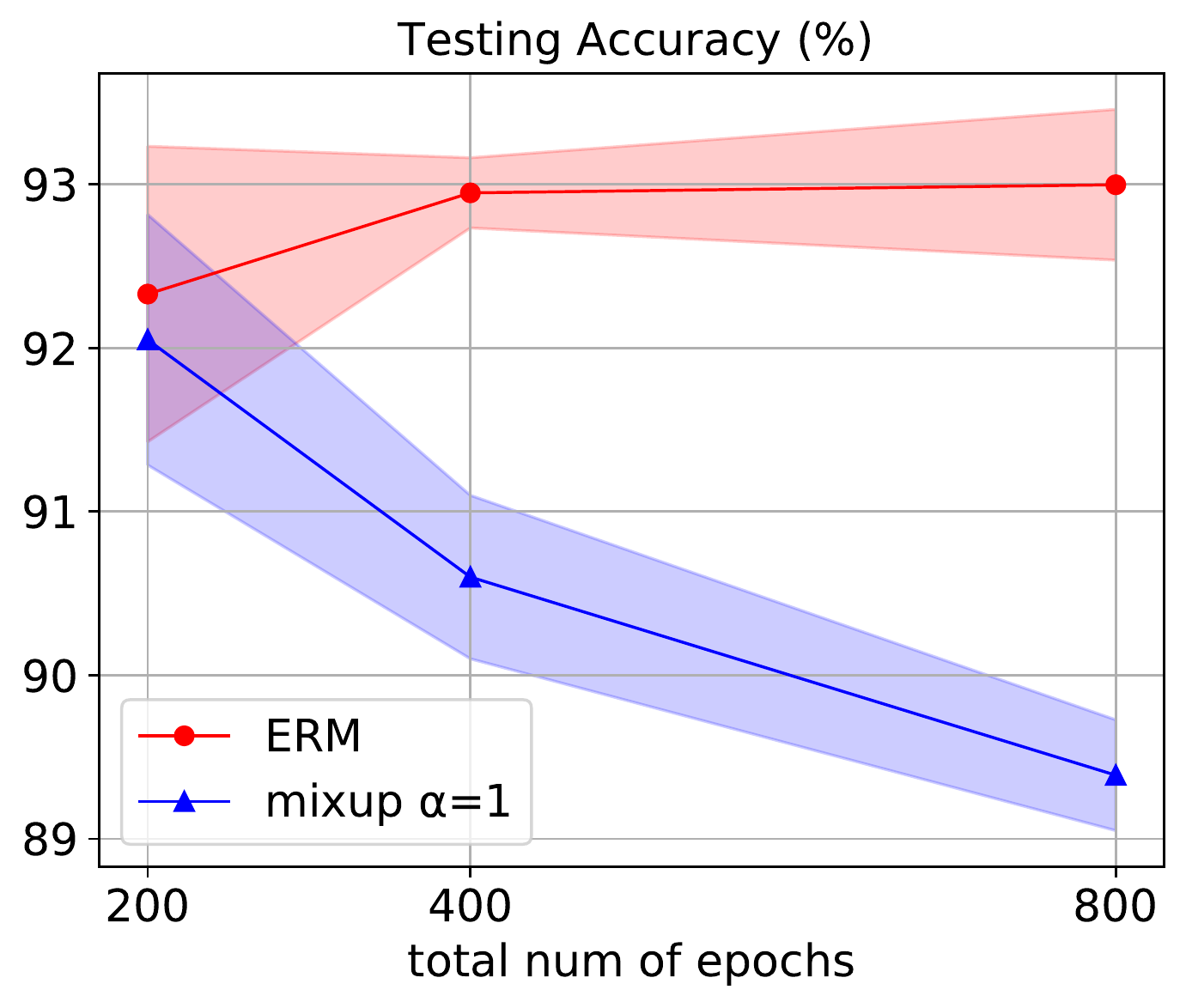}}
       \hfill
    \subfloat[Train loss ($100\%$ {data})]{%
       \includegraphics[width=0.262\linewidth]{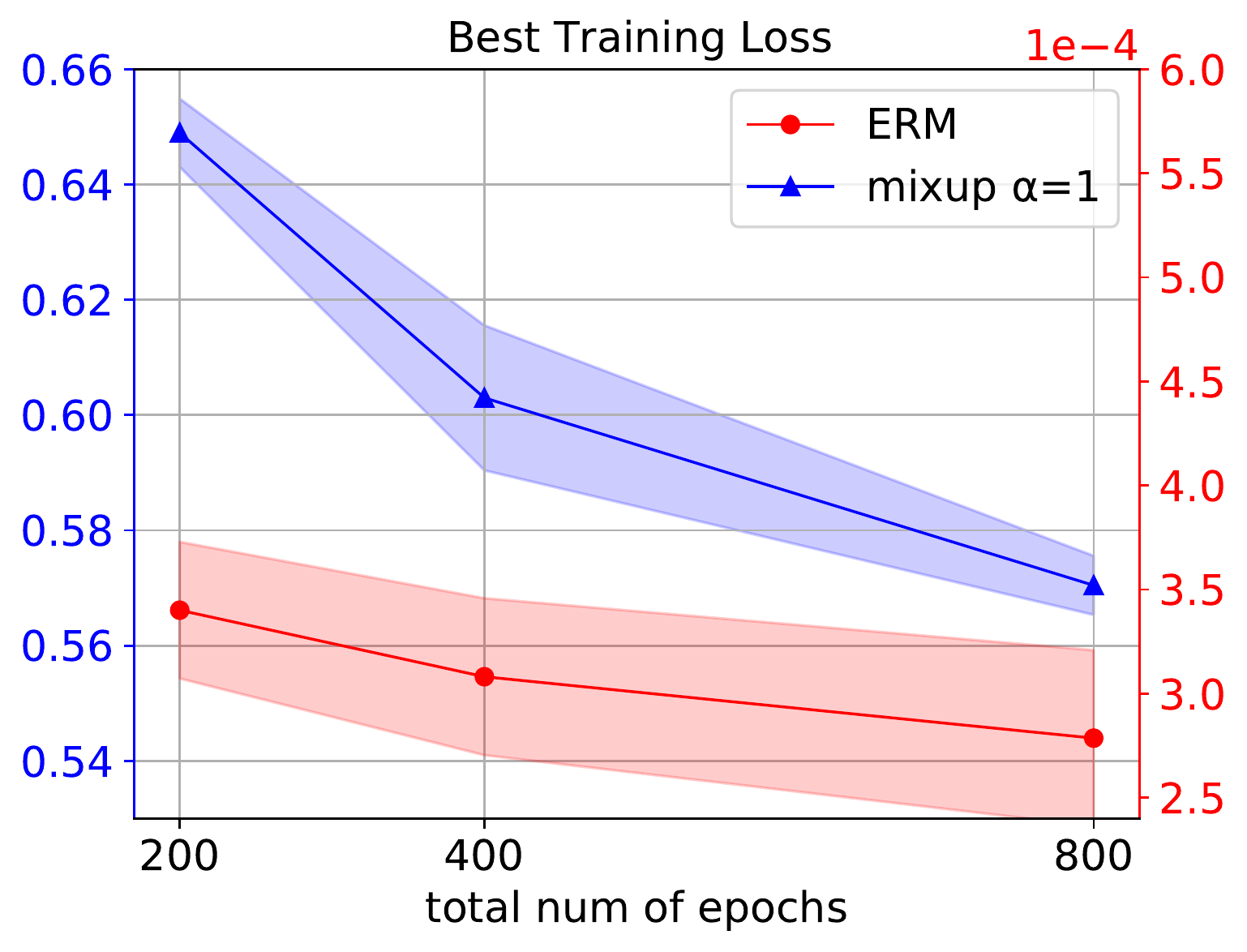}}
       \hfill
    \subfloat[Test acc ($100\%$ {data})]{%
       \includegraphics[width=0.241\linewidth]{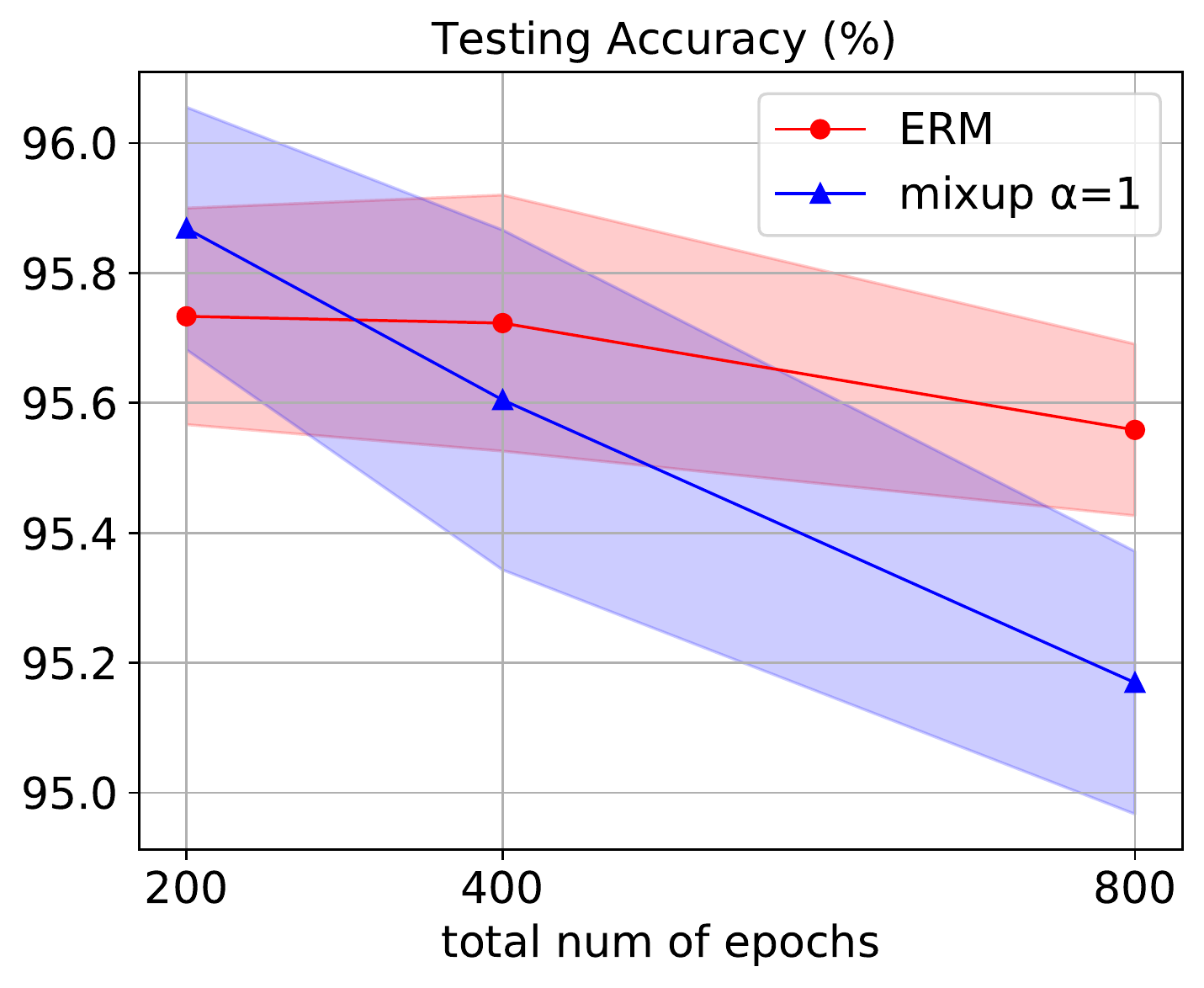}}
    \caption{
    Results of the recorded training losses and testing accuracies of training ResNet34 on SVHN  {training set} ($100\%$ data and $30\%$ data) without data augmentation.
    }
    \label{fig: svhn resnet34 loss & acc curves}
\end{figure}

The results for SVHN are shown in Figure~\ref{fig: svhn resnet34 loss & acc curves}. These results are also in accordance with those of training ResNet18 on both 30\% and 100\% of the SVHN dataset.

\begin{figure}[!ht]
    \centering
    \subfloat[Train loss ($30\%$ data)]{%
       \includegraphics[width=0.262\linewidth]{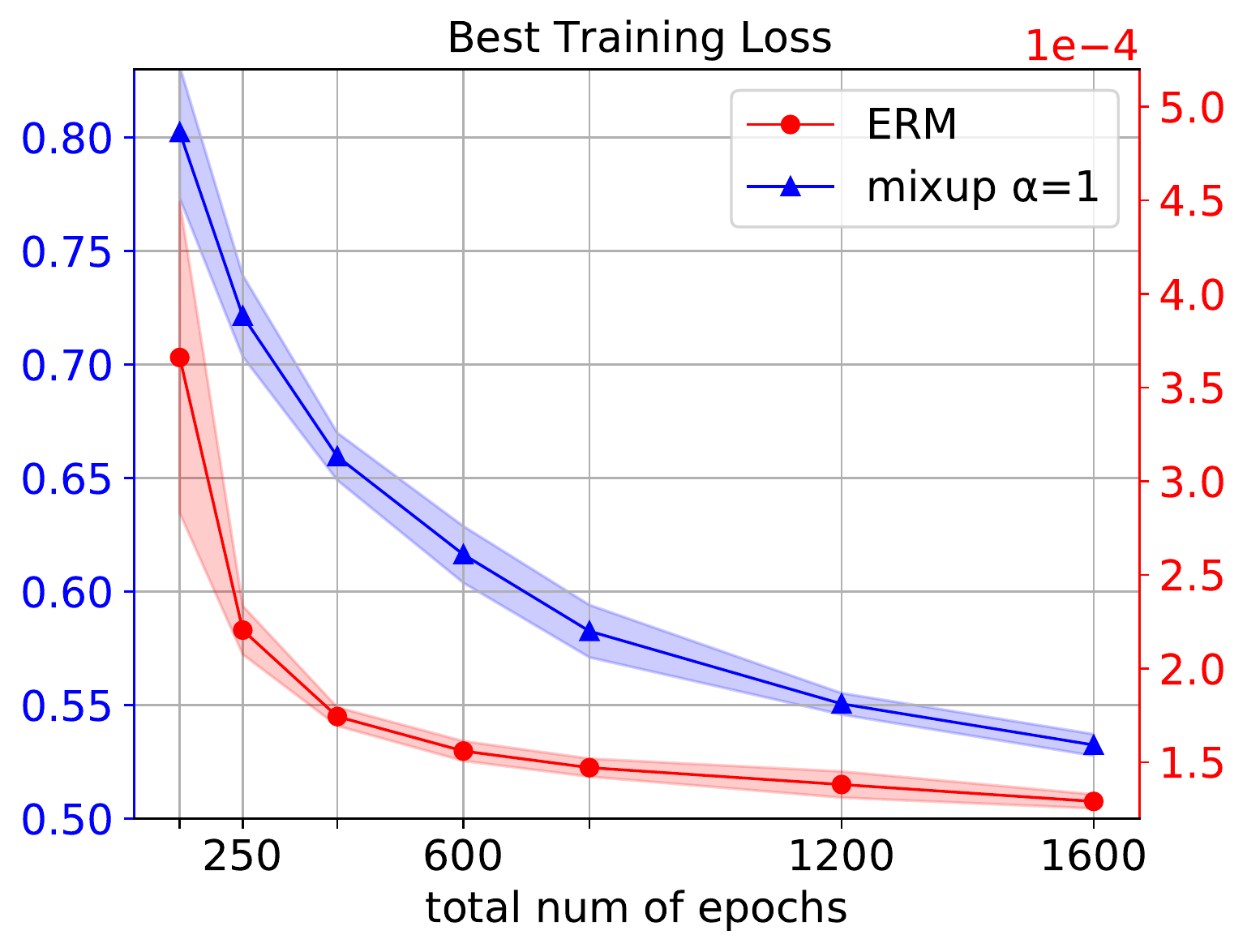}}
       \hfill
    \subfloat[Test acc ($30\%$ data)]{%
       \includegraphics[width=0.238\linewidth]{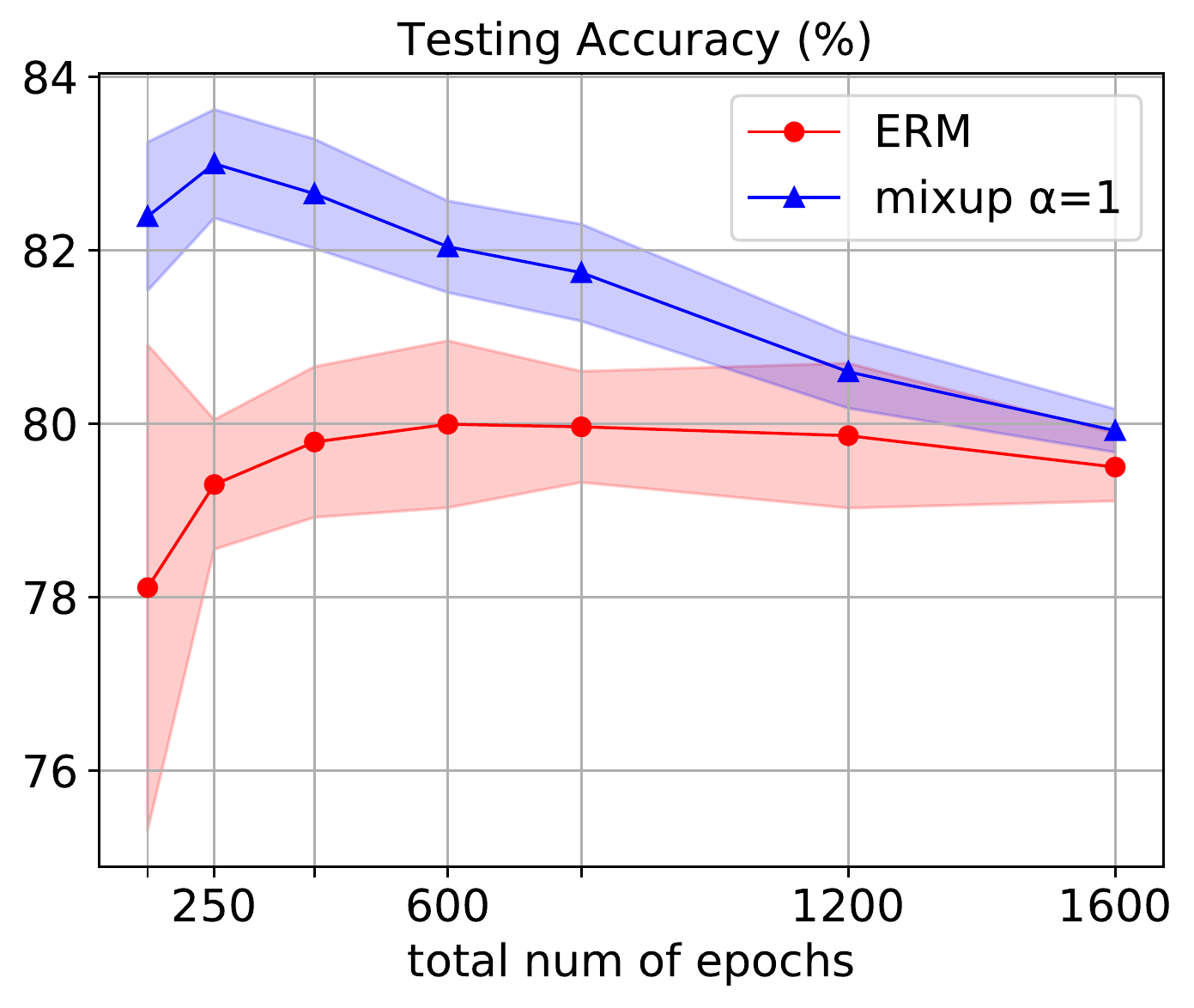}}
       \hfill
    \subfloat[Train loss ($100\%$ data)]{%
       \includegraphics[width=0.262\linewidth]{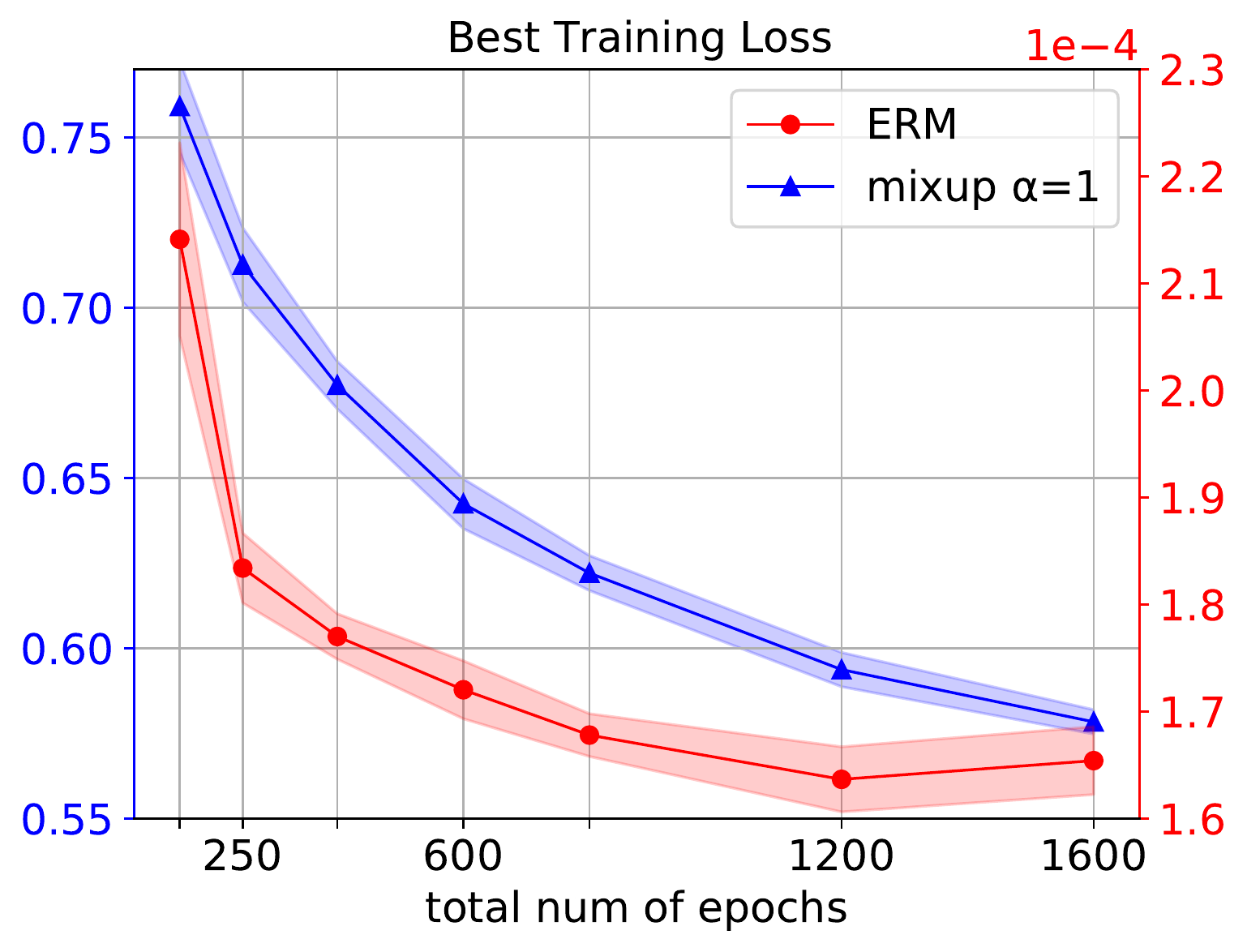}}
       \hfill
    \subfloat[Test acc ($100\%$ data)]{%
       \includegraphics[width=0.238\linewidth]{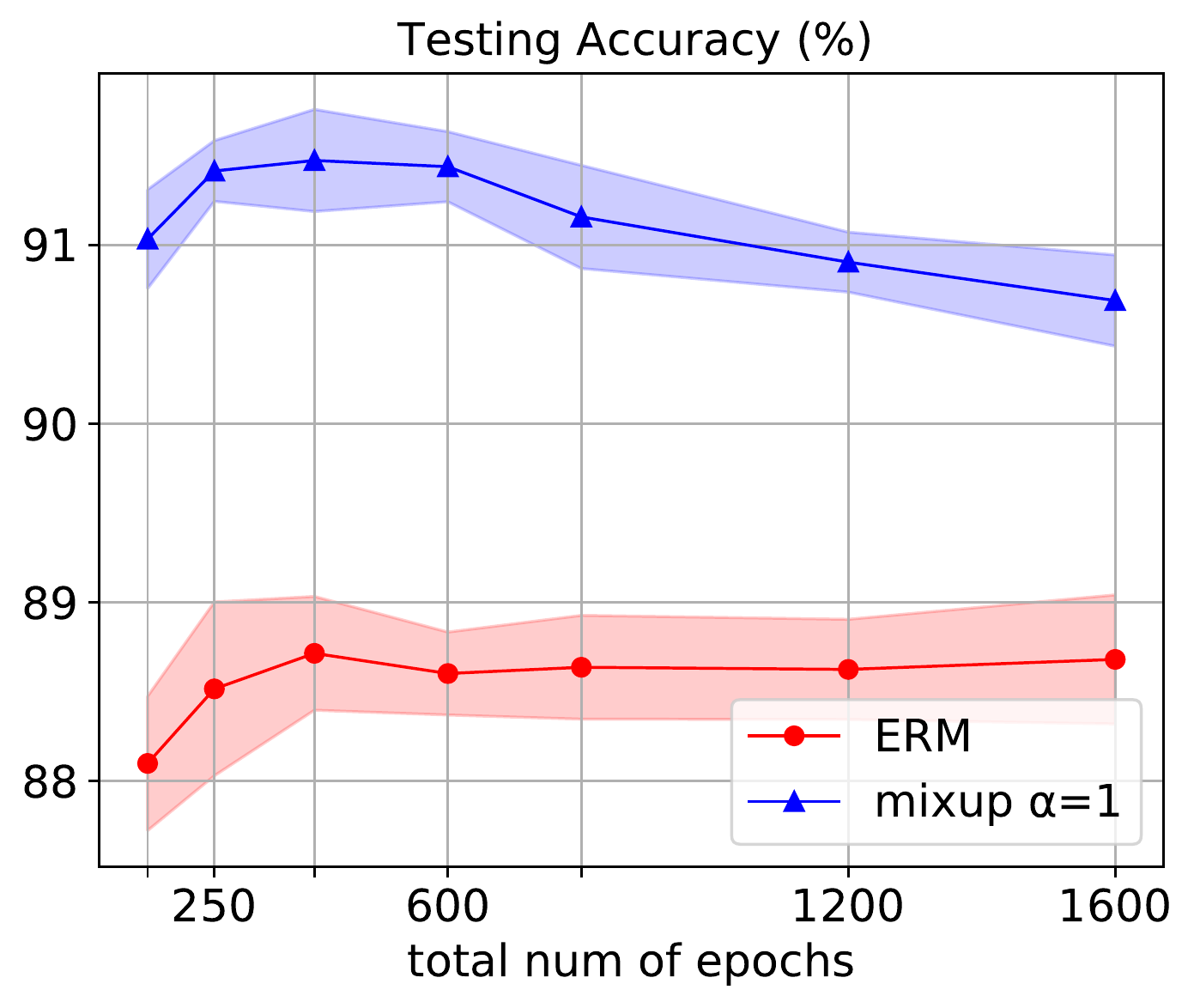}}
    \caption{
    Results of the recorded training losses and testing accuracies of training VGG16 on CIFAR10  training set ($30\%$ data and $100\%$ data) without data augmentation.
    }
    \label{fig: cifar10 vgg16 loss & acc curves}
\end{figure}

In addition, we have trained VGG16 on the CIFAR10 training set (100\% data and 30\% data) for up to in total $1600$ epochs without data augmentation. The results are provided in Figure~\ref{fig: cifar10 vgg16 loss & acc curves}. In both cases, over-training VGG16 with either ERM or Mixup can gradually reduce the best achieved training loss. However, the testing accuracy of the Mixup-trained network also decreases, while that of the ERM-trained network has no significant change.

\subsection{Results of Mean Square Error Loss Without Data Augmentation}
\begin{figure}[!ht]
    \centering
    \subfloat[CIFAR10 ($30\%$)\label{fig:cifar10-30-mse}]{%
       \includegraphics[width=0.25\linewidth]{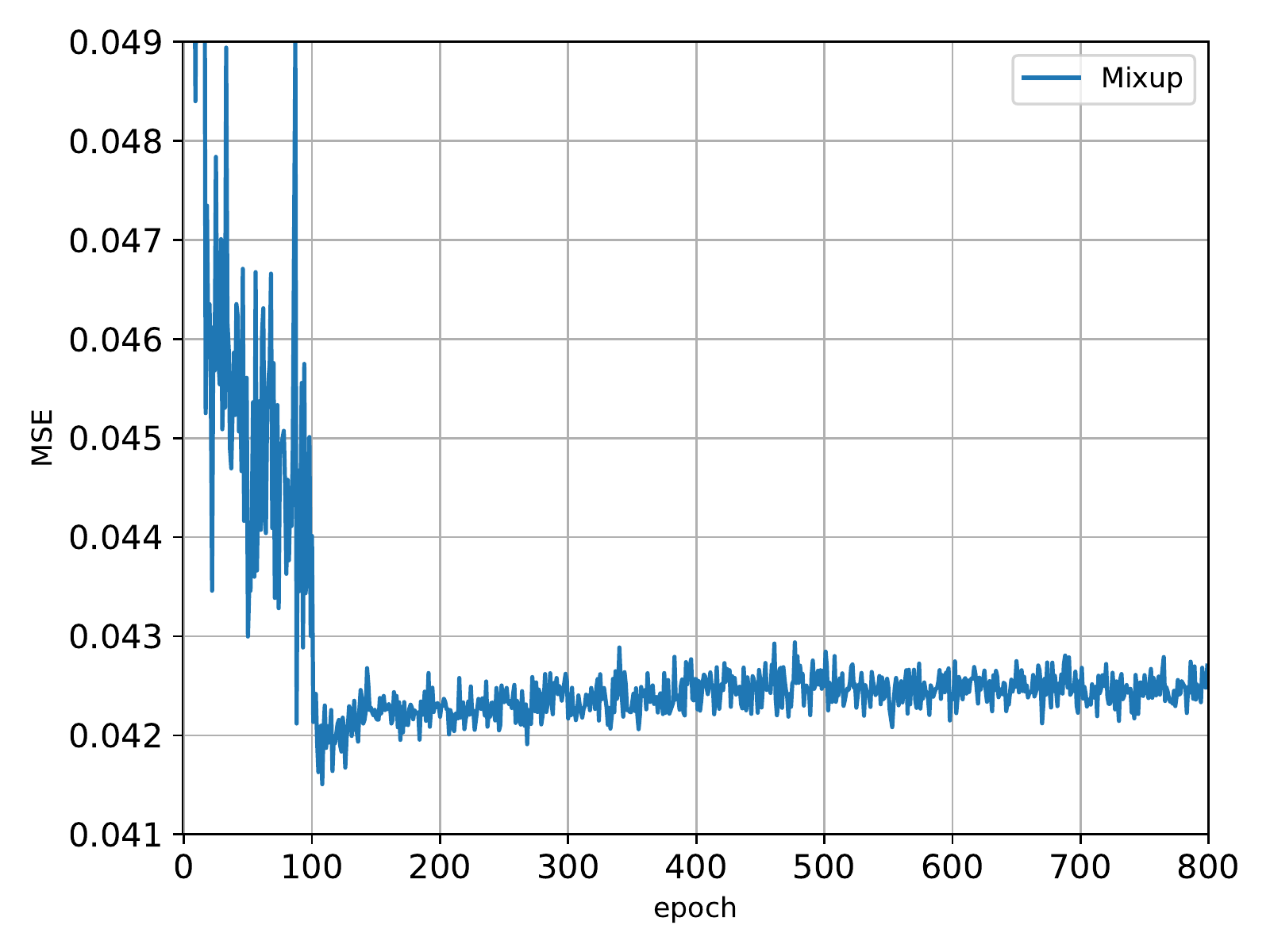}}
    \subfloat[CIFAR10 ($100\%$)\label{fig:cifar10-100-mse}]{%
       \includegraphics[width=0.25\linewidth]{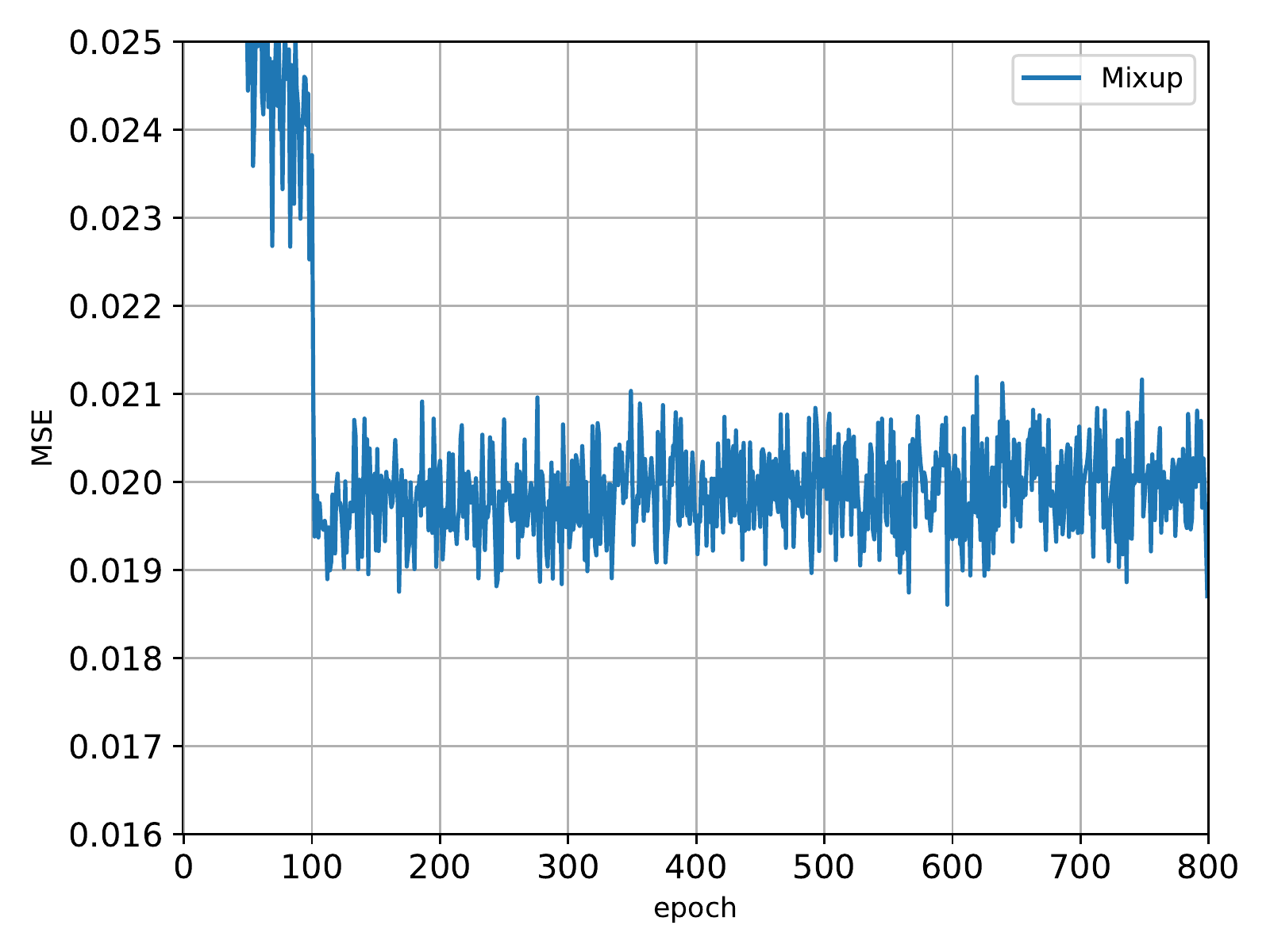}}
    \subfloat[SVHN ($30\%$)\label{fig:svhn-30-mse}]{%
       \includegraphics[width=0.25\linewidth]{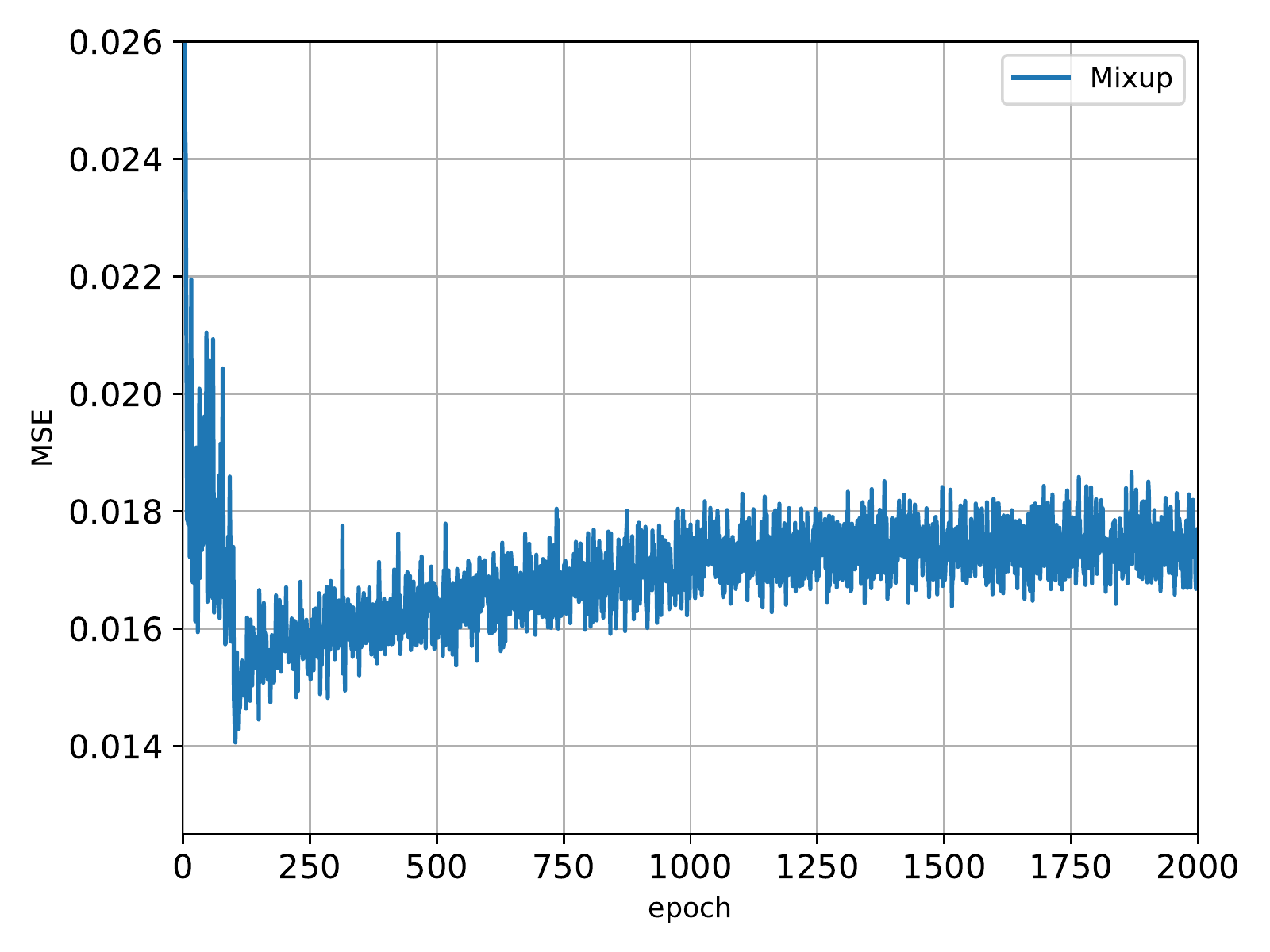}}
    \subfloat[SVHN ($100\%$)\label{fig:svhn-100-mse}]{%
       \includegraphics[width=0.25\linewidth]{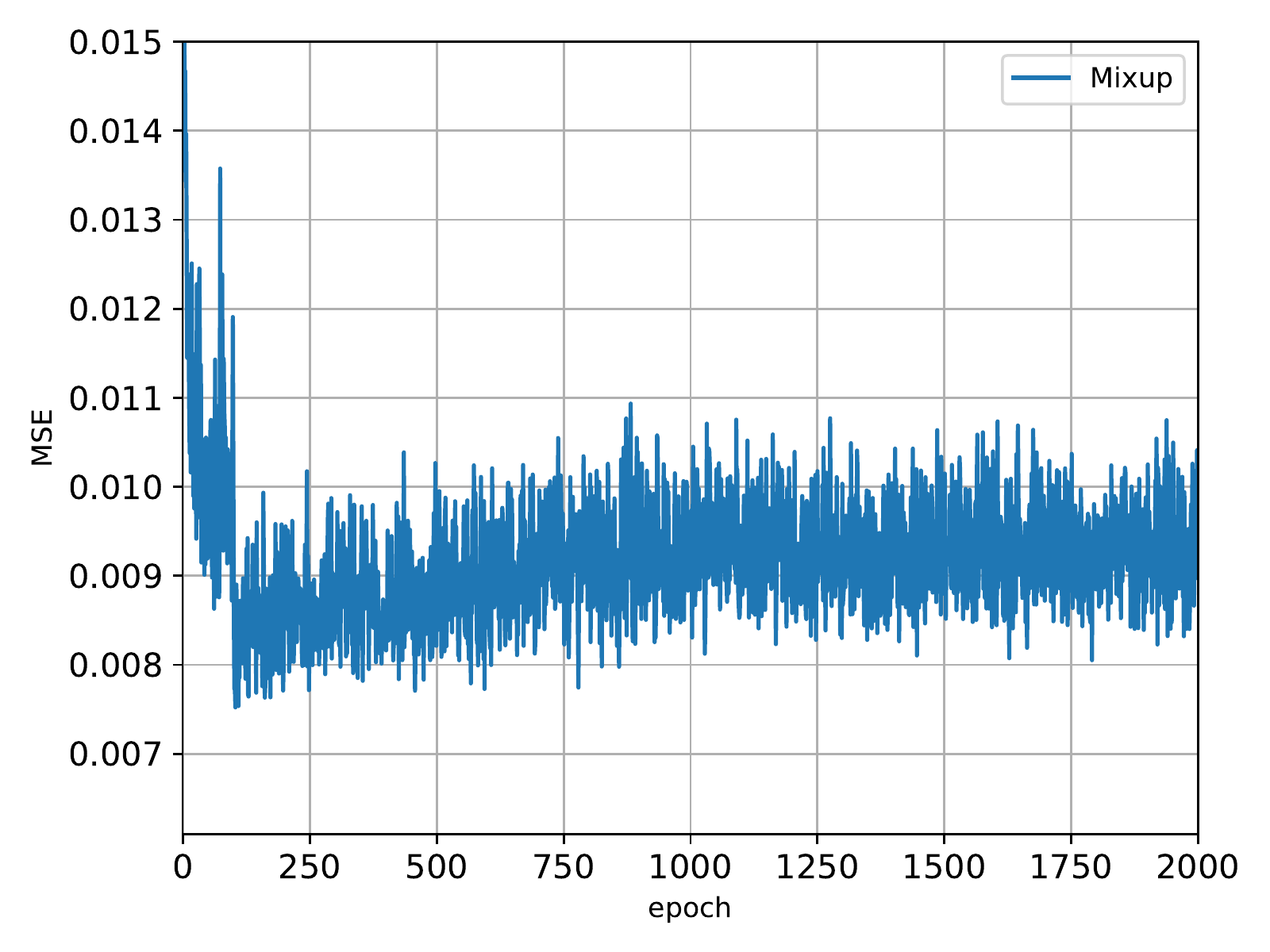}}
    \caption{Dynamics of MSE during Mixup training.}
    \label{fig:Mixu MSE}
\end{figure}

We also perform Mixup training experiments using the mean square error (MSE) loss function on both CIFAR10 and SVHN datasets. Figure~\ref{fig:Mixu MSE} illustrates that the U-shaped behavior observed in previous experiments is also present when using the MSE loss function. To ensure optimal training, the learning rate is decreased by a factor of 10 at epoch 100 and 150.
% We also conduct Mixup training experiments by using mean square error loss on CIFAR10 and SVHN. Figure~\ref{fig:Mixu MSE} shows that the U-shaped behavior also holds for the MSE loss. The learning rate is divided by $10$ at epoch $100$ and $150$.

\subsection{Gradient Norm Vanishes When Changing Mixup to ERM}
\label{sec:mixup-erm-gn}
\begin{figure}[!htbp]
    \centering
    \subfloat[CIFAR10 ($30\%$)\label{fig:cifar10-30-gn}]{%
       \includegraphics[width=0.25\linewidth]{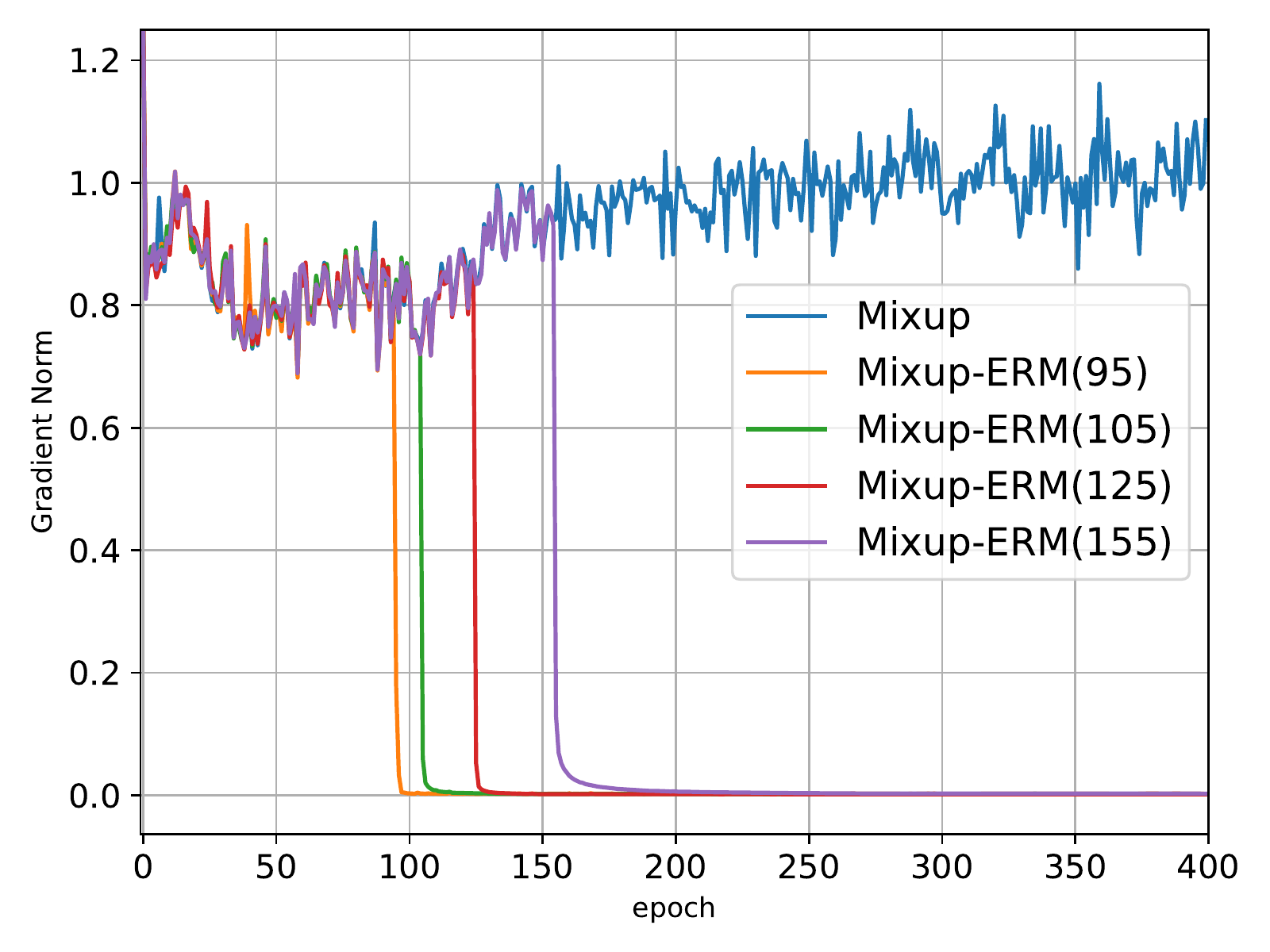}}
    \subfloat[CIFAR10 ($100\%$)\label{fig:cifar10-100-gn}]{%
       \includegraphics[width=0.25\linewidth]{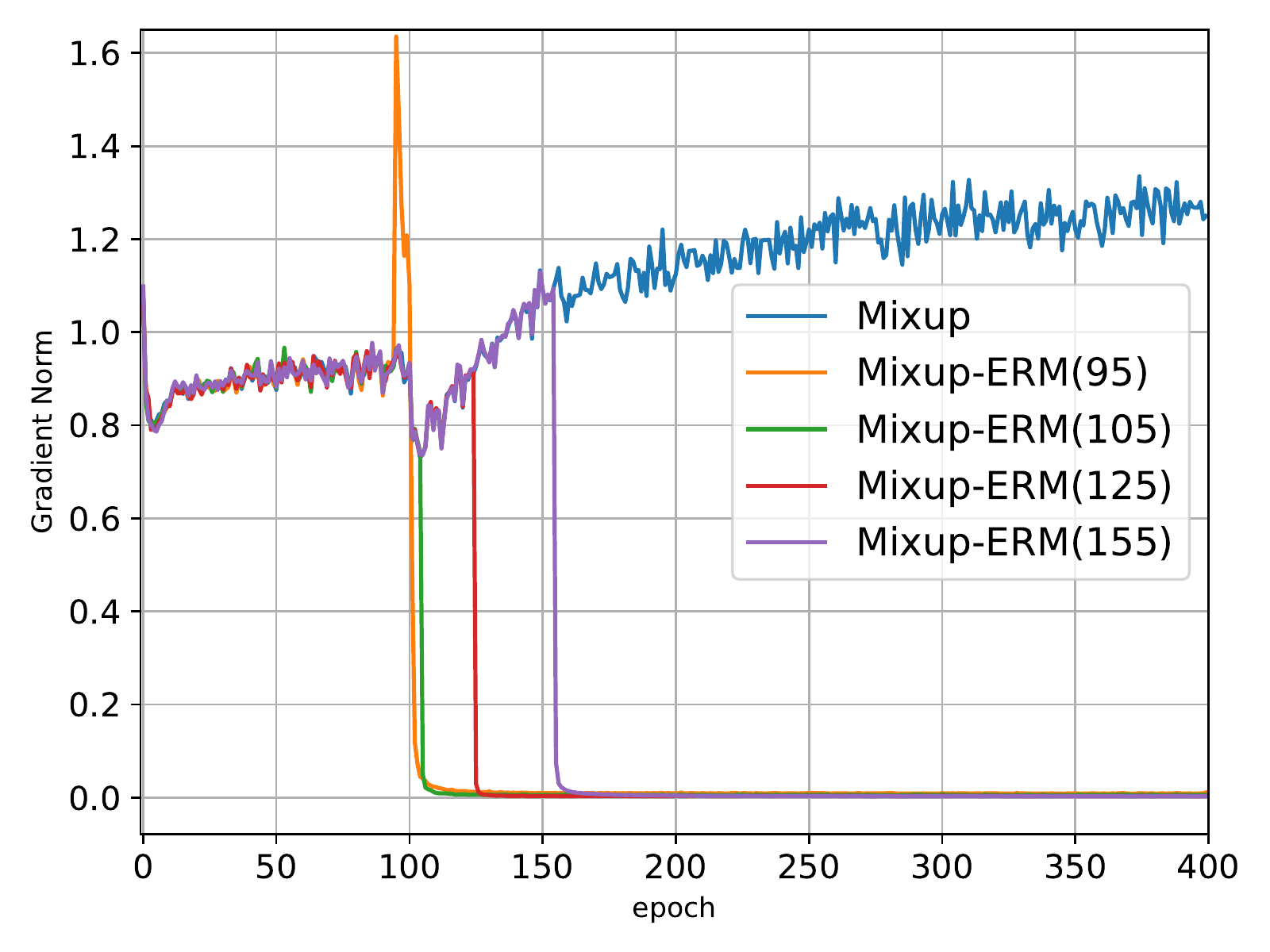}}
    \subfloat[SVHN ($30\%$)\label{fig:svhn-30-gn}]{%
       \includegraphics[width=0.25\linewidth]{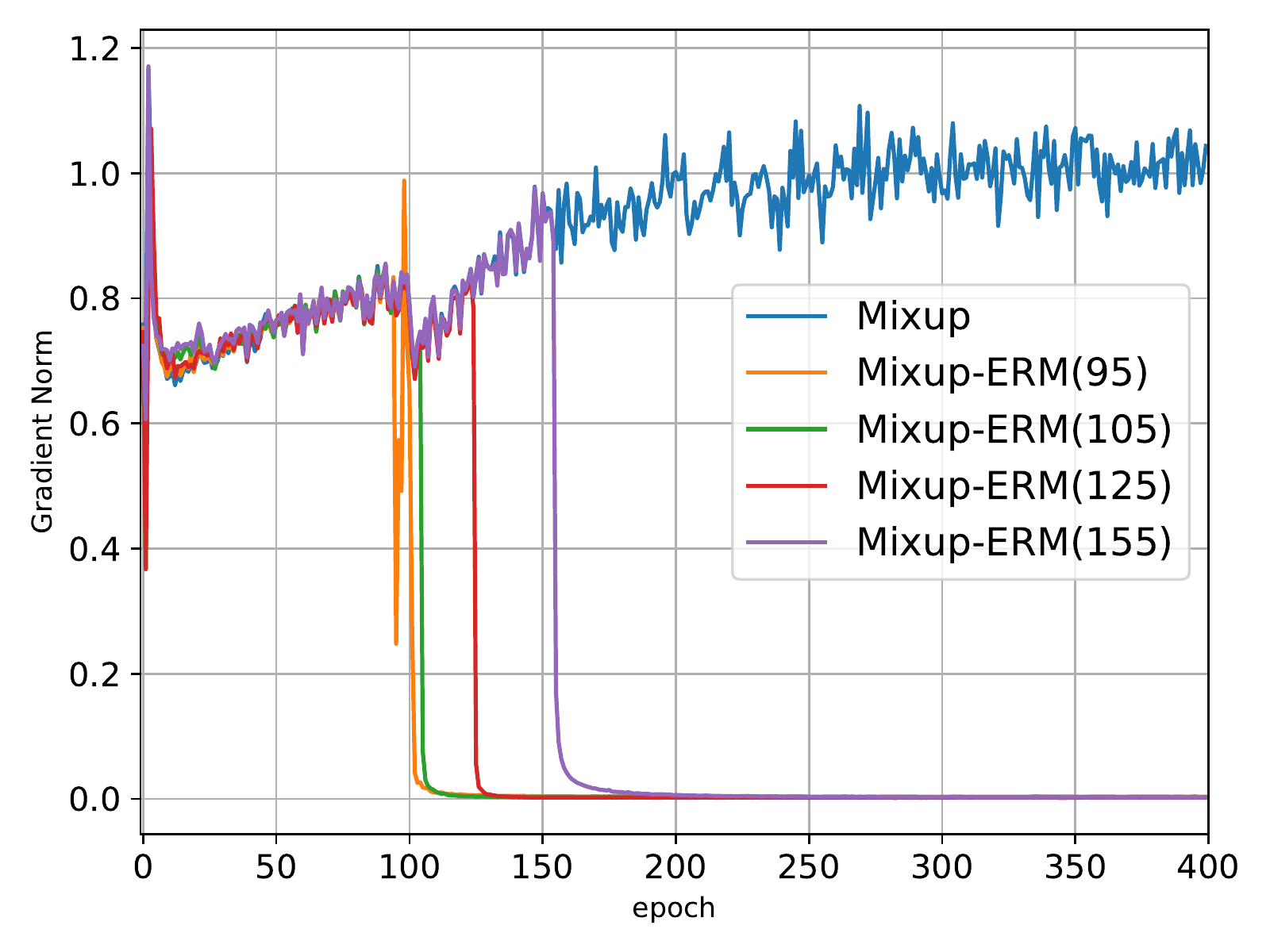}}
    \subfloat[SVHN ($100\%$)\label{fig:svhn-100-gn}]{%
       \includegraphics[width=0.25\linewidth]{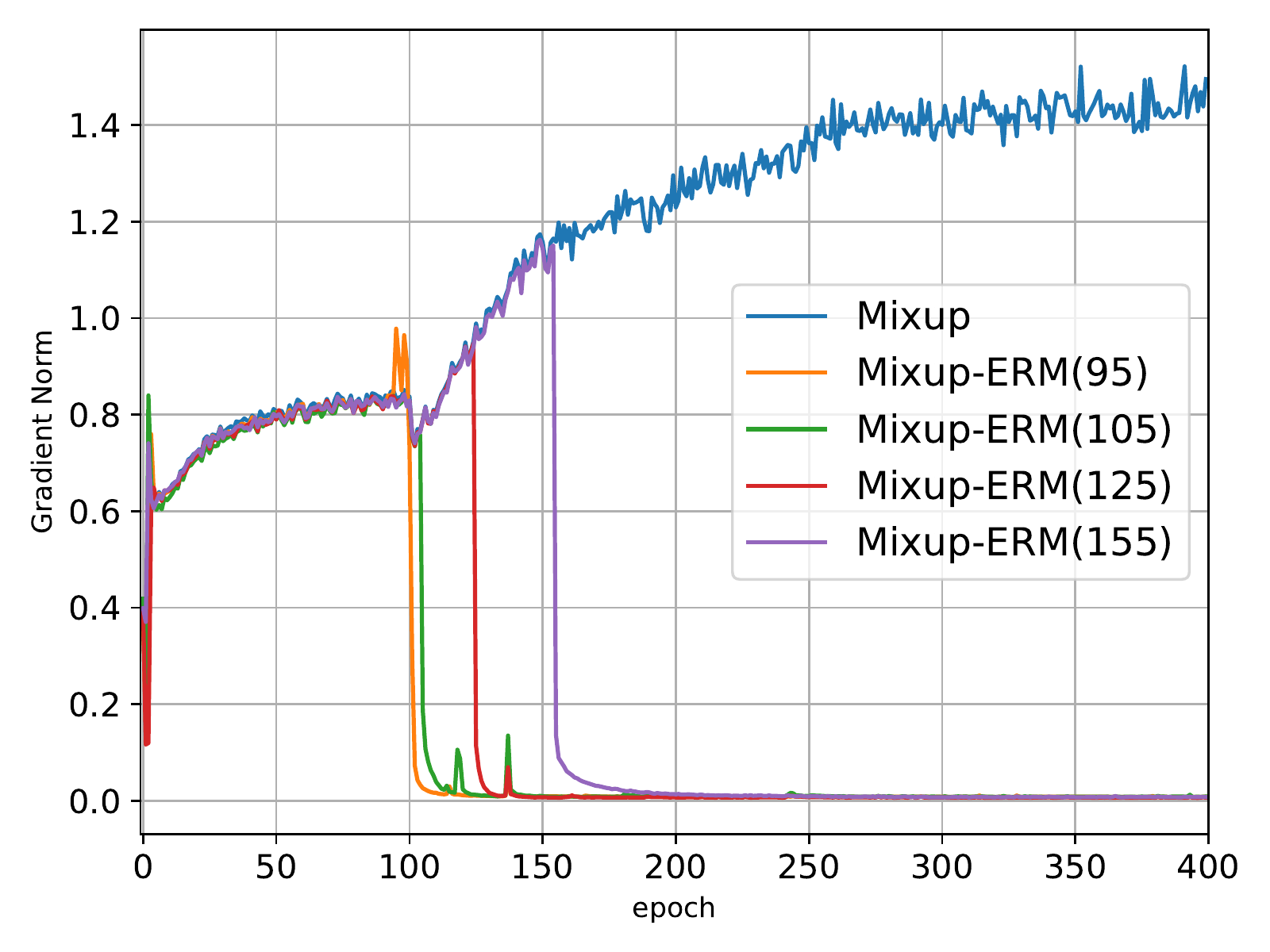}}
    \caption{Dynamics of gradient norm when changing Mixup training to ERM training.}
    \label{fig:Mixup-switch-norm}
\end{figure}

In Figure~\ref{fig:gradient-norm}, we can observe that the gradient norm of Mixup training does not diminish at the end of training and can even explode to a very high value. In contrast, ERM results in a gradient norm of zero at the end of training. Figure~\ref{fig:Mixup-switch-norm} illustrates that when switching from Mixup training to ERM training after a certain period, the gradient norm will rapidly become zero. This phenomenon occurs because Mixup-trained neural networks have already learned the ``clean patterns'' and the original data does not provide any useful gradient signal. Therefore, this further supports the idea that the latter stage of Mixup training is primarily focused on memorizing noisy data.
% In Figure~\ref{fig:gradient-norm}, we know that the gradient norm of Mixup training will not vanish at the end of training and will even explode to a very high value. In contrast, ERM will have zero gradient norm at the end. Figure~\ref{fig:Mixup-switch-norm} shows that the gradient norm will instantly become zero. This is because the ``clean patterns'' are already learned by Mixup trained neural networks and the original data will not provide any useful gradient signal. This further justifies that the latter stage of Mixup training is only for memorizing noisy data.

\subsection{Validation Results on Covariance-Shift Datasets}
\begin{figure}[!htbp]
    \centering
    \subfloat[CIFAR10.1]{%
       \includegraphics[width=0.252\linewidth]{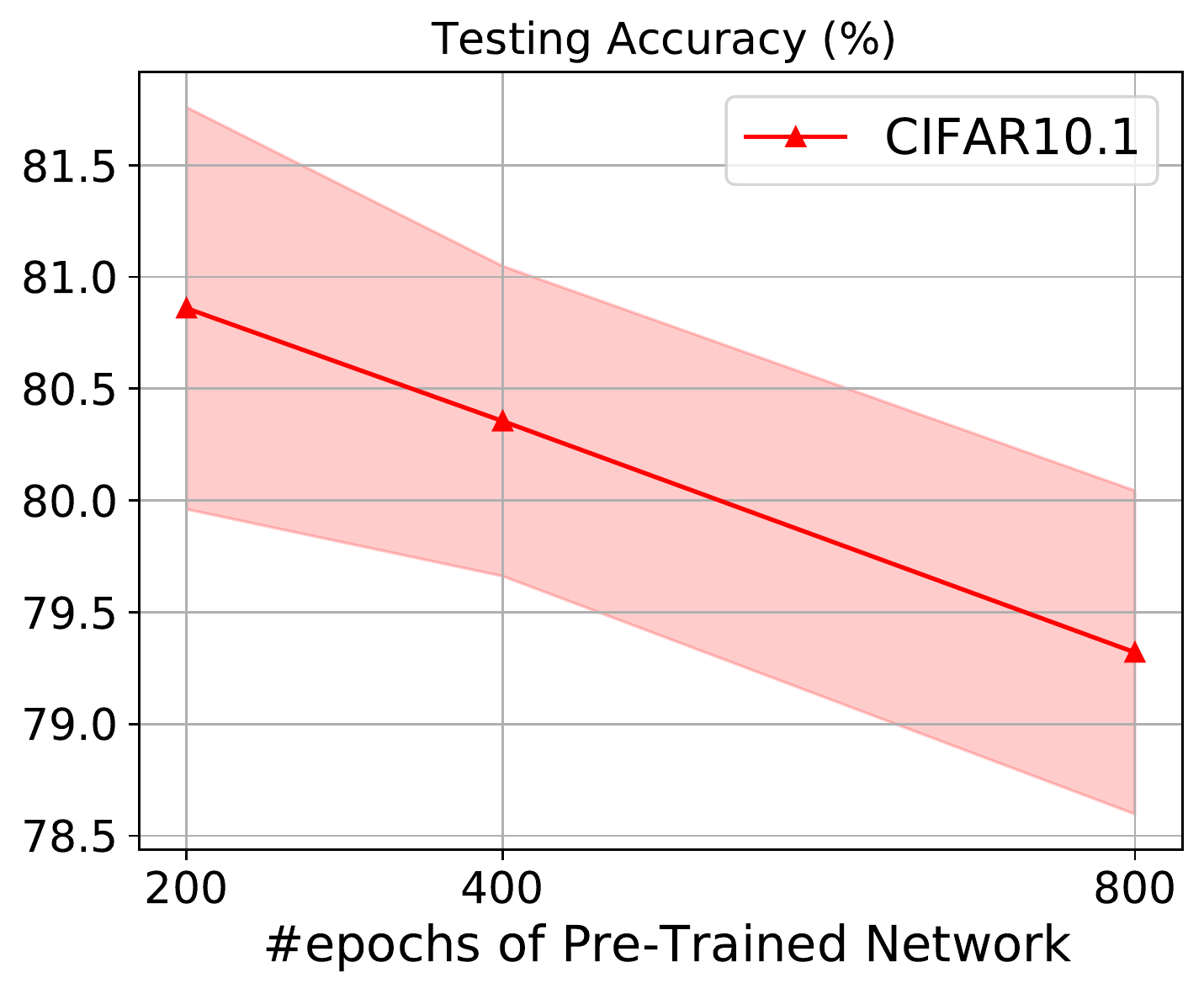}}
    \subfloat[CIFAR10.2]{%
       \includegraphics[width=0.252\linewidth]{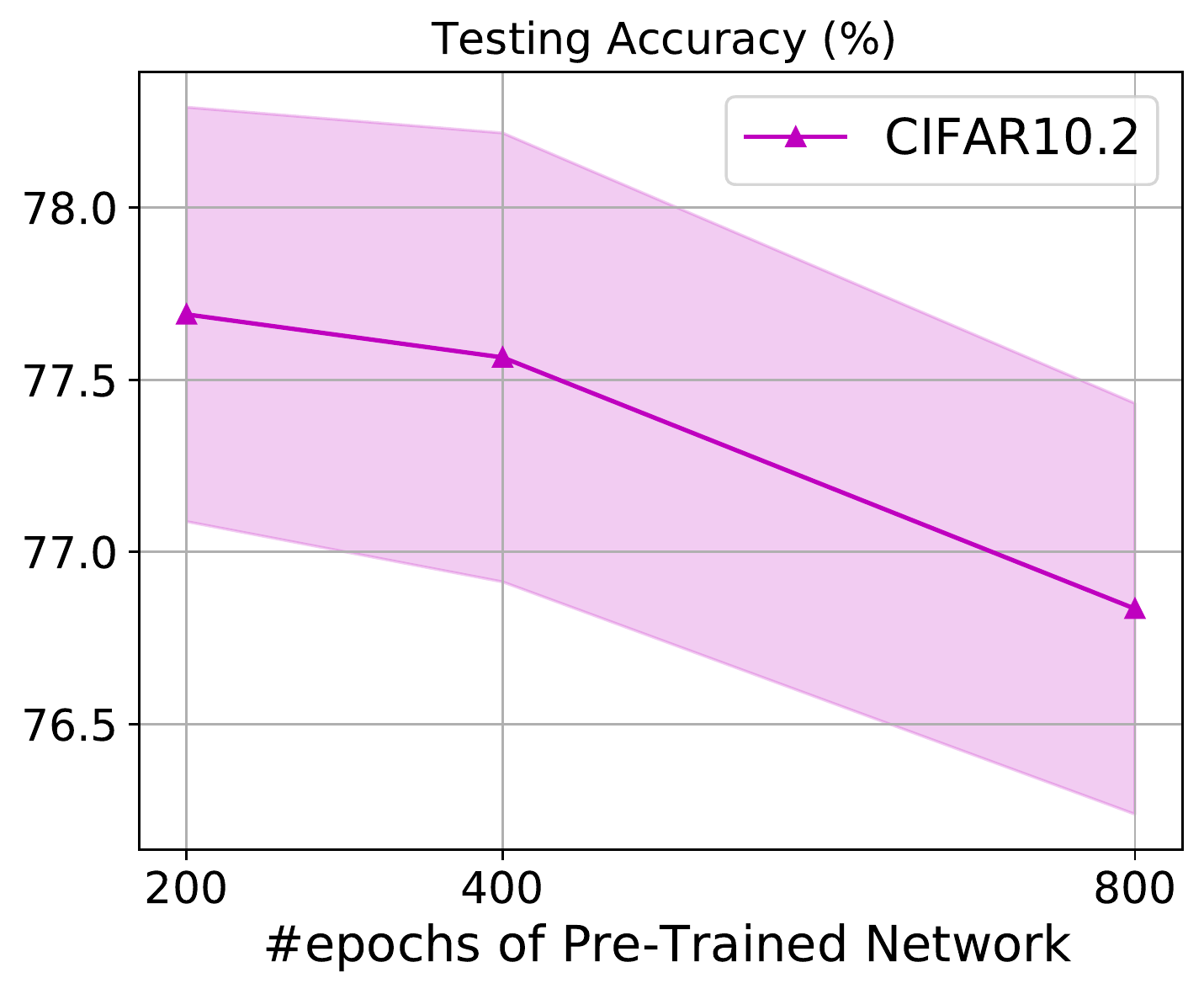}}
    \subfloat[CIFAR10-C-1]{%
       \includegraphics[width=0.252\linewidth]{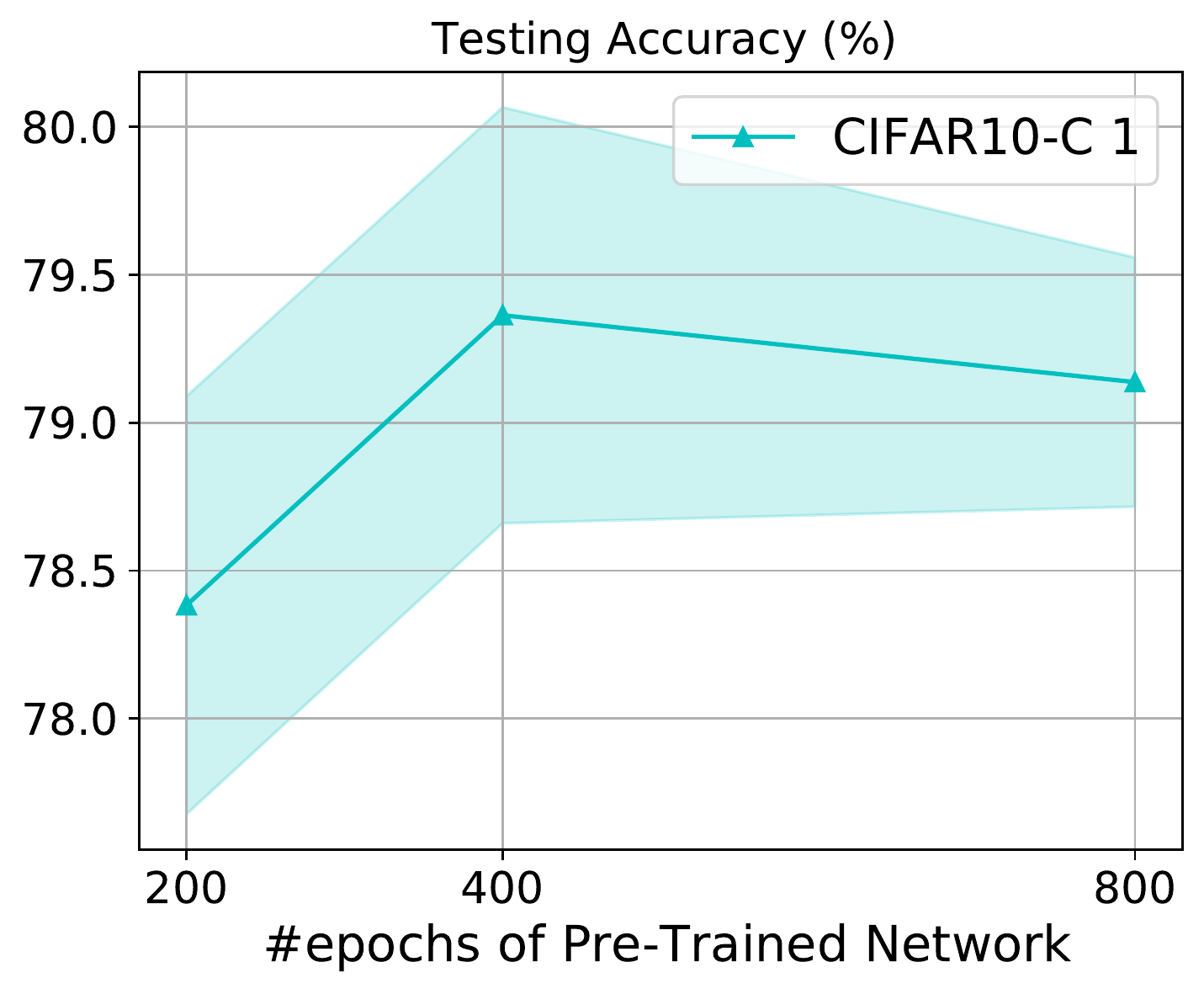}}
    \subfloat[CIFAR10-C-5]{%
       \includegraphics[width=0.244\linewidth]{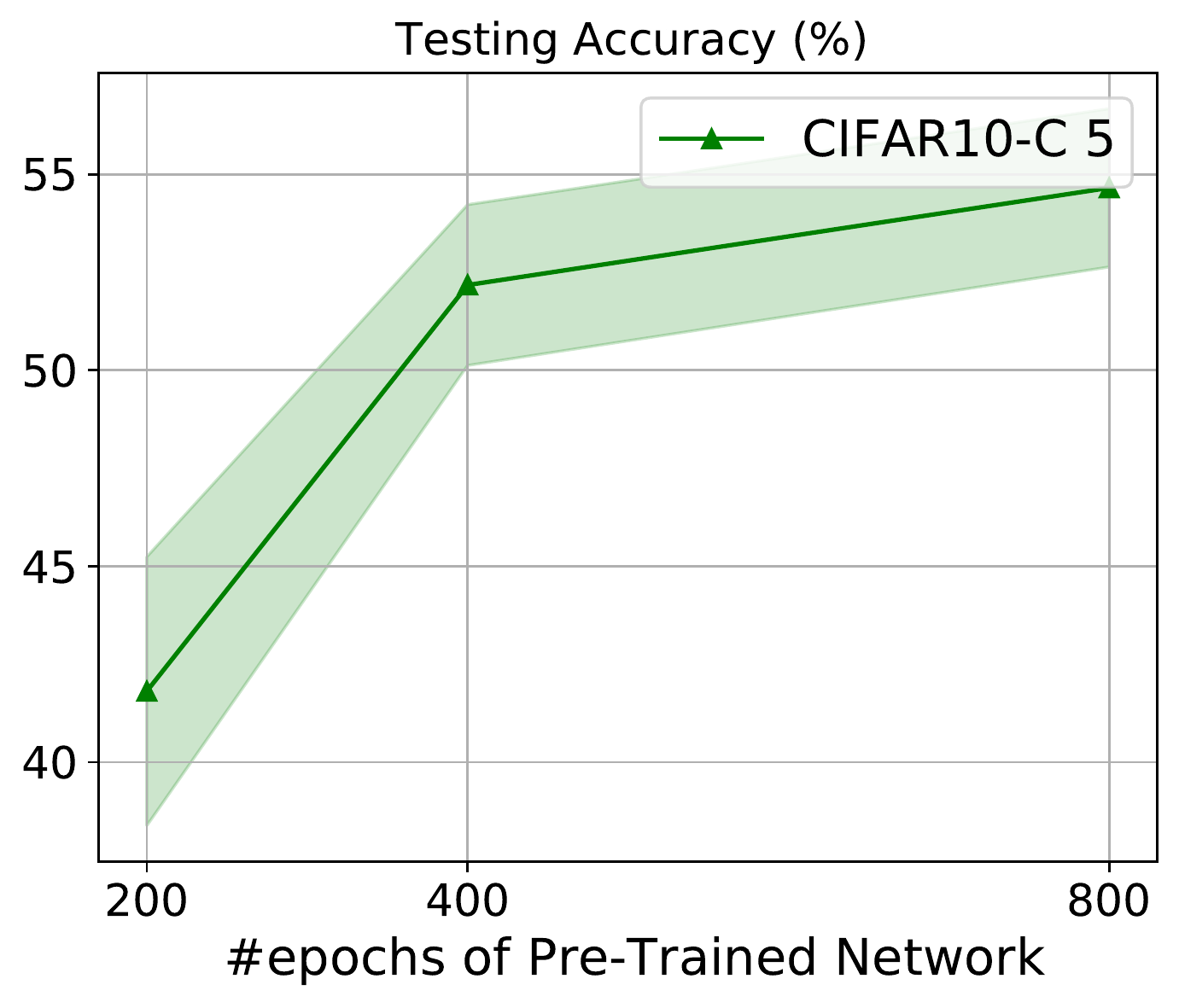}}
    \caption{Models Pre-Trained on 100\% CIFAR10 (without data augmentation)}
    \label{fig:covariate shift pretrain on 1.0 cifar10}
\end{figure}

\begin{figure}[!htbp]
    \centering
    \subfloat[CIFAR10.1]{%
       \includegraphics[width=0.25\linewidth]{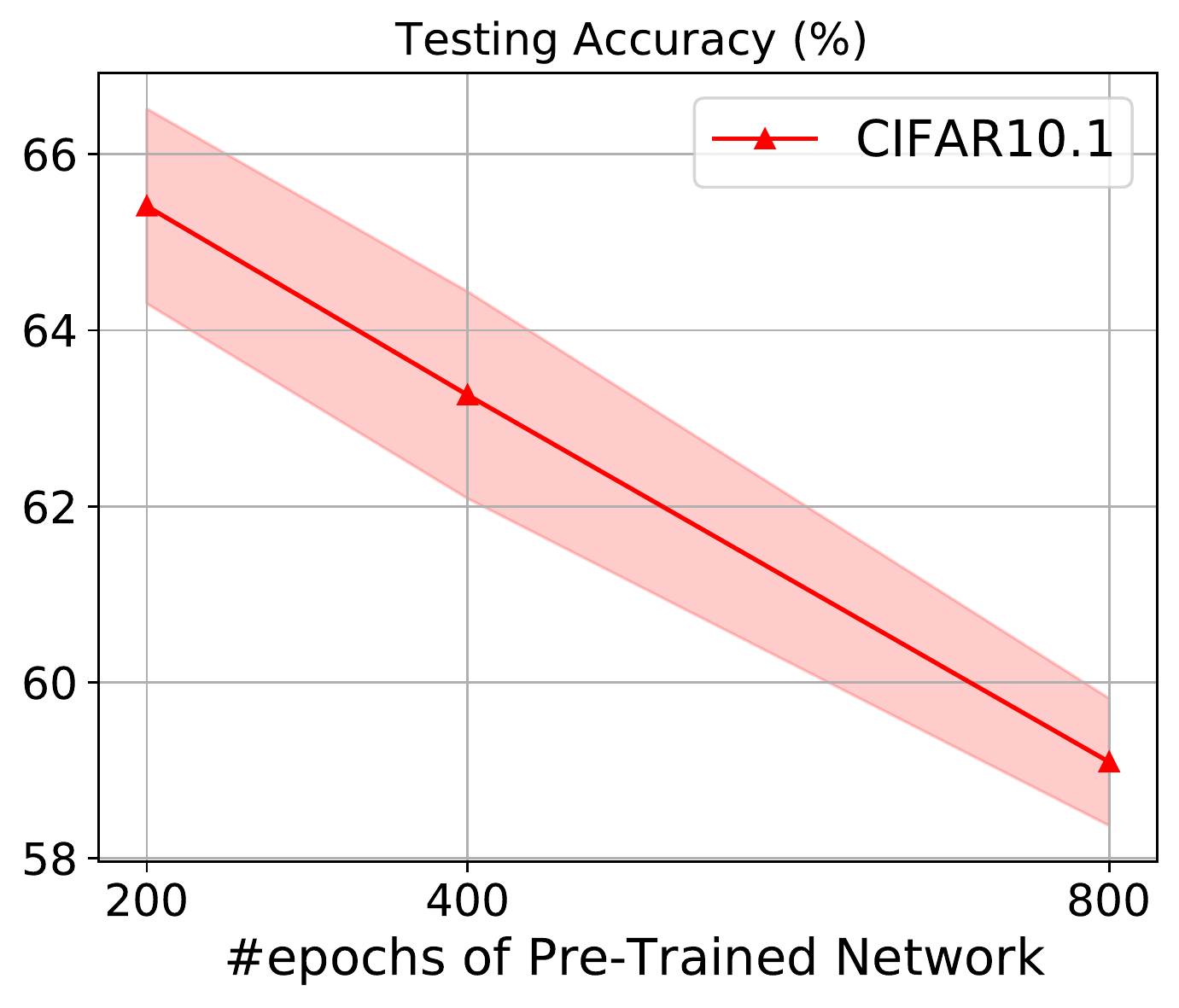}}
    \subfloat[CIFAR10.2]{%
       \includegraphics[width=0.25\linewidth]{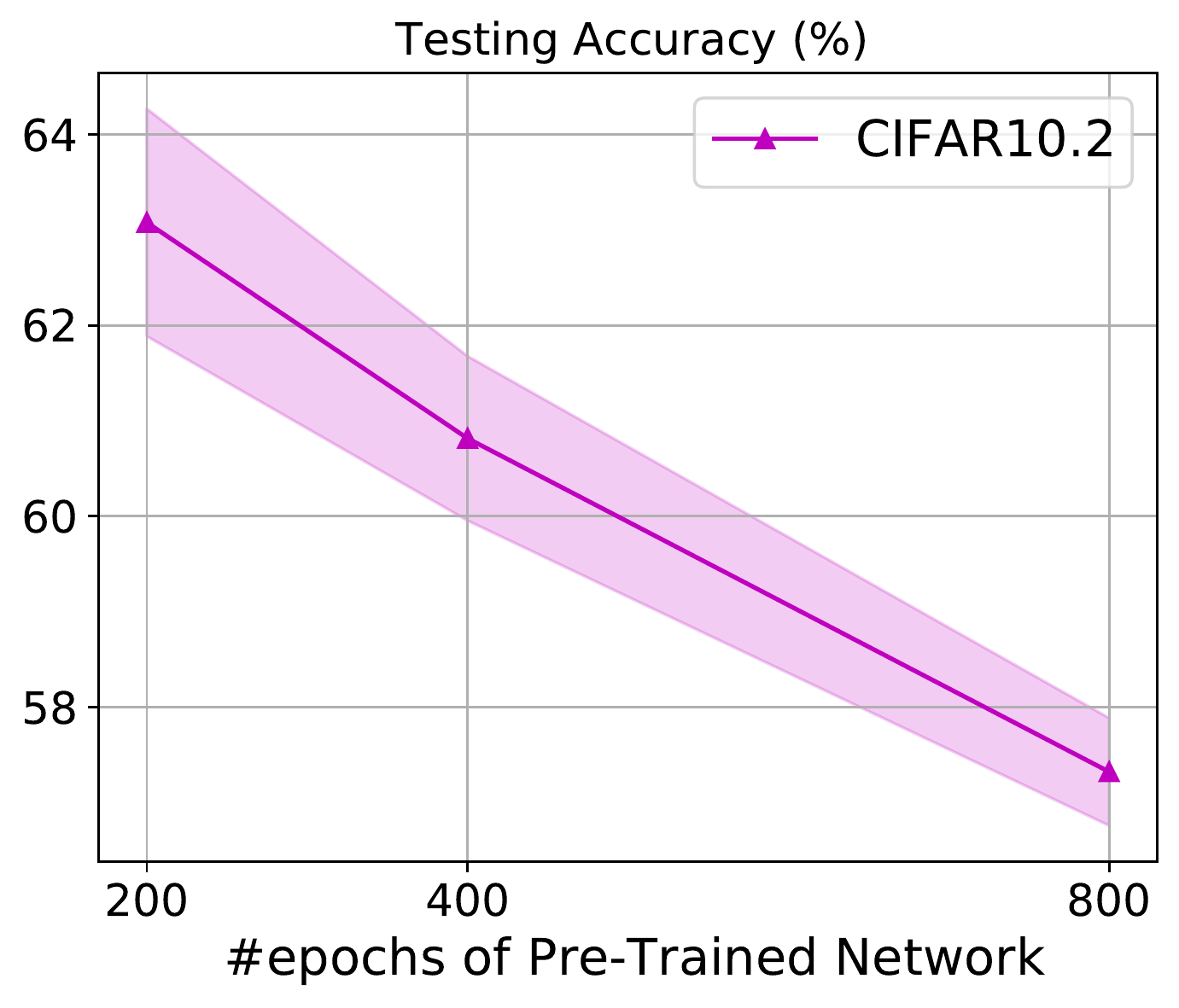}}
    \subfloat[CIFAR10-C-1]{%
       \includegraphics[width=0.25\linewidth]{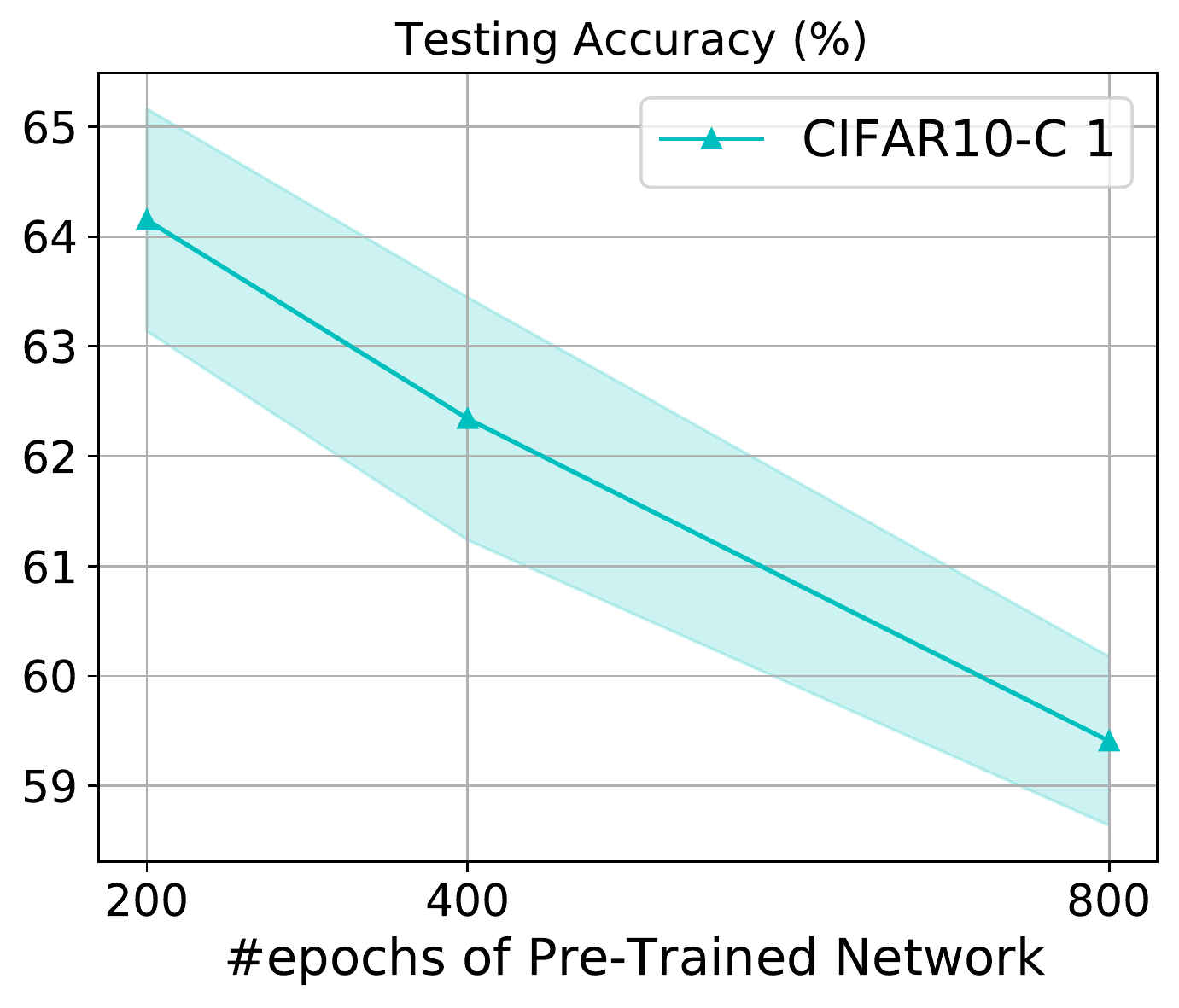}}
    \subfloat[CIFAR10-C-5]{%
       \includegraphics[width=0.25\linewidth]{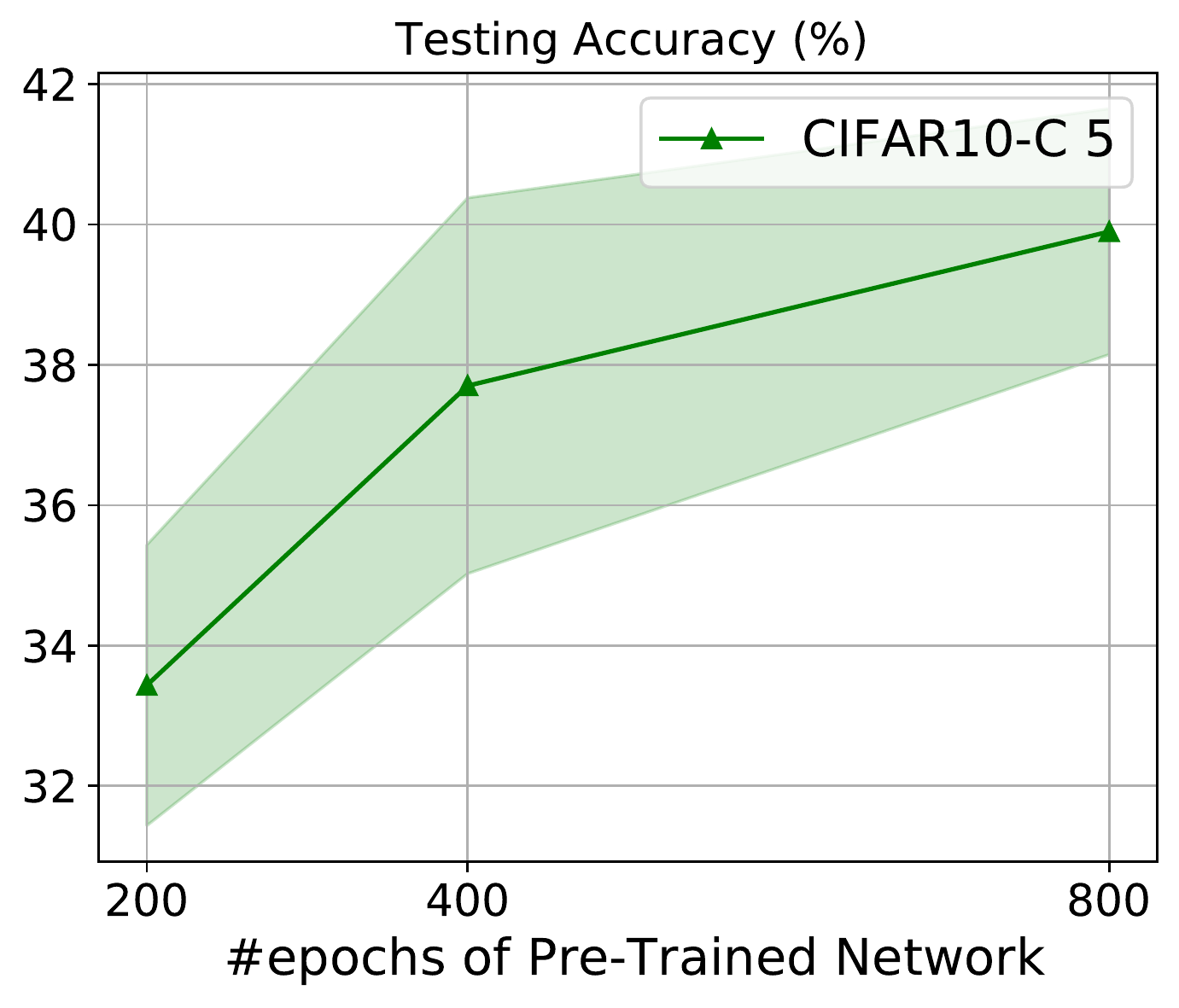}}
    \caption{Models Pre-Trained on 30\% CIFAR10 (without data augmentation)}
    \label{fig:covariate shift pretrain on 0.3 cifar10}
\end{figure}

Recall Figures~\ref{fig: CIFAR10 loss & acc curves} and \ref{fig: SVHN loss & acc curves}, we have seen that as we increase the training epochs ($200\rightarrow400\rightarrow800$), the local minima on the loss landscape (measure by the real training data) of the Mixup-trained model gradually becomes sharper. To validate the regular pattern of the relationship between the minima flatness and the generalization behavior on the covariate-shift datasets, we have ran some of the Mixup-trained ResNet18 networks and tested their accuracies on CIFAR10.1~\citep{recht2018cifar10.1}, CIFAR10.2~\citep{lu2020harder} and CIFAR10-C~\citep{hendrycks2019robustness} using Gaussian noise with severity 1 and 5 (denoted by CIFAR10-C-1 and CIFAR10-C-5). The results of the models pre-trained on 100\% CIFAR10 are given in Figure~\ref{fig:covariate shift pretrain on 1.0 cifar10}, and the results of the models pre-trained on 30\% CIFAR10 are given in Figure~\ref{fig:covariate shift pretrain on 0.3 cifar10}.

From the results, it is seen that with training epochs increase, the testing performance on the models on CIFAR10.1 and CIFAR10.2 decreases, taking a similar trend as our results in standard testing sets (i.e., the original CIFAR10 testing sets without covariate shift.) But on CIFAR10-C, this behaviour is not observed. In particular, the performance on CIFAR10-C-5 continues to improve over the training iterations. This seems to suggest that the flatness of empirical-risk loss landscape may impact generalization to covariate-shift datasets in more complex ways, possibly depending on the nature and structure of the covariate shift.
% \vspace{-5mm}
\subsection{Investigation of the Impact of RegMixup in Over-Training}
\begin{figure}[!htbp]
    \centering
       \includegraphics[width=0.6\linewidth]{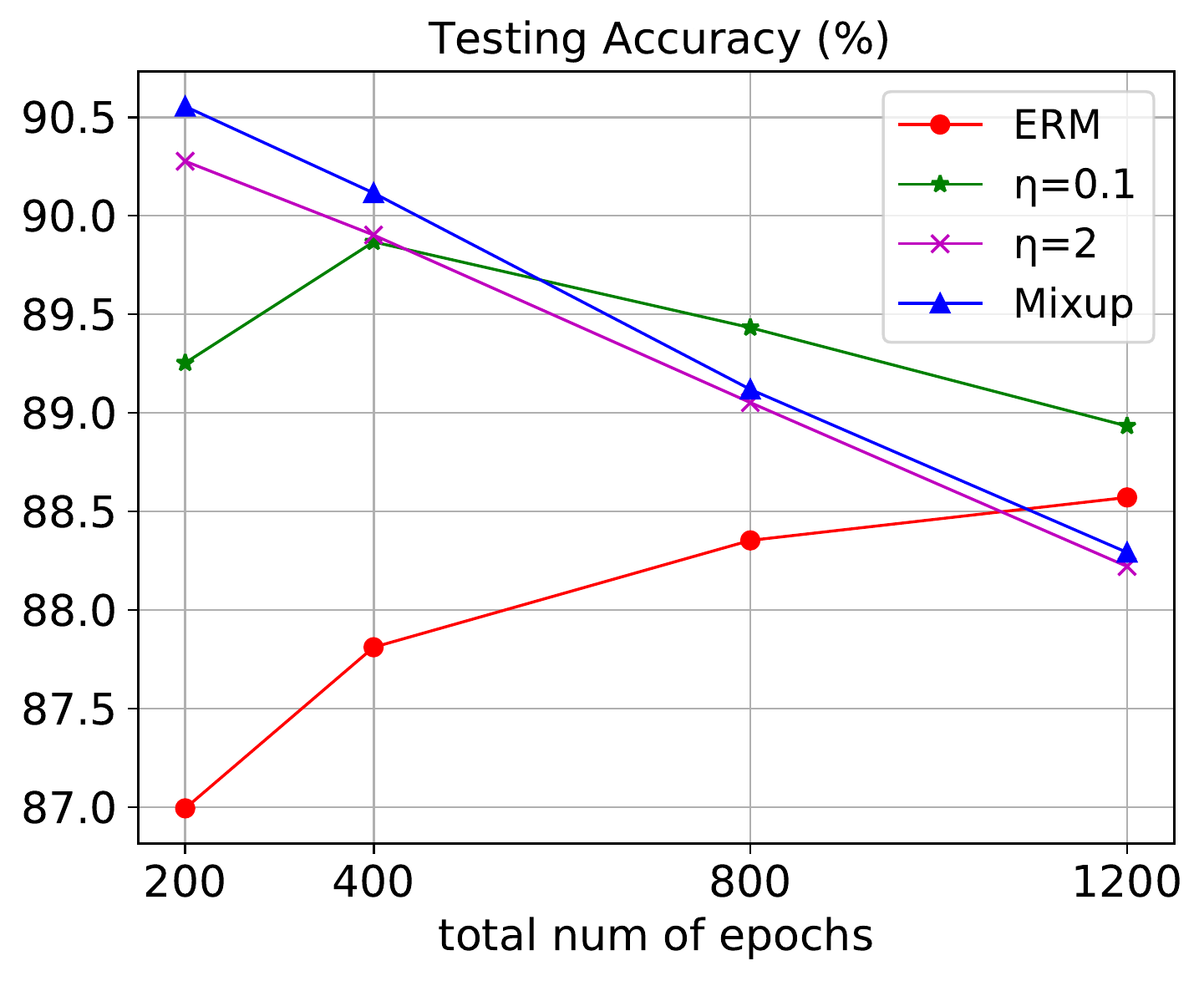}
    \caption{Results of Over-Training ResNet18 on CIFAR10 (without data augmentation) with RegMixup. Green: $\eta=0.1$; Purple: $\eta=2$; Red: ERM; Blue: standard Mixup.}
    \label{fig: regmixup overtrain}
\end{figure}

RegMixup is a variant algorithm of Mixup proposed by \citep{pinto2022regmixup}. For each synthetic example $(\tilde{\textbf{x}},\tilde{\textbf{y}})$ formulated by $(\textbf{x},\textbf{y})$ and $(\textbf{x}',\textbf{y}')$, RegMixup minimizes the following loss 
\[
\ell_{\text{CE}}(\theta,\textbf{x},\textbf{y})+\eta\ell_{\text{CE}}(\theta,\tilde{\textbf{x}},\tilde{\textbf{y}})
\]
where $\ell_{\text{CE}}(\cdot)$ denotes the cross-entropy loss, and $\eta$ is non-negative. The authors show that RegMixup can improve generalization on both in-distribution and covariate-shift datasets, and that it can also improve the out-of-distribution robustness. To validate the performance of RegMixup in the over-training scenario, we have trained ResNet18 using RegMixup with a few different settings of $\eta$. The network is trained on CIFAR10 without data augmentation for up to in total 1200 epochs. The results are given in Figure~\ref{fig: regmixup overtrain}.

The results show that when $\eta=2$, RegMixup performs nearly identically as standard Mixup in the over-training scenario. When $\eta=0.1$, RegMixup postpones the presenting of the turning point, and in the large epochs it outperforms standard Mixup. However, the phenomenon that the generalization performance of the trained model degrades with over-training still exists.

% \subsection{Dynamics of Original Training Loss in Mixup Training}
\section{Experiment Settings for the Teacher-Student Toy Example}
We set the dimension of the input feature as $d_0=10$. The teacher network consists of two layers with the activation function \textbf{Tanh}, and the hidden layer has a width of $5$. Similarly, the student network is a two-layer neural network with \textbf{Tanh}, where we train only the second layer and keep the parameters in the first layer fixed. The hidden layer has a dimension of $100$ (i.e. $d=100$).

To determine the value of $\lambda$, we either draw from a \textbf{Beta(1,1)} distribution in each epoch or fix it to $0.5$ in each epoch. We choose $n=20$, which puts us in the overparameterized regime where $n<d$, and the underparameterized regime where $m\geq n^2>d$. The learning rate is set to $0.1$, and we use full-batch gradient descent to train the student network with MSE. Here, the term "full-batch" means that the batch size is equal to $n$, enabling us to compare the fixed $\lambda$ and random $\lambda$ methods fairly.

For additional information, please refer to our code.
% The dimension of the input feature $d_0=10$. The teacher network is a two-layer neural networks with \textbf{Tanh} as the activation function, and the width of hidden layer is $5$. The student network is also a two-layer neural network with \textbf{Tanh}, where we fix the parameters in the first layer and only train the second layer. The hidden layer dimension is $100$ (i.e. $d=100$). For the value of $\lambda$, we either draw from \textbf{Beta(1,1)} distribution in each epoch or fix it to $0.5$ in each epoch. We choose $n=20$ (so that $n<d$ is the overparameterized regime and $m\geq n^2>d$ is the underparameterized regime) and the learning rate is $0.1$. We use full-batch gradient descent to train the student network with MSE. Notice that here the ``full-batch'' indicates the batch size is equal to $n$, so that we can fairly compare the fixed $\lambda$ and random $\lambda$. More details can be found in our codes.

\subsection{Ablation Study: Effect of Fixed Mixing Coefficient}

\begin{figure}[htbp]
    \centering
       \includegraphics[width=0.6\linewidth]{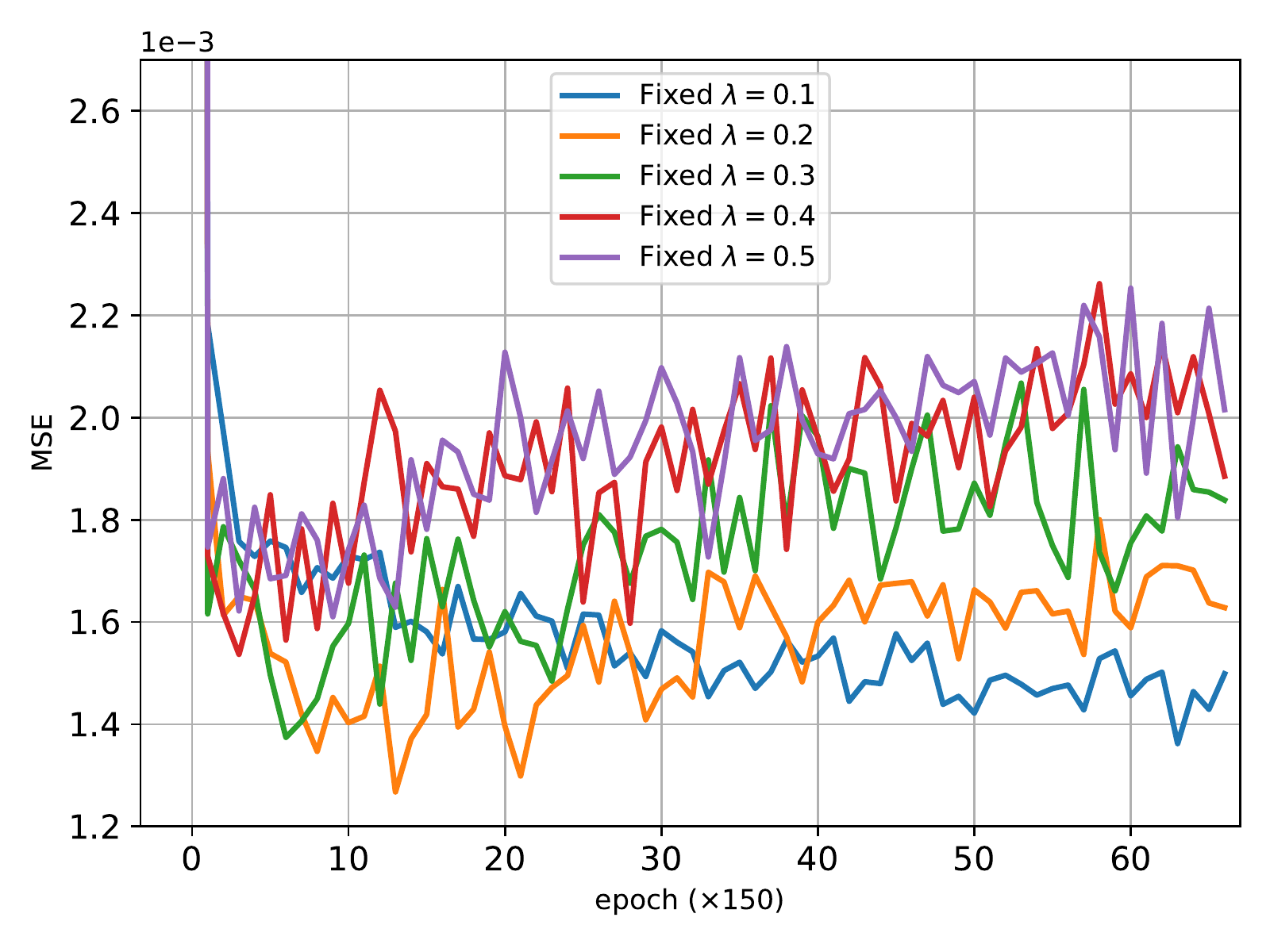}
    \caption{Results of the ablation study on $\lambda$.}
    \label{fig: TS-ablation}
\end{figure}

In the teacher-student setting, we experiment with different fixed values of $\lambda$, and the results are presented in Figure~\ref{fig: TS-ablation}. Of particular interest is the observation that as the noise level increases and $\lambda$ approaches 0.5, the turning point of testing error occurs earlier. This finding is consistent with our theoretical results.
% We compare different fixed values of $\lambda$ for the teacher-student setting, and the results are shown in Figure~\ref{fig: TS-ablation}. Notably, when $\lambda$ is close to $0.5$, i.e. having higher noise level, the turning point of testing error will come earlier, which is consistent with our theoretical results.

\section{Omitted Definitions and Proofs}
\label{sec:proofs}
\begin{defn}[Total Variation]
\label{defn:tv}
The total variation between two probability measures  $P$ and $Q$ is $\mathrm{TV}(P,Q)\triangleq\sup_E\left|P(E)-Q(E)\right|$, where the supremum is over all measurable set $E$.
\end{defn}

\begin{lem}[{\citep[Proposition~4.2]{levin2017markov}}]
\label{lem:TV}
Let $P$ and $Q$ be two probability distributions on $\mathcal{X}$. If $\mathcal{X}$ is countable, then
\[
\mathrm{TV}(P,Q)=\frac{1}{2}\sum_{x\in\mathcal{X}}\left|P(x)-Q(x)\right|.
\]
\end{lem}

\begin{lem}[Coupling Inequality {\citep[Proposition~4.7]{levin2017markov}}]
Given two random variables $X$ and $Y$ with probability distributions $P$ and $Q$, any
coupling $\hat{P}$ of $P$, $Q$ satisfies
\[
\mathrm{TV}(P,Q)\leq\hat{P}(X\neq Y).
\]
\label{lem:coupling}
\end{lem}

\subsection{Proof of Lemma~\ref{lem:mixup loss lower bound}}
\begin{proof}
We first prove the closed-form of the cross-entropy loss's lower bound. For any two discrete distributions $P$ and $Q$ defined on the same probability space $\mathcal{Y}$, the KL divergence of $P$ from $Q$ is defined as follows:
\begin{equation}
    D_\text{KL}(P\Vert{Q}):=\sum_{y\in\mathcal{Y}}P(y)\log\bigg(\dfrac{P(y)}{Q(y)}\bigg).
    \label{kl divergence}
\end{equation}
It is non-negative and it equals $0$ if and only if $P=Q$. 

Let's denote the $i^{th}$ element in $f_\theta(x)$ by $f_\theta(x)_i$. By adapting the definition of the cross-entropy loss, we have:
\begin{equation}
\begin{aligned}
    \ell(\theta,(\textbf{x},\textbf{y}))&=-\textbf{y}^\text{T}\log\big(f_\theta(\textbf{x})\big)\\
    &=-\sum_{i=1}^{K}y_i\log\big(f_\theta(\textbf{x})_i\big)\\
    &=-\sum_{i=1}^{K}y_i\log\bigg(\dfrac{f_\theta(\textbf{x})_i}{y_i}y_i\bigg)\\
    &=-\sum_{i=1}^{K}y_i\log\dfrac{f_\theta(\textbf{x})_i}{y_i}-\sum_{i=1}^{C}y_i\log{y_i}\\
    &=D_\text{KL}\big(\textbf{y}\Vert{f}_\theta(\textbf{x})\big)+\mathcal{H(\textbf{y})}\\
    &\geq\mathcal{H(\textbf{y})},
\end{aligned}
\label{eq: proof of lower bound of cross-entropy loss}
\end{equation}
where the equality holds if and only if ${f}_\theta(\textbf{x})=\textbf{y}$. Here $\mathcal{H(\textbf{y})}:=\sum_{i=1}^{C}y_i\log{y_i}$ is the entropy of the discrete distribution $\textbf{y}$. Particularly in ERM training, since $\textbf{y}$ is one-hot, by definition its entropy is simply $0$. Therefore the lower bound of the empirical risk is given as follows.
\begin{equation}
\begin{aligned}
    \hat{R}_S(\theta)&=\frac{1}{n}\sum_{i=1}^{n}\ell(\theta,(\textbf{x},\textbf{y}))\geq0
\end{aligned}
\end{equation}
The equality holds if ${f}_\theta(\textbf{x}_i)=\textbf{y}_i$ is true for each $i\in\{1,2,\cdots,n\}$.

We then prove the lower bound of the expectation of empirical Mixup loss. From Eq.~(\ref{eq: proof of lower bound of cross-entropy loss}), the lower bound of the general Mixup loss for a given $\lambda$ is also given by:
\begin{equation}
\begin{aligned}
    \ell(\theta,(\tilde{\textbf{x}},\tilde{\textbf{y}}))&\geq\mathcal{H(\tilde{\textbf{y}})}\\
    &=-\sum_{i=1}^{C}y_i\log{y_i}\\
    &=-\big(\lambda\log{\lambda}+(1-\lambda)\log(1-\lambda)\big).
\end{aligned}
\end{equation}
if $(\tilde{\textbf{x}},\tilde{\textbf{y}})$ is formulated via cross-class mixing. Recall the definition of the Mixup loss,
\begin{equation}
    \hat{R}_{\widetilde{S}}(\theta,\alpha)=\mathop{\mathbb{E}}\limits_{\lambda\sim{Beta(\alpha,\alpha)}}\frac{1}{n^2}\sum_{i=1}^{n}\sum_{j=1}^{n}\ell(\theta,(\tilde{\textbf{x}},\tilde{\textbf{y}})),
\end{equation}
we can exchange the computation of the expectation and the empirical average:
\begin{equation}
    \hat{R}_{\widetilde{S}}(\theta,\alpha)=\frac{1}{n^2}\sum_{i=1}^{n}\sum_{j=1}^{n}\mathop{\mathbb{E}}\limits_{\lambda\sim{Beta(\alpha,\alpha)}}\ell(\theta,(\tilde{\textbf{x}},\tilde{\textbf{y}}))
\end{equation}
Note that when $\alpha=1$, $Beta(\alpha,\alpha)$ is simply the uniform distribution in the interval $[0,1]$: $U(0,1)$. Using the fact that the probability density of $U(0,1)$ is constantly $1$ in the interval $[0,1]$,  the lower bound of $\mathop{\mathbb{E}}\limits_{\lambda\sim{Beta(1,1)}}\ell(\theta,(\tilde{\textbf{x}},\tilde{\textbf{y}}))$ where $\textbf{y}\neq\textbf{y}'$ is given by:
\begin{equation}
    \begin{aligned}
    \mathop{\mathbb{E}}\limits_{\lambda\sim{Beta(1,1)}}\ell(\theta,(\tilde{\textbf{x}},\tilde{\textbf{y}}))&\geq-\mathop{\mathbb{E}}\limits_{\lambda\sim{U(0,1)}}\big(\lambda\log{\lambda}+(1-\lambda)\log(1-\lambda)\big)\\
    &=-\int_0^1{\lambda\log{\lambda}+(1-\lambda)\log(1-\lambda)}\ d\lambda\\
    &=-2\int_0^{1}{\lambda\log{\lambda}}\ d\lambda\\
    &=-2\Bigg(\log\lambda\int_0^{1}\lambda\ {d}\lambda-\int_0^{1}{\frac{1}{\lambda}\bigg(\int_0^{1}\lambda\ {d}\lambda}\bigg)\ d\lambda\Bigg)\\
    &=-2\bigg(\dfrac{\lambda^2\log\lambda}{2}-\dfrac{\lambda^2}{4}\bigg)\bigg|_0^1\\
    &=0.5
    \end{aligned}
\end{equation}
Note that if the synthetic example is formulated via in-class mixing, the synthetic label is still one-hot, thus the lower bound of its general loss is $0$. In a balanced $C$-class training set, with probability $\frac{1}{C}$ the in-class mixing occurs. Therefore, the lower bound of the overall Mixup loss is given as follows,
\begin{equation}
    \hat{R}_{\widetilde{S}}(\theta,\alpha=1)\geq\dfrac{C-1}{2C}
\end{equation}
The equality holds if ${f}_\theta(\tilde{\textbf{{x}}})=\tilde{\textbf{y}}$ is true for each synthetic example $(\tilde{\textbf{x}},\tilde{\textbf{y}})\in\widetilde{S}$. This completes the proof.
\end{proof}

\subsection{Proof of Theorem~\ref{thm: noise-lower-bound}}
\begin{proof}
By the coupling inequality i.e. Lemma~\ref{lem:coupling}, we have
\[
\mathrm{TV}(P(\widetilde{Y}_{\rm h}|\widetilde{X}), P(\widetilde{Y}^*_{\rm h}|\widetilde{X}))\leq P(\widetilde{Y}_{\rm h}\neq \widetilde{Y}^*_{\rm h}|\widetilde{X}),
\]
Since $\mathrm{TV}(P(\widetilde{Y}_{\rm h}|\widetilde{X}), P(Y|X))=\mathrm{TV}(P(\widetilde{Y}_{\rm h}|\widetilde{X}), P(\widetilde{Y}^*_{\rm h}|\widetilde{X}))$, then the first inequality is straightforward.

For the second inequality, by Lemma~\ref{lem:TV}, we have
\begin{align*}
    \mathrm{TV}(P(\widetilde{Y}_{\rm h}|\widetilde{X}), P(\widetilde{Y}^*_{\rm h}|\widetilde{X}))=&\frac{1}{2}\sum_{j=1}^C\left|P(\widetilde{Y}^*=j|\widetilde{X})-P(\widetilde{Y}=j|\widetilde{X})\right| \\
    =&\frac{1}{2}\sum_{j=1}^C\left|f_{j}(\widetilde{X})-\left((1-\lambda) f_{j}(X) + \lambda f_{j}(X')\right)\right| \\
    \geq& \sup_j\frac{1}{2}\left|f_{{j}}(\widetilde{X})-\left((1-\lambda) f_{{j}}(X) + \lambda f_{{j}}(X')\right)\right|. 
    % \\
    % \geq& \frac{1}{2} \rho\lambda(1-\lambda)||X-X'||_2^2,
\end{align*}
This completes the proof.
\end{proof}

\subsection{Proof of Lemma~\ref{lem:model-dynamic}}
\begin{proof}
The ordinary differential equation of Eq.~(\ref{eq:gradient-flow}) (Newton's law of cooling) has the closed form solution
\begin{align}
\label{eq:ode-solution}
    \theta_t = \widetilde{\Phi}^{\dagger}\widetilde{\mathbf{Y}} + (\theta_0-\widetilde{\Phi}^{\dagger}\widetilde{\mathbf{Y}})e^{-\frac{\eta}{m}\widetilde{\Phi}\widetilde{\Phi}^T t}.
\end{align}
Recall that $\widetilde{\mathbf{Y}}=\widetilde{\mathbf{Y}}^*+\mathbf{Z}$, 
\begin{align*}
    \theta_t =& \widetilde{\Phi}^{\dagger}\left(\widetilde{\mathbf{Y}}^*+\mathbf{Z}\right) + (\theta_0-\widetilde{\Phi}^{\dagger}\left(\widetilde{\mathbf{Y}}^*+\mathbf{Z}\right))e^{-\frac{\eta}{m}\widetilde{\Phi}\widetilde{\Phi}^T t}\\
    =& \widetilde{\Phi}^{\dagger}\widetilde{\mathbf{Y}}^*+\widetilde{\Phi}^{\dagger}\mathbf{Z} + \left(\theta_0-\widetilde{\Phi}^{\dagger}\widetilde{\mathbf{Y}}^*\right)e^{-\frac{\eta}{m}\widetilde{\Phi}\widetilde{\Phi}^T t}-\widetilde{\Phi}^{\dagger}\mathbf{Z}e^{-\frac{\eta}{m}\widetilde{\Phi}\widetilde{\Phi}^T t}\\
    =&\theta^*+(\theta_0-\theta^*)e^{-\frac{\eta}{m}\widetilde{\Phi}\widetilde{\Phi}^T t} + (\mathbf{I}_d - e^{-\frac{\eta}{m}\widetilde{\Phi}\widetilde{\Phi}^T t})\theta^{\mathrm{noise}},
\end{align*}
which concludes the proof.
\end{proof}

\subsection{Proof of Theorem~\ref{thm:mixup-dynamic}}
\begin{proof}
We first notice that
\begin{align}
    R_t =& \mathbb{E}_{\theta_t,X,Y}\left|\left|\theta_t^T\phi(X)-Y\right|\right|_2^2\notag\\
    =&\mathbb{E}_{\theta_t,X,Y}\left|\left|\theta_t^T\phi(X)-\theta^{*T}\phi(X)+\theta^{*T}\phi(X)-Y\right|\right|_2^2\notag\\
    =&\mathbb{E}_{\theta_t,X}\left|\left|\theta_t^T\phi(X)-\theta^{*T}\phi(X)\right|\right|_2^2+\mathbb{E}_{X,Y}\left|\left|\theta^{*T}\phi(X)-Y\right|\right|_2^2+2\mathbb{E}_{\theta_t,X,Y}\langle\theta_t^T\phi(X)-\theta^{*T}\phi(X),\theta^{*T}\phi(X)-Y\rangle\notag\\
    \leq&\mathbb{E}_{X}\left|\left|\phi(X)\right|\right|_2^2\mathbb{E}_{\theta_t}\left|\left|\theta_t^T-\theta^{*T}\right|\right|_2^2+R^*+2\sqrt{\mathbb{E}_{\theta_t,X}\left|\left|\theta_t^T\phi(X)-\theta^{*T}\phi(X)\right|\right|_2^2\mathbb{E}_{X,Y}\left|\left|\theta^{*T}\phi(X)-Y\right|\right|_2^2}\notag\\
    \leq&\frac{C_1}{2}\mathbb{E}_{\theta_t}\left|\left|\theta_t^T-\theta^{*T}\right|\right|_2^2+R^*+2\sqrt{\frac{C_1R^*}{2}\mathbb{E}_{\theta_t}\left|\left|\theta_t^T-\theta^{*T}\right|\right|_2^2},\label{ineq:risk-bound}
    % \leq& 2\mathbb{E}_{\theta_t,X}\left|\left|\theta_t^T\phi(X)-\theta^{*T}\phi(X)\right|\right|_2^2+2R^*\\
    % +2\mathbb{E}_{\theta_t,X,Y}\left|\left|\theta_t^T\phi(X)-\theta^{*T}\phi(X)\right|\right|_2\left|\left|\theta^{*T}\phi(X)-Y\right|\right|_2\\
    %  \leq& 2\mathbb{E}_{X}\left|\left|\phi(X)\right|\right|_2^2\mathbb{E}_{\theta_t}\left|\left|\theta_t^T-\theta^{*T}\right|\right|_2^2+2R^*,
\end{align}
where the first inequality is by the Cauchy–Schwarz inequality and the second inequality is by the assumption.

Recall Eq.~(\ref{eq:ode-solution-2}),
\[
    \theta_t-\theta^* =   (\theta_0-\theta^*)e^{-\frac{\eta}{m}\widetilde{\Phi}\widetilde{\Phi}^T t} + (\mathbf{I}_d - e^{-\frac{\eta}{m}\widetilde{\Phi}\widetilde{\Phi}^T t})\widetilde{\Phi}^{\dagger}\mathbf{Z}.
\]

By eigen-decomposition we have
\[
\frac{1}{m}\widetilde{\Phi}\widetilde{\Phi}^T = V\Lambda V^T= \sum_{k=1}^d {\mu}_kv_k v_k^T,
\]
where $\{v_k\}_{k=1}^d$ are orthonormal eigenvectors and $\{{\mu}_k\}_{k=1}^d$ are corresponding eigenvectors.

Then, for each dimension $k$,
\begin{align*}
    \left(\theta_{t,k}-\theta_k^*\right)^2 \leq 2(\theta_{0,k}-\theta_k^*)^2e^{-2\eta \mu_k t}+2(1-e^{-\eta \mu_k t})^2\frac{mC_2}{m\mu_k},
\end{align*}

Taking expectation over $\theta_0$ for both side, we have
\begin{align}
     \mathbb{E}_{\theta_0}\left(\theta_{t,k}-\theta_k^*\right)^2 \leq 2(\xi^2_{k}+\theta_k^{*2})e^{-2\eta \mu_k t}+2(1-e^{-\eta \mu_k t})^2\frac{C_2}{\mu_k}.
     \label{ineq:dimension-square}
\end{align}

Notich that the RHS in Eq.~\ref{ineq:dimension-square} first monotonically decreases and then monotonically increases, so the maximum value of RHS  is achieved either at $t=0$ or $t\rightarrow \infty$. That is,
\begin{align}
    \mathbb{E}_{\theta_0}\left|\left|\theta_t^T-\theta^{*T}\right|\right|_2^2\leq\sum_{k=1}^d2\max\{\xi^2_{k}+\theta_k^{*2},\frac{C_2}{\mu_k}\}.
    \label{ineq:weight-norm}
\end{align}

Plugging Eq.~\ref{ineq:dimension-square} and Eq.~\ref{ineq:weight-norm} into Eq.~\ref{ineq:risk-bound}, we have
\begin{align*}
    R_t \leq &
    \frac{C_1}{2}\mathbb{E}_{\theta_t}\left|\left|\theta_t^T-\theta^{*T}\right|\right|_2^2+R^*+2\sqrt{\frac{C_1R^*}{2}\mathbb{E}_{\theta_t}\left|\left|\theta_t^T-\theta^{*T}\right|\right|_2^2}\\
    \leq& R^* + C_1\sum_{k=1}^d (\xi^2_{k}+\theta_k^{*2})e^{-2\eta \mu_k t}+(1-e^{-\eta \mu_k t})^2\frac{C_2}{\mu_k} + 2\sqrt{C_1R^*\zeta},
\end{align*}
where $\zeta=\sum_{k=1}^d\max\{\xi^2_{k}+\theta_k^{*2},\frac{C_2}{\mu_k}\}$.
This concludes the proof.
\end{proof}

% \section{Experiments on Covariate-Shift Datasets}

\end{document}